%% file: schatten.tex
\documentclass[backref,12pt]{colt2017} 

\usepackage[parfill]{parskip} 
\usepackage{fullpage, prettyref}
\usepackage{hyperref}
\hypersetup{backref} 

\usepackage{amssymb,amsfonts,amsmath,amscd,dsfont,mathrsfs}
 \usepackage{graphicx,float,psfrag}
\usepackage{wrapfig}
\usepackage{algorithm,algorithmic}
\usepackage{float}
\usepackage{cases}
\usepackage{pgf,pgfarrows,pgfnodes}
\usepackage{multicol, algorithmic,multirow}
\usepackage{float,psfrag}

\usepackage{wasysym}
\usepackage{pgfplots}
\usepackage{tikz}
\usetikzlibrary{arrows}
\usetikzlibrary{positioning}

\DeclareMathAlphabet{\mathpzc}{OT1}{pzc}{m}{it}

\usepackage{caption}
\usepackage{subcaption}
\usepackage{natbib}
\usepackage{array}
\newcolumntype{C}[1]{>{\centering\let\newline\\\arraybackslash\hspace{0pt}}m{#1}}

\newtheorem{propo}{Proposition}
\newtheorem{lemma}[propo]{Lemma}
\newtheorem{definition}[propo]{Definition}
\newtheorem{coro}{Corollary}
\newtheorem{thm}{Theorem}

\newtheorem{remark}[propo]{Remark}

\def\cH{\mathcal{H}}

\def\I{\mathbb{I}}

\def\i{i^\prime}

\def\cP{{\mathcal{P}}}
\def\Tr{{ \rm Tr}}
\def\diag{{\rm diag}}
\def\E{{\mathbb{E}}}
\def\P{{\mathbb{P}}}
\def\H{{\mathcal{H}}}
\def\reals{{\mathbb{R}}}

\def\<{\langle}
\def\>{\rangle}

\def\ones{{\mathds 1} }

\def\hTheta{{\widehat \Theta}}

\newcommand{\norm}[1]{\|#1\|}

\def\cM{{\mathcal{M}}}

\def\cD{{\mathcal{D}}}
\def\cE{{\mathcal{E}}}
\def\tv{{\rm{TV}}}
\newcommand{\ip}[2]{\left\langle #1, #2 \right\rangle}

\def\gam{{\gamma}}
\def\wgt{{\omega}}

\def\trace{{\rm tr}}
\def\hd{{\odot}}
\def\mm{{*}}
\def\sm{{\rm sum}}
\def\diag{{\rm diag}}
\newcommand{\val}[1]{\gam(#1)}

\title[Spectrum Estimation from a Few Entries]{Spectrum Estimation from a Few Entries}
\usepackage{times}

 \coltauthor{\Name{Ashish Khetan} \Email{khetan2@illinois.edu}\\
 \addr University of Illinois Urbana Champaign
 \AND
\Name{Sewoong Oh} \Email{swoh@illinois.edu}\\
    \addr University of Illinois Urbana Champaign}

\begin{document}
%
\maketitle
\begin{abstract}
	
	Singular values of a data in a matrix form provide insights 
	on the structure of the data, the effective dimensionality, and the choice of 
	hyper-parameters on higher-level data analysis tools. However, 
	in many practical applications such as collaborative filtering and network analysis,  
	we only get a partial observation. 
	Under such scenarios, we consider the fundamental problem of recovering 
	spectral properties of the underlying matrix from a sampling of its entries. 
	We are particularly interested in directly recovering 
	the spectrum, which is the set of singular values, 
	and also in sample-efficient approaches for recovering  a 
	spectral sum function, which is 
	an aggregate sum of the same function applied to  each of the singular values. 
	We propose first estimating the Schatten $k$-norms of a matrix, 
	and then applying Chebyshev approximation 
	to the spectral sum function or applying moment matching in Wasserstein distance 
	to recover the singular values. 
	The main technical challenge is in accurately estimating the Schatten norms 
	from a sampling of a matrix. 
	We introduce  a novel unbiased estimator based on counting small structures in a graph 
	and provide  guarantees that match its empirical performance.  
	Our theoretical analysis shows that Schatten norms can be recovered accurately 
	from strictly smaller number of samples compared to what is needed to recover the 
	underlying low-rank matrix. 
	Numerical experiments suggest that we significantly 
	improve upon a competing approach of 
	using matrix completion methods. 
	
%
\end{abstract}


%
%

\section{Introduction} 
\label{sec:intro}

Computing and analyzing the set of singular values of a data in a matrix form, 
which is called the spectrum, provide  
insights into the geometry and topology of the data. 
Such a spectral analysis is routinely a first step in general data analysis  
with the goal of checking if there exists a lower dimensional subspace explaining 
the important aspects of the data, which itself might be high dimensional. 
Concretely, it is a first step in dimensionality reduction methods such as 
 principal component analysis or canonical correlation analysis. 

However, spectral analysis becomes challenging in practical scenarios where 
the data is only partially observed. 
We commonly observe 
pairwise relations of randomly chosen pairs: 
 each user only rates a few movies in recommendation systems, 
 each player/team only plays against a few opponents in sports leagues, 
 each word appears in the same sentence with a small number of 
 other words  in word count matrices, and 
 each worker answers a few questions in crowdsourcing. 
In other applications, we have more structured samples. 
For example,  in a 
network analysis we might be interested in the 
spectrum of a large network, but only get to see the connections within a
small subset of nodes corresponding to sampling a sub-matrix of the adjacency matrix. 
Whatever the sampling pattern is, 
typical number of paired relations we observe is significantly 
smaller than the dimension of the data matrix. 

We study  all such variations in sampling patterns for partially observed data matrices, 
and ask the following fundamental question:
{\em can we  estimate spectral properties of a data matrix from partial observations?} 
We propose novel approaches for recovering 
spectral sum functions of the form \eqref{eq:defspectralsum} 
and the also the spectrum itself. 
A crucial building block in our approach is that 
spectral properties can be accurately approximated from the 
first few  moments of the spectrum  known as the 
 Schatten $k$-norms defined as 
\begin{eqnarray}
	\| M \|_k   &=& \Big(\sum_{i=1}^d \sigma_i(M)^k \Big) ^{1/k} \;, 
	\label{eq:defnorm1}
\end{eqnarray}
where $\sigma_1(M)\geq \sigma_2(M)\geq \cdots \geq \sigma_d(M)\geq 0$ are the 
singular values of the data matrix $M \in \reals^{d\times d}$. 
Once we obtain accurate 
estimates of Schatten $k$-norms, 
these estimates, as well as corresponding performance guarantees,   
can readily be  translated into accurate estimates of any spectral sum functions and also the spectrum of the matrix.

\subsection{Setup}
\label{sec:introsetup}

We want to estimate the Schatten $k$-norm   of a 
positive semidefinite matrix $M\in\reals^{d\times d}$
from a subset of its entries.  
The restriction to positive semidefinite matrices is primarily for notational convenience, and 
our analyses, the estimator, and the efficient algorithms naturally generalize to any non-square matrices. 
Namely, we can extend our framework to  bipartite graphs and estimate Schatten $k$-norm 
of any matrix for any even $k$. 
Let $\Omega$ denote the set of indices of samples we are given and let 
$\cP_\Omega(M)  =   \{ (i,j, M_{ij})\}_{(i,j) \in \Omega}$ denote the samples. 
With a slight abuse of notation, we used  $\cP_\Omega(M)$ to also denote the $d\times d$ sampled matrix: 
\begin{eqnarray*}
	\cP_\Omega(M)_{ij} = \left\{ 
	\begin{array}{rl}
		M_{ij}& \text{ if }(i,j)\in \Omega\;,\\
		0&\text{ otherwise} \;,
	\end{array}
	\right.
\end{eqnarray*}
and it should be clear from the context which one we refer to. 
Although we propose a framework that generally applies to any probabilistic sampling, 
it is necessary to propose specific sampling scenarios to provide tight analyses on the performance.
Hence, we focus on two types of sampling models: {\em Erd\"os-R\'enyi sampling} and {\em graph sampling}.

\begin{figure}[h]
 	\begin{center}
	\includegraphics[width=0.45\textwidth]{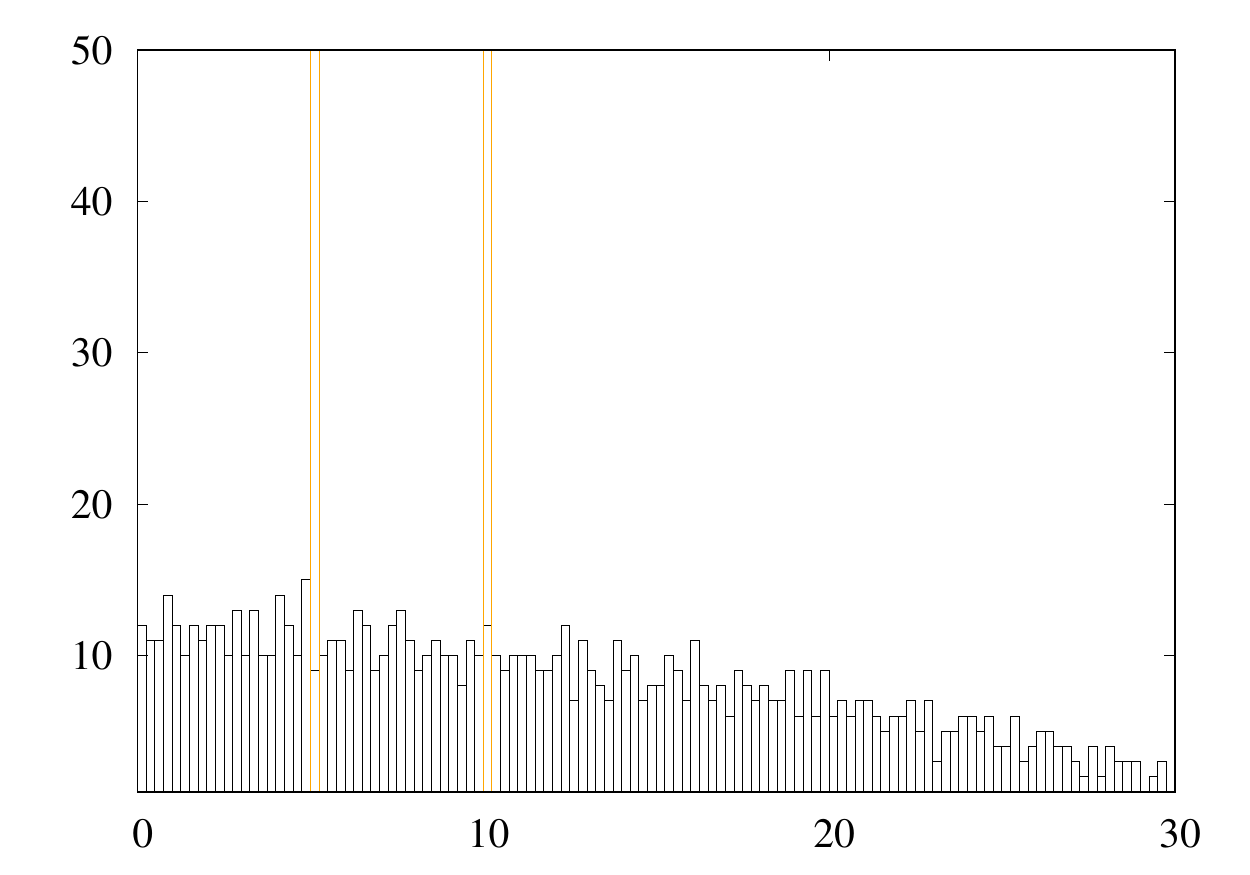} 
	\put(-105,120){{true spectrum}}
	\put(-127,117){\huge{$\leftarrow$}}
	\put(-120,60){{sampled spectrum}}
	\put(-100,43){\huge{$\downarrow$}}
	\caption{Histogram of singular values of a positive semi-definite matrix $M \in \reals^{d \times d}$ of size $d = 1000$ with rank $r=100$, and singular values of the sampled matrix where each entry of $M$ is sampled with probability $p=(1/d)r^{1-2/7}$ (properly rescaled by $1/p$ to best match the original spectrum).}  
	\label{fig:hist}
	\end{center}	
	\end{figure}
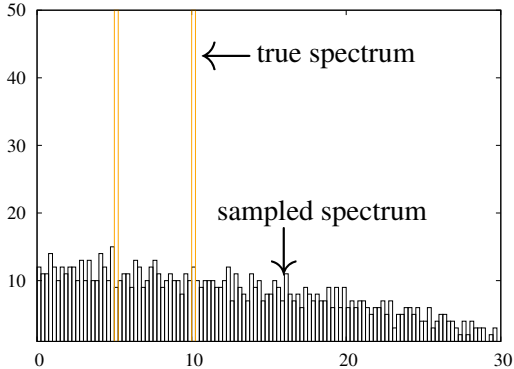

There is an extensive line of research in  low-rank matrix completion problems \citep{CR09,KMO10IT}, which addresses 
a fundamental question of how many samples are required to {\em complete} a matrix 
(i.e. estimate all the missing entries) from a small subset of sampled entries. 
It is typically assumed that each entry of the matrix is sampled independently 
with a probability $p\in(0,1]$. We refer to this scenario as 
 {\em Erd\"os-R\'enyi sampling}, as the resulting pattern of the samples encoded as  a graph  is distributed as 
 an Erd\"os-R\'enyi random graph. 
The spectral properties of such an sampled matrix   have been 
well studied in the literature \citep{FKS89,AM01,FeO05,KMO10IT,LLV15}. 
In particular, it is known that the original matrix is close in spectral norm 
to the sampled one where the missing entries are filled in with zeros and properly rescaled 
under certain incoherence assumptions. 
This suggests  using the singular values of $(d^2/|\Omega|)\cP(M)$ directly for estimating the Schatten norms. 
 However, in the sub-linear regime in which the number of samples $|\Omega| = d^2p $ is  comparable to or significantly smaller than
the degrees of freedom in representing a symmetric rank-$r$ matrix, which is $dr-r^2$, the spectrum of the sampled matrix is significantly different from the spectrum of the original matrix as shown in Figure \ref{fig:hist}.   
We need to design novel estimators that are more sample efficient
in the sub-linear regime where $d^2p \ll dr$. 

The Erd\"os-R\'enyi sampling has been criticized as being too strict for explaining how real-world datasets are sampled. 
When working with natural data, we typically only get one instance of a sampled matrix 
without the knowledge of how those entries are sampled. 
We propose {\em graph sampling}, a new sampling model that makes minimal assumptions about how the 
data was sampled. 
We assume that the {\em pattern} has been determined a priori, 
which is represented by a deterministic graph $G=(V,E)$ with $d$  nodes 
denoted by $V$ and undirected edges denoted by $E$.  
The random sampling $\Omega$ is chosen uniformly at random over all relabeling 
of the nodes in $G$. Formally, for a given $G=(V,E)$, a permutation $\pi:[d]\to V$ is drawn uniformly at random and samples are drawn according to 
\begin{eqnarray} 
	\label{eq:sample} 
	\cP_\Omega(M) \; = \;  \{ (i,j, M_{ij})\}_{(\pi(i),\pi(j)) \in E}\;.  
\end{eqnarray} 
As the sampling pattern $G$ is completely known to the statistician who 
only has one instance of a random sampling, 
we are only imposing that the samples are drawn uniformly at random from all instances that share the same pattern. 
Further, understanding this graph sampling model 
has a potential to reveal the subtle dependence of the estimation problem to 
the underlying pattern, which is known to be hard even for an established area of matrix completion.  



\subsection{Summary of the approach and preview of results}
\label{sec:intromain}
We propose first estimating one or a few  Scahtten norms, 
which can be  accurately estimated from samples, 
and using these estimated Schatten norms to approximate the spectral properties of interest: 
spectral sum functions and the spectrum. 
We use  
 an alternative expression of  the Schatten $k$-norm for positive semidefinite 
matrices as the trace of the $k$-th power of $M$, i.e.  
 $ (\| M \|_k)^k   =  \Tr(M^k)$. 
 This sum of the entries along the diagonal of $M^k$ is the  
 sum of total weights of all the closed walks of length $k$. 
 Consider the entries of $M$ as weights on a complete graph $K_d$ over $d$ nodes (with self-loops).   
 A closed walk of length $k$ is defined as a sequence of nodes 
$w=(w_1,w_2,\ldots,w_{k+1})$ with $w_1=w_{k+1}$, where we allow repeated nodes and repeated edges. 
The {\em weight} of a closed walk $w=(w_1,\ldots,w_{k},w_1)$ 
is defined as $\wgt_M(w) \equiv \prod_{i=1}^k M_{w_iw_{i+1}}$, 
which is the product  of the weights along the walk. 
It follows that 
\begin{eqnarray}
	\|M\|_k^k \;\; =\;\;  \sum_{w:\text{ all length $k$ closed walks}}  \wgt_M(w) \;. \label{eq:defnorm2}
\end{eqnarray}
Following the notations from enumeration of small simple cycles in a graph by \cite{AYZ97}, 
we partition this summation  
into those with the same pattern $H$ that we call a {\em $k$-cyclic pseudograph}. 
Let  $C_k=(V_k,E_k)$  denote the undirected simple cycle graph with $k$ nodes, e.g. $A_3$ in Figure \ref{fig:3cyclic} is $C_3$. 
We expand the standard notion of simple $k$-cyclic graphs to include multiedges and loops, 
hence the name {\em pseudograph}.  

\begin{definition}{\rm 
We define an unlabelled and undirected pseudograph $H=(V_H,E_H)$ to be  
a {\em $k$-cyclic pseudograph} for $k\geq3$ if   
there exists an onto node-mapping from $C_k=(V_k,E_k)$, i.e. $f:V_k \to V_H$, and a one-to-one edge-mapping $g:E_k\to E_H$ 
such that $g(e)=(f(u_e),f(v_e))$ for all $e=(u_e,v_e) \in E_k$. 
We use $\cH_k$ to denote the set of all $k$-cyclic pseudographs. 
We use   $c(H)$ to 
the number of different node mappings $f$ from $C_k$ to a $k$-cyclic pseudograph $H$. 
}
\label{def:pseugograph}
\end{definition}

\input{cycles3.tex}
In the above example, 
each member of $\cH_3$ is a distinct pattern that can be mapped from $C_3$. 
For $A_1$, it is clear that there is only one mapping from $C_3$ to $A_1$ (i.e.~$c(A_1)=1$). 
For $A_2$, one can map any of the three nodes to the left-node of $A_2$, hence 
$c(A_2)=3$. 
For $A_3$, any of the three nodes can be mapped to the bottom-left-node of $A_3$ and 
also one can map the rest of the nodes clockwise or counter-clockwise, resulting in 
$c(A_3)=6$. For $ k \leq 7$, all the $k$-cyclic pseudo graphs are given in the Appendix \ref{sec:graphs} (See Figures \ref{fig:4cyclic}--\ref{fig:7cyclic}).

Each closed walk $w$ of length $k$  is associated with one of the graphs in $\cH_k$, 
as there is a unique $H$ that the walk is an Eulerian cycle of (under a one-to-one mapping of the nodes). 
We denote this graph by $H(w)\in \cH_k$. 
Considering the weight of a walk $\wgt_M(w)$, there are multiple distinct walks 
with the same weight. 
For example, a length-3 walk $w=(v_1,v_2,v_2,v_1)$ has $H(w) = A_2$ and there are $3$ walks with the same weight $\wgt(w)=(M_{v_1v_2})^2M_{v_2v_2}$, i.e. 
$(v_1,v_2,v_2,v_1)$, $(v_2,v_2,v_1,v_2)$, and $(v_2,v_1,v_2,v_2)$.  
This multiplicity of the weight depends only on the structure $H(w)$ of a walk, and it is exactly $c(H(w))$ the number of 
mappings from $C_k$ to $H(w)$  in Definition \ref{def:pseugograph}. 
The total sum of the weights of closed walks of length $k$ can be partitioned into their respective pattern, which will make computation 
of such terms more efficient (see Section \ref{sec:algorithm}) and also de-biasing straight forward (see Equation \eqref{eq:estimate}):  
\begin{eqnarray} 
	\label{eq:defnorm3}
	\|M\|_k^k &=& \sum_{H \in \cH_k} \wgt_{ M }(H) \, c(H)\,, 
\end{eqnarray}
where with a slight abuse of a notation, 
we let $\wgt_M(H)$ for $H\in \cH_k$ be the sum of all {\em distinct} weights of walks $w$ with $H(w) = H$, 
and $c(H)$ is the multiplicity of each  of those distinct weights.
This gives an alternative tool for computing the Schatten $k$-norm without explicitly computing the singular values. 

Given only the access to a subset of sampled entries, 
one might be tempted to apply the above formula to the sampled matrix with an appropriate scaling, 
i.e.  
$\|(d^2/|\Omega|) \cP_\Omega(M)\|_k^k= (d^2/|\Omega|) \sum_{H \in \cH_k} \wgt_{ \cP_\Omega(M) }(H) \, c(H)\,, $ 
to estimate $\|M\|_k^k$. 
However, this is significantly biased. 
To eliminate the bias, we propose rescaling each term in \eqref{eq:defnorm2}
by the inverse of the probability of sampling that particular walk $w$ 
(i.e. the probability that all edges in $w$ are sampled). 
A crucial observation is that, for any sampling model that is invariant under a relabelling of the nodes, 
this probability only depends on the pattern $H(w)$.  
In particular, this is true for both Erd\"os-R\'enyi and graph sampling. 
Based on this observation, we introduce a novel estimator that 
 de-biases each group separately: 
\begin{eqnarray}
	\label{eq:estimate}
	   \hTheta_k(\cP_\Omega(M)) \;\; = \;\; \sum_{H \in \cH_k} \frac{1}{p(H)} \,\wgt_{\cP_{\Omega}(M)}(H) \, c(H) \;. 
\end{eqnarray}
It immediately follows that this estimator is unbiased, i.e. $\E_\Omega[\hTheta_k(\cP_\Omega(M))] = \|M\|_k^k$,    
where the randomness is in $\Omega$. 
However, computing this estimate can be challenging. 
Naive enumeration over all closed walks of length $k$ takes time scaling as $O(d \, \Delta^{k-1})$, 
where $\Delta$ is the maximum degree of the graph. 
Except for extremely sparse graphs, this is impractical.   
Inspired by the work of \cite{AYZ97} in counting short cycles in a graph, 
we introduce a novel and efficient method for computing the proposed estimate   for small values of $k$. 
\begin{propo} 
	For a positive semidefinite matrix $M$ and any sampling pattern $\Omega$, the proposed estimate $\hTheta_k(\cP_\Omega(M))$ in \eqref{eq:estimate} can be computed in time 
 $O(d^\alpha)$ for $k\in\{3,4,5,6,7\}$, where $\alpha<2.373$ is the exponent of matrix multiplication.
For $k=1$ or $2$, $\hTheta_k(\cP_\Omega(M))$ can be computed in time $O(d)$ and $O(d^2)$, respectively. 
	\label{pro:algorithm}
\end{propo} 
This bound holds regardless of the degree, and 
the complexity can be even smaller for sparse graphs as matrix multiplications are more efficient.  
We give a constructive proof by introducing a novel algorithm achieving this complexity in Section \ref{sec:algorithm}. 
For $k\geq8$, our approach can potentially be extended, but 
the complexity of the problem fundamentally changes as it is 
at least as hard as counting $K_4$ in a graph, for which 
the best known run time is $O(d^{\alpha+1})$ for general graphs  \citep{KKM00}. 

We make the following contributions in this paper: 
\begin{itemize}
	\item We give  in \eqref{eq:estimate} an unbiased estimator of the Schatten $k$-norm of a positive semidefinite matrix $M$, from 
	a random sampling of its entries. In general, the complexity of computing the estimate scales as $O(d \Delta^k)$ 
	where $\Delta$ is the maximum degree (number of sampled entries in a column) in the sampled matrix. 
	We propose a novel efficient algorithm for computing the estimate 
	in \eqref{eq:estimate} exactly for small  $k\leq 7$, which involves only matrix operations. 
	This algorithm is significantly more efficient and has 
	run-time scaling as $O(d^\alpha)$ independent of the degree and for all $k\leq 7$ (see Proposition \ref{pro:algorithm}) .  
	
	\item Under the typical Erd\"os-R\'enyi sampling, 
	we show that the Schatten $k$-norm of an incoherent rank-$r$ matrix can be approximated 
	within any constant multiplicative error, with number of samples scaling as $O(dr^{1-2/k})$ (see Theorem \ref{thm:er_ub}). 
	In particular, this is strictly smaller than the number of samples necessary to complete the matrix, 
	which scales as $O(d r \log d)$.
	Below this matrix completion threshold, 
	numerical experiments confirm that 
	the proposed estimator  significantly outperforms simple 
	heuristics of using singular values of the sampled matrices directly 
	or applying state-of-the-art matrix completion methods (see Figure \ref{fig:compare}).

	\item Given estimation of first $K$ Schatten norms, 
	it is straight forward to approximate 
	spectral sum functions of the form \eqref{eq:defspectralsum} 
	using Chebyshev's expansion, 
	and also estimate the spectrum itself using 
	moment matching in Wasserstein distance. 
	We apply our Schatten norm estimates to the application of 
	estimating the generalized rank studied in \cite{ZWJ15} and 
	estimating the spectrum studied in \cite{KV16}. 
	We provide performance guarantees
	for both applications and provide 
	experimental results suggesting we improve upon other competing methods. 

	\item We propose a new  model of  {\em graph sampling}  that preserves the structural properties of the pattern of the samples. 
	We identify a fundamental property of the structure of the pattern $\lambda^*_{G,r}$ that captures the 
	difficulty of estimating the Schatten $k$-norm from such graph sampling (see Theorem \ref{thm:lb}).  
	Under this graph sampling, we show that there are sampling patterns that are significantly more efficient 
	than Erd\"os-R\'enyi sampling. 
	If we sample from a clique, then it is necessary to have number of samples scaling as 
	$O(r^{2-4/k})$ to accurately estimate the Schatten $k$-norm for general values of $k$ 
	and it is also sufficient for special case of $k=3$ (see Lemma \ref{lem:lb} and Theorem \ref{thm:graph_ub}). 
	
	Although our analysis does not give a matching 
	lower bound on sample complexity for Erd\"os-R\'enyi sampling, 
	we show that there exists sampling patterns that 
	require significantly more samples than clique sampling. 
	It is necessary to have number of samples scaling as $O(dr^{1-4/k})$, if the pattern is a clique-star (see Lemma \ref{lem:lb2}).
\end{itemize}

In the remainder, we review existing work in Schatten norm approximation, 
and  provide an efficient implementation of the estimator \eqref{eq:estimate} for small $k$ in Section \ref{sec:algorithm}.  In Section \ref{sec:er}, we provide a theoretical analysis of 
our estimator under the Erd\"os-R\'enyi sampling scenario. 
In Section \ref{sec:gs}, we provide a theoretical analysis under the graph sampling scenario. 
We conclude with a discussion on interesting observations and remaining challenges 
in Section \ref{sec:discussion}.

\subsection{Related work}

We  review existing methods in approximating the Schatten norms, counting small structures in graphs, and various applications of Schatten norms. 

 \noindent
{\bf Estimating $k$-Schatten norms of a data matrix.} 
The proposed Schatten norm estimator  can be used as a black box in 
 various applications 
where we want to test the property of a data matrix or a network but limited to observe only a small portion of the data. 
These include, for example, network forensics, matrix spectral property testing, and testing for graph isospectral properties. 
Relatively little is known under the matrix completion setting studied in this paper.  
However, Schatten norm estimation under different resource constrained scenarios have been studied. 
\cite{Hut90} propose a randomized algorithm for approximating the trace of any large matrix, 
where the constraint is on the computational complexity. 
The goal is to design a random rank-one linear mapping such that the trace is preserved in expectation and the variance is small \citep{AT11, RA15}. 
\cite{LNW14} propose an optimal bilinear sketching of a data matrix, where the constraint is on the memory, 
i.e. the size of the resulting sketch. 
The goal is to design a sketch of a data matrix $M$ using minimal storage  
and a corresponding approximate reconstruction method for $\|M\|_k^k$. 
 \cite{LW16} propose an optimal streaming algorithm where only one-pass on the data is allowed in a data stream model  
 and the constraint is on the space complexity of the algorithm. 
 The goal is to design a streaming algorithm using minimal space to estimate $\|M\|_k^k$. 
\cite{ZWJ15} propose an estimator under a distributed setting where columns of the data are store in distributed storage and 
the constraint is on the communication complexity. 
The goal is to design a distributed protocol minimizing the communication to estimate $\|M\|_k^k$. 
Given a random vector $X$, \cite{KV16} propose an optimal estimator for the Schatten $k$-norm of the covariance matrix, 
where the constraint is on the number of samples $n$. 
The goal is to design an estimator using minimum number of samples to estimate $\|\E[XX^T]\|_k^k$. 

\begin{figure}[h]
 \begin{center}
\includegraphics[width=0.4\textwidth]{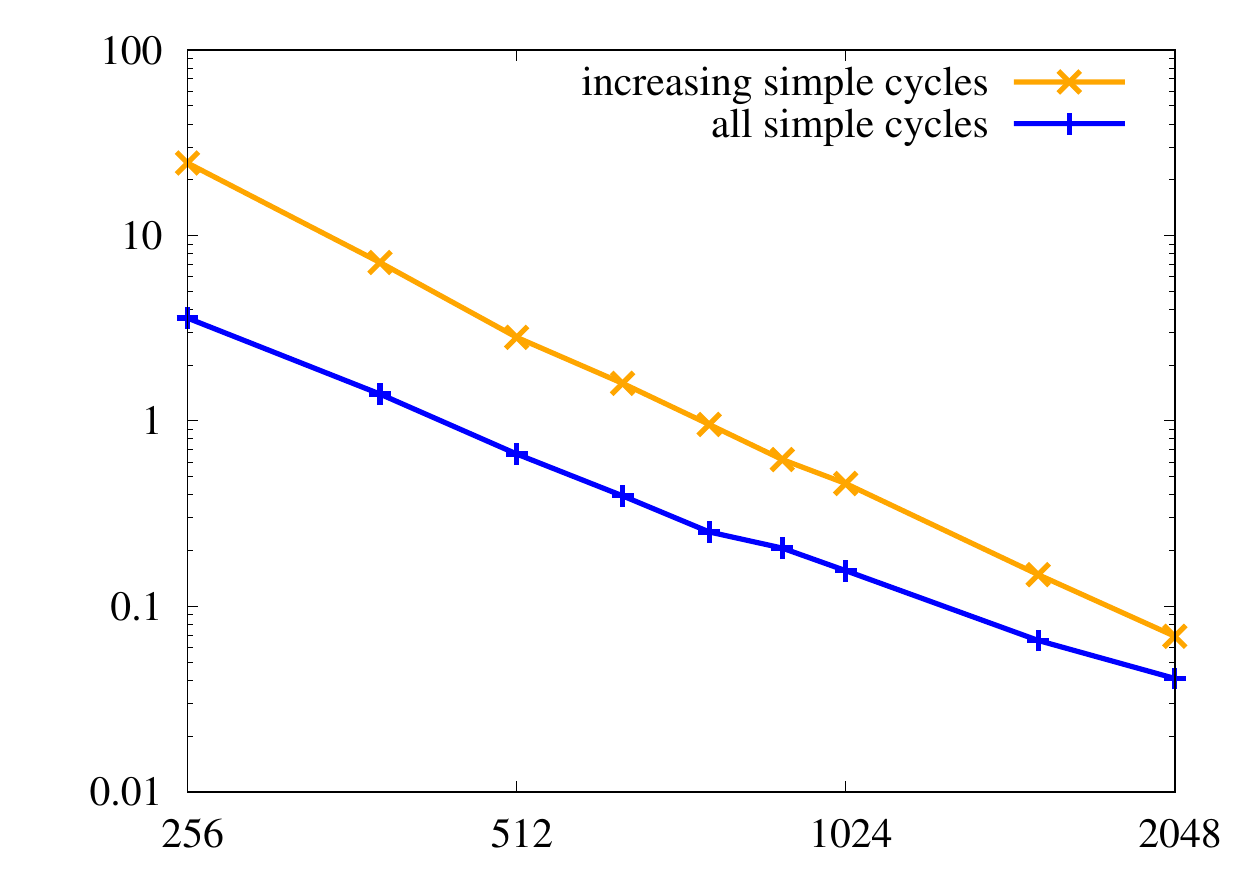} 
 	\put(-135,-7){{\small number of samples, $n$}}
 	\put(-265,60){{\small $\frac{\widehat {|\norm{\E[ XX^T]}_k^k} - \norm{\E[ XX^T]}_k^k|}{\norm{\E[ XX^T]}_k^k}$}}
	\caption{By replacing \citep[Algorithm 1]{KV16} that only counts increasing cycles with 
	our proposed algorithm that counts all cycles, significant gain is acheived in 
	 estimating $\norm{\E[XX^T]}_k^k$, for $k = 7$.}
	\label{fig:figSPComp}   
\end{center}	
\end{figure}

One of our contribution is that we propose an efficient algorithm for computing the weighted counts of small structures in Section \ref{sec:algorithm}, 
which can significantly improve upon less sample-efficient counterpart in, for example,  \citep{KV16}. 
Under the setting of \citep{KV16} (and also  \citep{LNW14}), 
the main idea of the estimator is that 
the weight of each length-$k$ cycle in the observed empirical covariance matrix $(1/n)\sum_{i=1}^nX_iX_i^T$ 
provides an unbiased estimator of $\|\E[XX^T]\|_k^k$. 
One prefers to sum over the weights of as many cycles as computationally allowed in order to reduce the variance. 
As counting all cycles is in general computationally hard, they propose counting only increasing cycles 
(which only accounts for only 1/k! fraction of all the cycles), which can be computed in time $O(d^\alpha)$. 
If one has an efficient method to count all the (weighted) cycles, then the variance of the estimator could potentially decrease 
by an order of $k!$. 
For $k\leq 7$, our proposed algorithm in Section \ref{sec:algorithm} provides exactly such an estimator. 
We replace \cite[Algorithm 1]{KV16} with ours, and run the same experiment to showcase the improvement in 
 Figure \ref{fig:figSPComp}, for dimension $d = 2048$ and various values of number of samples $n$ comparing 
 the multiplicative error in estimating $\|\E[XX^T]\|_k^k$, for $k = 7$.
 With the same run-time, significant gain is achieved by simply substituting our proposed algorithm for counting small structures, in the sub-routine. 
In general, the efficient algorithm we propose   might be of independent interest to 
various applications, and 
 can directly substitute (and significantly improve upon) other popular but less efficient counterparts. 

One of the main challenges 
under the sampling scenario considered in this paper is that  
existing counting methods like  that of \citep{KV16} cannot be applied, regardless of how much computational power we have.  
Under the matrix completion scenario, 
we need to $(a)$ sum over all small structures $H\in \cH_k$ and not just $C_k$ as in \citep{KV16}; 
and $(b)$ for each structure we need to sum over all subgraphs with the same structure 
and not just those walks whose labels form a monotonically increasing sequence as in \citep{KV16}.

\bigskip
\noindent
{\bf Algorithms for counting structures.} 
An important problem in graph theory is to count the number of small structures, also called network motifs,  in a given graph. 
This has many practical applications in 
designing good LDPC codes \citep{tian2004selective}, 
understanding the  properties social networks \citep{UBK13},  
and explaining gene regulation networks \citep{SMM02}. 
Exact and approximate algorithms have been proposed in \citep{AYZ97,KKM00,LW06,HC06,KB13,WLR14}. 
The most relevant one is the work of  \cite{AYZ97}  on counting the number of cycles $C_k$, 
where counts of various small structures called $k$-cyclic graphs are used as sub-routines and efficient 
approaches are proposed  for $k\leq 7$. 
These are  similar to $k$-cyclic pseudographs, but with multiedges condensed to a single edge.  
When counting cycles in a simple (unweighted) graph, $k$-cyclic graphs are sufficient as all the edges have weight one. 
Hence, one does not need to track how many times an edge has been traversed; 
the weight of that walk is one, regardless.  
In our setting, the weight of a walk depends on how many times an edge has been  traversed, which we track using multiedges. 
It is therefore crucial to introduce the class of $k$-cyclic pseudographs  in our estimator. 

Iin a distributed environment, fast algorithms for counting small structures 
have been proposed by 
\cite{elenberg2015beyond} and 
\cite{elenberg2016distributed} for small values of $k\in\{3,4\}$. 
However, the main strength of this approach is in distributed computing, and 
under the typical centralized setting we study, this approach can be slower by a factor exponential in $k$  for, say $k\leq 7$.  

\bigskip \noindent 
{\bf From Schatten norms to spectral sum functions.}  
A  dominant application of Schatten norms is in 
 approximating a family of 
 functions of a matrix, which are called 
{\em spectral sum functions}   \citep{HMAS16} of the form 
\begin{eqnarray}
	F(M;f) &\equiv& \sum_{i=1}^d f(\sigma_i(M)) \;\; \simeq \;\; \sum_{k=0}^K a_k \Big\{\sum_{i=1}^d\sigma_i(M)^k\Big\}  \label{eq:defspectralsum}
	\;. 
\end{eqnarray}  
A typical approach is to 
compute the coefficients of a Chebyshev approximation of $f$, 
which  immediately leads to an approximation of the spectral sum function of interest as 
the weighted sum of  Schatten $k$-norms. 
This approach has been widely used in 
fast methods for approximating the log-determinant \citep{PL04,ZL07,BDK15,ASE14,HMS15}, 
corresponding to $f(x)=\log x$. 
Practically,  log-determinant computations are routinely (approximately) required in 
applications including 
Gaussian graphical models \citep{RH05},  
minimum-volume ellipsoids \cite{VR09}, and metric learning \cite{DKJ07}. 
Fast methods for approximating trace of matrix inverse has been studied in \citep{WLK16,Che16}, corresponding to $f(x)=x^{-1}$, 
motivated by applications in  lattice quantum chromodynamics 
\citep{SLO13}.   
Fast methods for approximating the Estarada index has been studied in \citep{HMAS16}, corresponding to $f(x)=\exp(x)$. 
Practically, it is used in 
characterizing 3-dimensional molecular structure \citep{Est00} and  
measuring graph centrality \citep{EH07}, 
 the entropy of a graph \citep{Car08}, and 
 the bipartivity of a graph \citep{ER05}. 
Approximating the generalized rank under communication constraints 
has been studied in \citep{ZWJ15}, corresponding to $f(x;c_1)={\mathbb I}(x\leq c_1)$. 
The generalized rank approximates 
a necessary tuning parameter in a number of problems 
where low-rank solutions are sought including 
robust PCA \citep{rpca1,rpca2} and matrix completion \citep{KMO10JMLR,KMO10IT,JNS13}, 
and also is required in 
sampling based methods in numerical analysis \citep{rand1,rand2}. 
Similarly, \citep{saade2015matrix} studied the number of singular values in an interval, corresponding to $f(x;c_1,c_2) = {\mathbb I}( c_1\leq x \leq c_2)$. 
In practice, a number of eigensolvers \citep{eigen1,eigen2,eigen3} require the number of eigenvalues in an given interval. 
For more comprehensive list of references and applications of this framework, we refer to the related work section in \citep{HMAS16}. 

In a recent work, 
\cite{KV16} provide a novel approach to tackle the challenging problem of estimating the singular values themselves.  
Considering the histogram of the singular values as a 
one-dimensional distribution and the Schatten $k$-norm as the $k$-th moment of this distribution, the authors provide an innovative algorithm to estimate 
the histogram that best matches the moments in Wasserstein distance. 

\bigskip\noindent{\bf Matrix completion.} 
Low-rank matrix completion addresses the problem of 
recovering a low-rank matrix from its  sampled entries. 
Tight lower and upper bounds on the 
sample complexity is well studied 
in both cases where you want exact recovery when samples are noiseless \citep{CR09,KMO10IT,BJ14}, and also 
when samples are noisy and where you want approximate recovery \citep{KMO10JMLR,NW11}. 
In practical applications, one might not have enough samples to estimate all the missing entries with sufficient accuracy. 
However, one might still be able to infer important spectral properties of the data, such as 
the singular values or the rank. 
Such spectral properties can also assist in making decisions on 
how many more samples to collect in order to make accurate inference on the quantity of interest. 
In this paper, one of the fundamental question we ask and answer affirmatively is:  Can we accurately recover the spectral properties of a low-rank matrix from sampling of its entries, 
below the matrix completion threshold?

%
%
%

\section{Efficient Algorithm}
\label{sec:algorithm}

In this section we give a constructive proof of Proposition \ref{pro:algorithm}, inspired by the seminal work of 
 \cite{AYZ97} and generalize their counting algorithm for $k$-cyclic graphs for counting (weighted) $k$-cyclic pseudographs. 
 In computing the estimate in \eqref{eq:estimate}, 
$c(H)$ can be computed in time $O(k!)$ and suppose $p(H)$ has been computed 
(we will explain how to compute $p(H)$ for Er\"os-R\'enyi sampling and graph sampling in Sections \ref{sec:er} and \ref{sec:gs}).
The bottleneck then 
is computing the weights $\wgt_{\cP_\Omega(M)}(H)$ for each $H\in\cH_k$. 
Let $\gam_M(H) \equiv  \wgt_M(H)c(H)$.  
We give matrix multiplication based equations to compute $\gam_M(H)$ 
for every $H \in \mathcal{H}_k$ for $k \in \{3,4,5,6,7\}$. 
This establishes that $\gam_M(H)$, and hence $ \wgt_M(H)$, 
can be computed in time $O(d^\alpha)$, proving Proposition \ref{pro:algorithm}. 

For any matrix $A \in \mathbb{R}^{d \times d}$, let $\diag(A)$ to be a diagonal matrix such that $(\diag(A))_{ii} = A_{ii}$, for all $i \in [d]$ and $(\diag(A))_{i,j} = 0$, for all $i \neq j \in [d]$.  
For a given matrix $M \in \mathbb{R}^{d \times d}$, define the following: $O_{M}$ to be matrix of off-diagonal entries of $M$ that is $O_{M} \equiv M - \diag(M)$ and we let $D_{M} \equiv \diag(M)$. 
Let $\trace(A)$ denote trace of $A$, that is $\trace(A) = \sum_{i \in [d]} A_{ii}$, 
and let $A \mm B$ denote the standard matrix multiplication of two matrices $A$ and $B$ to make it more explicit. 
Consider computing $\gam_M(H)$ for $H \in \cH_3$ as labeled in Figure \ref{fig:3cyclic}: 
	\begin{eqnarray}
		\gam_M(A_1) &=& \trace(D_M \mm D_M \mm D_M) \label{eq:matrix3start}\\
		\gam_M(A_2) &=& 3\;\trace(D_M \mm O_M \mm O_M) \label{eq:matrix3middle}\\
		\gam_M(A_3) &=& \trace(O_M \mm O_M \mm O_M) \label{eq:matrix3end}
	\end{eqnarray}
The first weighted sum $\gam_M(A_1)$ is sum of all weights of walks of length 3 that consists of three self-loops. 
One can show that $\gam_M(A_1) = \sum_{i\in [d]} M_{ii}^3$, which in our matrix operation notations is \eqref{eq:matrix3start}. 
Similarly, $\gam_M(A_3)$ is the sum of weights of length 3 walks with no self-loop, which leads to \eqref{eq:matrix3end}.
$\gam_M(A_2)$ is the sum of weights of length 3 walks with a single self-loop, which leads to \eqref{eq:matrix3middle}. 
The factor 3 accounts for the fact that the self loop could have been placed at first, second, or third in the walk.

Similarly, for each $k$-cyclic pseudographs in $\cH_k$ for $k\leq 7$, 
computing $\gam_M(H)$ involves a few matrix operations with run-time $O(d^\alpha)$. 
We provide the complete set of explicit expressions in Appendix \ref{sec:matrix}. 
A MATLAB implementation of the estimator \eqref{eq:estimate}, 
that includes as its sub-routines the computation of the weights of 
all $k$-cyclic pseudographs, is available for download at \url{https://github.com/khetan2/Schatten_norm_estimation}. 
The explicit formulae in Appendix \ref{sec:matrix}  
together with the implementation in the above url might be of interest to 
other problems involving counting small structures in graphs. 

For $k=1$, the estimator simplifies to $\hTheta_k(\cP_\Omega(M))= (1/p) \sum_{i} \cP_\Omega(M)_{ii}$, which 
can be computed in time $O(d)$. 
For $k=2$, the estimator simplifies to $\hTheta_k(\cP_\Omega(M))= (1/p) \sum_{i,j} \cP_\Omega(M)_{ij}^2$, 
which can be computed in time $O(|\Omega |)$. 
However, for $k \geq 8$, there exists walks over $K_4$, a clique over 4 nodes, that cannot be decomposed 
into simple computations involving matrix operations. The best known algorithm for a simpler task 
of counting $K_4$ has run-time scaling as $O(d^{\alpha+1})$, which is fundamentally different. 
We refer to Section \ref{sec:discussion} for further discussions on the computational complexity beyond $k=7$.


\begin{algorithm}[ht]
\caption{Schatten $k$-norm estimator }
\label{algo:est}
  \begin{algorithmic}[1]
    \REQUIRE $\cP_\Omega(M)$, $k$, $\cH_k$, $p(H)$ for all $H \in \cH_k$
	\ENSURE $\hTheta_k(\cP_\Omega(M))$    
    	\IF{$k\leq 7$} 
		\STATE For each $H\in \cH_k$, compute $\gamma_{\cP_\Omega(M)}(H)$ using the formula from 
		Eq.~\eqref{eq:matrix3start}--\eqref{eq:matrix3end}  for $k=3$ and 
		Eq.~\eqref{eq:matrix4start} -- \eqref{eq:matrix74end} for $k\in\{4,5,6,7\}$
		\STATE $	   \hTheta_k(\cP_\Omega(M)) \leftarrow  \sum_{H \in \cH_k} \frac{1}{p(H)} \,\gamma_{\cP_\Omega(M)}(H) $		
    	\ELSE 
    	\STATE $\hTheta_k(\cP_\Omega(M)) \leftarrow$ Algorithm \ref{algo:kgreaterthan7}$[\cP_\Omega(M)$, $k$, $\cH_k$, $p(H)$ for all $H \in \cH_k]$  \hfill 	[Appendix \ref{sec:kgreaterthan7}] 
    	\ENDIF	
      \end{algorithmic}
\end{algorithm}

\section{Erd\"os-R\'enyi sampling}
\label{sec:er}

Under the stylized but canonical Erd\"os-R\'enyi sampling, 
notice that 
the probability $p(H)$  that we observe all edges in a walk  with pattern $H$ is 
\begin{eqnarray}
	p(H) \;\;= \;\; p^{m(H)} \;,
\end{eqnarray}
where $p$ is the probability an edge is sampled and 
$m(H)$ is the number of distinct edges in a $k$-cyclic pseudograph $H$. 
Plugging in this value of $p(H)$, which can be computed in time linear in $k$, 
into the estimator \eqref{eq:estimate}, we get an estimate customized for Erd\"os-R\'enyi sampling. 

Given a rank-$r$ matrix $M$, 
the difficulty of estimating properties of $M$ from sampled entries is captured by 
the {\em incoherence} of the original matrix $M$, which we denote by $\mu(M)\in\reals$ \citep{CR09}. 
Formally, let $M \equiv U \Sigma U^{\top}$ be the singular value decomposition of a positive definite matrix 
 where $U$ is a $d \times r$ orthonormal matrix and 
  $\Sigma \equiv {\rm diag}(\sigma_1,\cdots,\sigma_r)$ with singular values 
  $\sigma_1\geq \sigma_2\geq\cdots\geq \sigma_r >0$.
   Let $U_{i,r}$ denote the $i$-th row and $j$-th column entry of matrix $U$.
The incoherence $\mu(M)$ is defined as the smallest positive value $\mu$
such that the following holds: 
\begin{itemize}
\item[A1.] For all $i \in [d]$, we have $\sum_{a=1}^r U_{ia}^2 (\sigma_a/\sigma_1) \leq  {\mu r}/{d}$.
\item[A2.] For all $i \neq j \in [d]$, we have $|\sum_{a=1}^r U_{ia}U_{ja}(\sigma_a/\sigma_1)| \leq {\mu \sqrt{r}}/{d}$.
\end{itemize}
The incoherence measures how well spread out the matrix is 
and is a common measure of difficulty in completing a matrix from random samples \citep{CR09,KMO10IT}. 
The lower the incoherence, the more spread out the entries are, and estimation is easier. 
On the other hand,  if there a a few entries that are much larger than the rest, 
estimating a property of the matrix (such as the Schatten $k$-norm) 
from sampled entries can be extremely challenging.


\subsection{Performance guarantee} 
\label{sec:er_guarantee}

For any $d \times d$ positive semidefinite matrix $M$ of rank $r$ with 
incoherence $\mu(M)=\mu$ and the effective condition number $\kappa=\sigma_{\rm max}(M)/\sigma_{\rm min}(M)$,
we define 
\begin{eqnarray}
	\rho^2 \;\;\equiv\;\;  (\kappa \mu)^{2k}g(k)  \max \Bigg\{1, \frac{(dp)^{k-1}}{d}, \frac{r^{k}p^{k-1}}{d^{k-1}} \Bigg\}\,,
\end{eqnarray}
such that the variance of our estimator is bounded by 
${\rm Var}( \hTheta(\cP_\Omega(M))/\|M\|_k^k ) \leq \rho^2 (r^{1-2/k}/dp)^k$ as we show in the proof of Theorem \ref{thm:er_ub} in Section \ref{sec:proof_er_ub}. 
Here, $g(k)=O(k!)$ is a function  depending only on $k$.

\begin{thm}[Upper bound under the Erd\"os-R\'enyi sampling]
For any integer $k \in [3,\infty)$,  any $\delta > 0$, 
any rank-$r$ positive semidefinite matrix $M\in\reals^{d\times d}$,  and  
given i.i.d. samples of the entries of $M$ with probability $p$, 
the proposed estimate of \eqref{eq:estimate} 
achieves normalized error $\delta$ with probability bounded by 
	\begin{eqnarray}\label{eq:ub0}
\P\Bigg( \frac{\big|\hTheta_k(\cP_\Omega(M)) - \norm{M}_k^k \big|}{\norm{M}_k^k} & \geq & \delta  \Bigg) \;\; \leq \;\; 
	 \frac{\rho^2}{\delta^2} \Big(\frac{r^{1-2/k}}{dp}\Big)^k \;. 
	\end{eqnarray}
	\label{thm:er_ub}
\end{thm}


Consider a typical scenario where $\mu$, $\kappa$, and $k$ are  finite with respect to $d$ and $r$. 
Then the Chebyshev's bound in \eqref{eq:ub0} implies that 
the sample $d^2p = O(dr^{1-2/k})$ is sufficient to recover $\|M\|_k^k$ up to arbitrarily small multiplicative error 
and arbitrarily small (but strictly positive) error probability. 
This is strictly less than 
the known minimax sample complexity for recovering the entire low-rank matrix, which scales is $\Theta(rd\log d)$.
As we seek to estimate only a property of the matrix (i.e. the Schatten $k$-norm) and not the whole matrix itself, 
we can be more efficient on the sample complexity by a factor of $r^{2/k}$ in rank and a factor of $\log d$ in the dimension.  
We emphasize here that such a gain can only be established using the proposed estimator based on the structure of the $k$-cyclic pseudographs. We will show empirically that the standard matrix completion approaches fail in the critical regime of samples below the recovery 
threshold of $O(rd\log d)$.  


\begin{figure}[h]
 \begin{center}
	\includegraphics[width=0.5\textwidth]{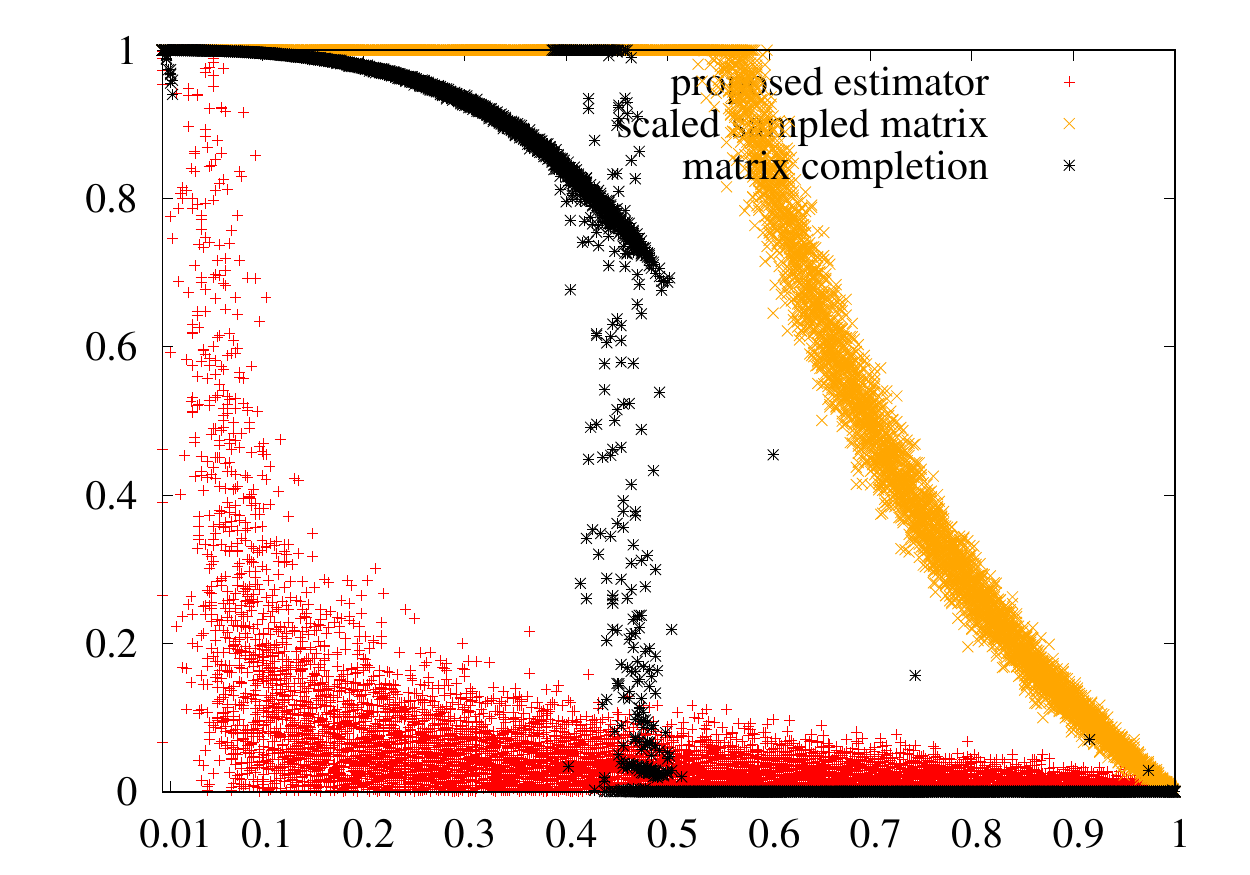}
	\put(-160,-3){\small{$\;$sampling probability, $p$}}
	\put(-230,47){\rotatebox{90}{\parbox{4cm}{\small{relative error}}}} 
	\put(-150,163){\large{$d=500$, $r =100$}}
	\includegraphics[width=0.5\textwidth]{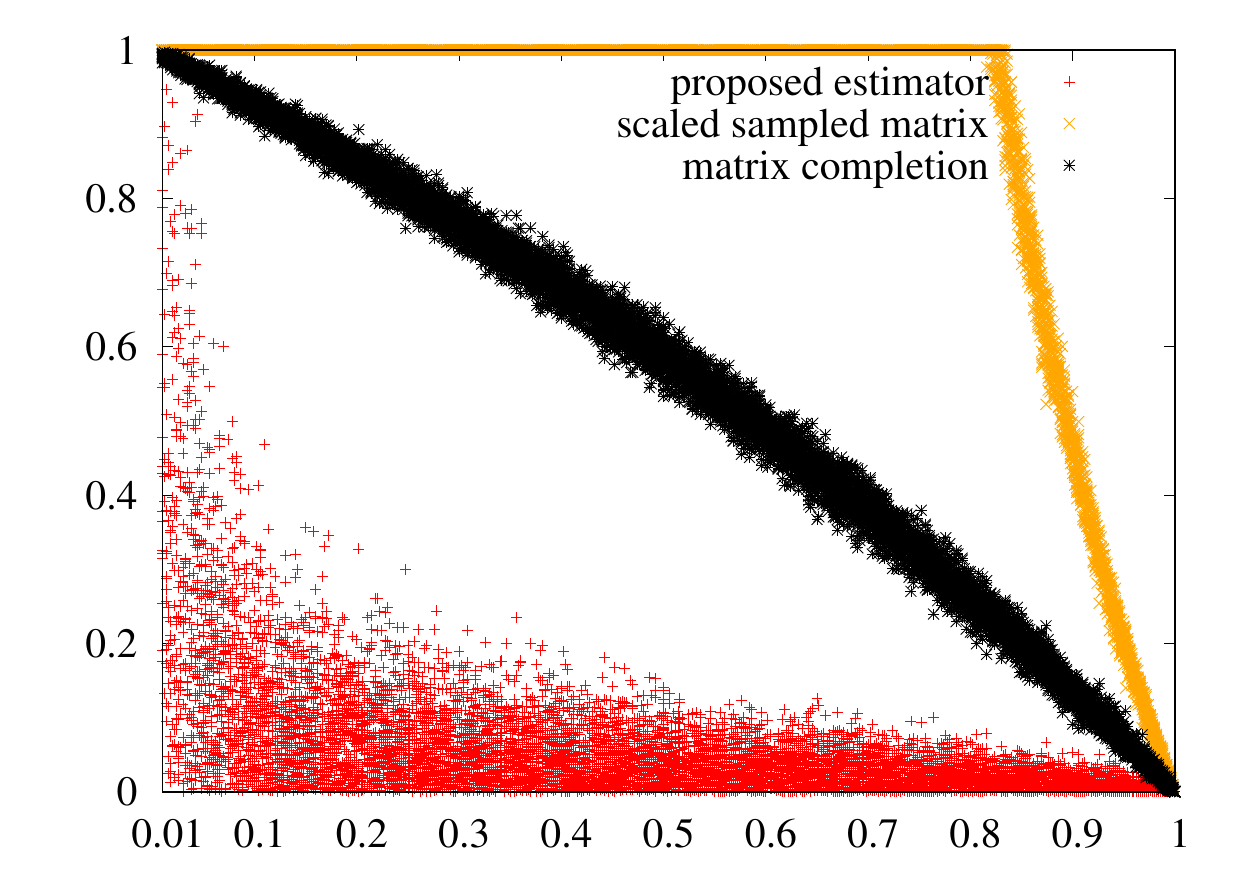}
	\put(-160,-3){\small{$\;$sampling probability, $p$}}
	\put(-230,47){\rotatebox{90}{\parbox{4cm}{\small{relative error}}}}
	\put(-150,163){\large{$d=500$, $r =500$}} 	
\end{center}
\caption{The proposed estimator outperforms both baseline approaches below the matrix completion threshold.  
For $k = 5$, comparison of the absolute relative error in estimated Schatten norm that is $\big|\|M\|_k^k - \widehat{\|M\|_k^k}\big|/\|M\|_k^k$ for the three algorithms: $(1)$ the proposed estimator, $\widehat {\|M\|_k^k} = \hTheta_k(\cP_\Omega(M))$, $(2)$ Schatten norm of the scaled sampled matrix, $\widehat{\|M\|_k^k} = \norm{(1/p)\cP_r(\cP_{\Omega}(M))}_k^k$, $(3)$ Schatten norm of the completed matrix, $\widetilde M = {\rm AltMin}(\cP_{\Omega}(M))$ from \citep{JNS13},  $\widehat{\|M\|_k^k} = \norm{\widetilde M}_k^k$, where $\cP_r(\cdot)$ is the standard best rank-$r$ projection of a matrix. $\Omega$ is generated by Erd\"os-R\'enyi  sampling of matrix $M$ with probability $p$.}
\label{fig:compare}
\end{figure}

Figure \ref{fig:compare} is a scatter plot of the absolute relative error in estimated Schatten $k$-norm, 
$\big|\|M\|_k^k - \widehat{\|M\|_k^k}\big|/\|M\|_k^k$, for $k=5$, for three approaches: 
the proposed estimator, Schatten norm of the scaled sampled matrix (after rank-$r$ projection), and Schatten norm of the completed matrix, using state-of-the-art alternating minimization algorithm \citep{JNS13}. 
All the three estimators are evaluated $20$ times for each value of $p$. $M$ is a symmetric positive semi-definite matrix of size $d = 500$, and rank $r = 100$ (left panel) and $r=500$ (right panel). Singular vectors $U$ of $M = U \Sigma U^\top$, are generated by QR decomposition of $\mathcal{N}(0,\I_{d \times d})$ and $\Sigma_{i,i}$ is uniformly distributed over  $[1,2]$.
For a low rank matrix on the left, there is a clear critical value of $p\simeq 0.45 $, above which matrix completion is exact with high probability. 
However, this algorithm knows the underlying rank and crucially exploits the fact that the underlying matrix is exactly low-rank. 
In comparison, our approach is agnostic to the low-rank assumption but 
 finds the accurate estimate that is adaptive to the actual rank in a data-driven manner. 
Using the first $r$ singular values of the (rescaled) sampled matrix fails miserably for all regimes (we truncate the error at one for illustration purposes). 
In this paper, we are interested in the regime where 
exact matrix completion is impossible as 
we do not have enough samples to exactly recover the underlying matrix: $p\leq 0.45$ in the left panel 
and all regimes in the right panel.

\begin{figure}[h]
 \begin{center}
	\includegraphics[width=0.45\textwidth]{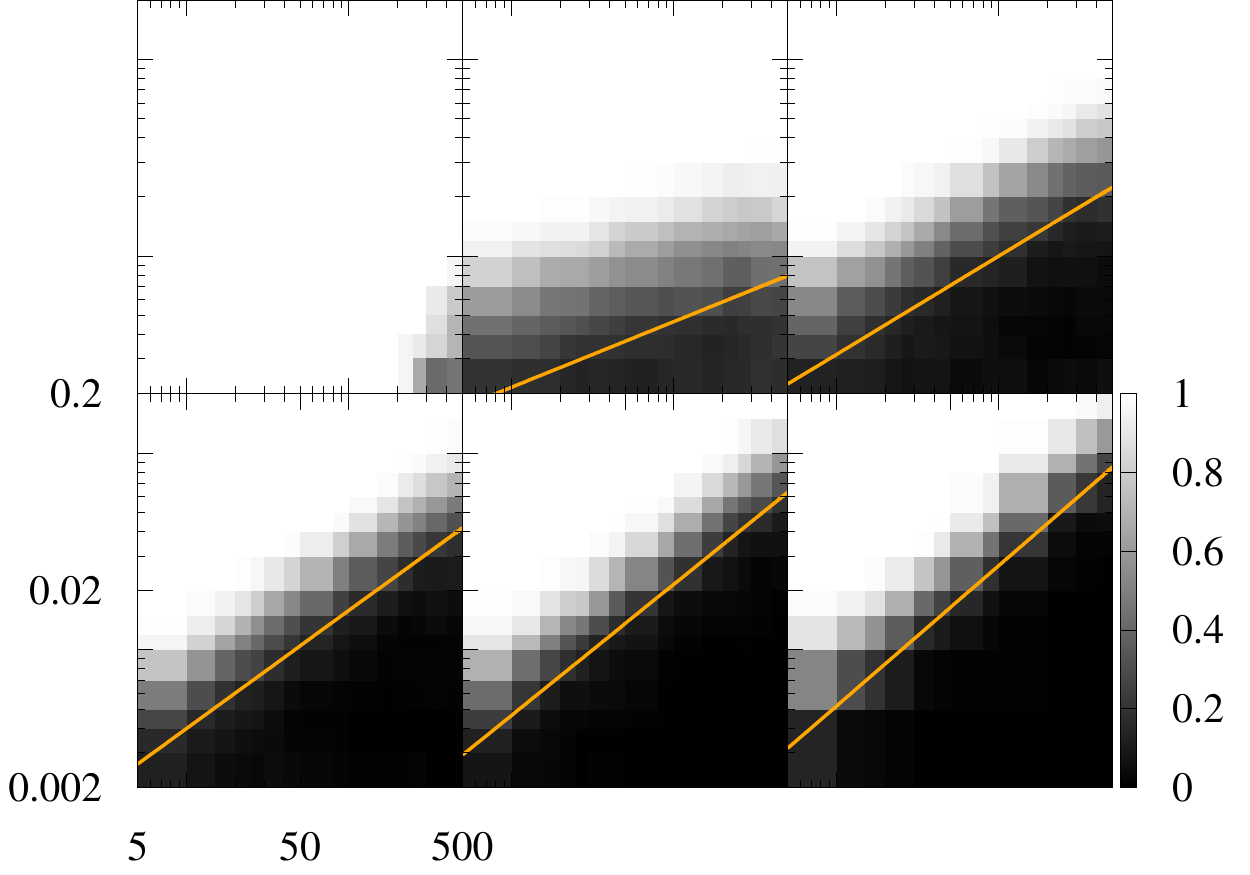} %
	\put(-170,-10){\small{rank, $r$}}
	\put(-220,27){\rotatebox{90}{\parbox{4cm}{\small{$\;$sampling probability, $p$}}}}
	\put(-180,138){\small{$k = 2$}}
	\put(-125,138){\small{$k = 3$}}
	\put(-70,138){\small{$k = 4$}}
	\put(-180,70){\small{$k = 5$}}
	\put(-125,70){\small{$k = 6$}}
	\put(-70,70){\small{$k = 7$}}
	\includegraphics[width=0.45\textwidth]{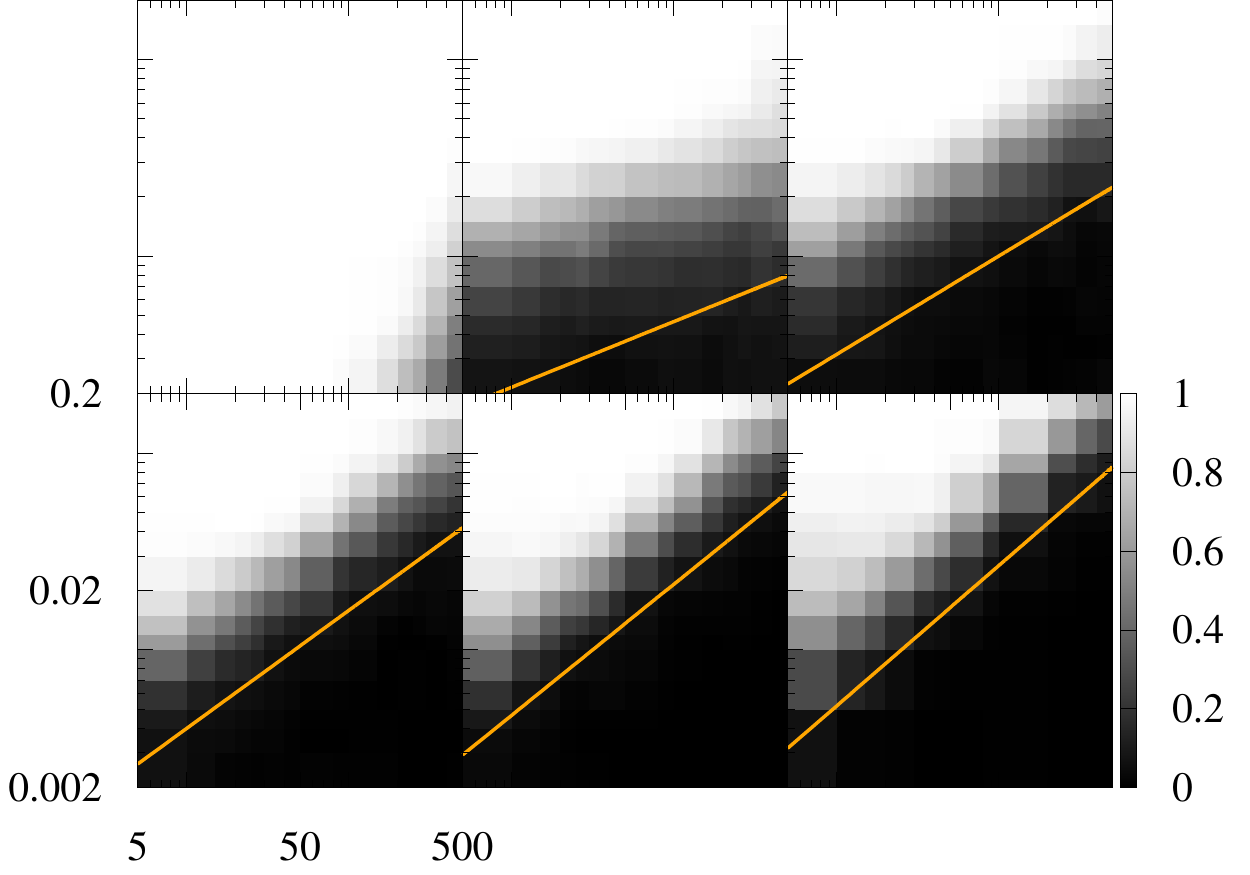} 
	\put(-180,138){\small{$k = 2$}}
	\put(-125,138){\small{$k = 3$}}
	\put(-70,138){\small{$k = 4$}}
	\put(-180,70){\small{$k = 5$}}
	\put(-125,70){\small{$k = 6$}}
	\put(-70,70){\small{$k = 7$}}
	\caption{Each colormap in each block for $k \in \{2,3,4,5,6,7\}$ show empirical probability of the event $\big\{\big|\|M\|_k^k - \widehat \Theta_k(\mathcal{P}_{\Omega}(M))\big|/\|M\|_k^k \leq \delta\big\}$, for $\delta = 0.5$ (left panel) and $\delta = 0.2$ (right panel). $\Omega$ is generated by Erd\"os-R\'enyi  sampling of matrix $M$ with probability $p$ (vertical axis). $M$ is a symmetric positive semi-definite matrix of size $d = 1000$. The solid lines correspond to  our theoretical prediction $p = (1/d)r^{1-2/k}$.}
	\label{fig:figER}    
\end{center}
\end{figure}

The sufficient condition of $d^2p = O(dr^{1-2/k})$  in Theorem \ref{thm:er_ub} 
holds for a broad range of parameters where 
the rank is sufficiently small $r=O(d^{k/((k-1)(k-2))})$ (to ensure that the first term in $\rho^2$ dominates).  
However, the following results in Figure \ref{fig:figER} on numerical experiments suggest that our analysis holds 
 more generally for all regimes of the rank $r$, even those close to $d$. 
 $M$ is generated using settings similar to that of Figure \ref{fig:compare}. Empirical probabilities are computed by averaging over $100$ instances.

One might hope to tighten the Chebyshev bound 
by exploiting the fact that the correlation among the summands in our estimator  \eqref{eq:estimate} is weak. 
This can be made precise using recent  result from \cite{schudy2011bernstein}, where 
a Bernstein-type bound was proved for sum of polynomials of independent random variables that  are weakly correlated. 
The first term in the bound \eqref{eq:ubChernoff1} is the natural Bernstein-type bound corresponding to the 
Chebyshev's bound in \eqref{eq:ub0}. 
However, under the regime where $k$ is large or $p$ is large, 
the correlation among the summands become stronger, and the second and third term in the bound \eqref{eq:ubChernoff1} starts to dominate. 
In the typical regime of interest where $\mu$, $\kappa$, $k$ are finite,  
$d^2p=O(dr^{1-2/k})$, and sufficiently small rank $r=O(d^{k/((k-1)(k-2))})$, 
the error probability is dominated by the first term in the right-hand side of $\eqref{eq:ubChernoff1}$. 
Neither one of the two bounds in \eqref{eq:ub0} and \eqref{eq:ubChernoff1} dominates the other, and depending on the 
values of the problem parameters, we might want to apply the one that is tighter. 
We provide a proof in 
Section \ref{sec:proof_er_ub2}. 

\begin{thm}
	Under the hypotheses of Theorem \ref{thm:er_ub}, 
        the error probability is upper bounded by 
\begin{eqnarray}\label{eq:ubChernoff1}
\P\Bigg( \frac{\big|\hTheta_k(\cP_\Omega(M)) - \norm{M}_k^k \big|}{\norm{M}_k^k}  \geq  \delta  \Bigg)  \leq 
	e^2 \max\Bigg\{ e^{- \frac{\delta^2}{\rho^2} \Big( \frac{dp}{r^{1-2/k}}\Big)^k},  e^{-(dp)\Big(\frac{\delta d}{ \rho r^{k-1}}\Big)^{1/k}},e^{-(dp)\Big(\frac{\delta d}{ \rho r^{k-1}}\Big) }, e^{-\frac{\delta dp}{ \rho}}\Bigg\}\,.
\end{eqnarray}
	\label{thm:er_ub2}
\end{thm}

These two results show that the sample size of $d^2p=O(dr^{1-2/k})$ is sufficient to 
estimate a Schatten $k$-norm accurately. 
In general, we do not expect to get a universal upper bound that is significantly tighter for all $r$, 
because for a special case of $r=d$, the following corollary of \cite[Theorem 3.2]{LNW14} provides a lower bound; 
it is necessary to have sample size $d^2p=O(d^{2-4/k})$ when $r=d$. Hence, the gap is at most a factor of 
$r^{2/k}$ in the sample complexity.

\begin{coro}
	\label{cor:er_lb}
	Consider any linear observation $X\in\reals^n$ of a matrix $M\in\reals^{d\times d}$ 
	and any estimate $\theta(X)$ satisfying $(1-\delta_k)\|M\|_k^k \leq \theta(X) \leq (1+\delta_k)\|M\|_k^k$ 
	for any $M$ with probability at least $3/4$, where $\delta_k=(1.2^k-1)/(1.2^k+1)$. Then, $n=\Omega(d^{2-4/k})$.
	
\end{coro}

For $k \in \{1,2\}$, precise bounds can be obtained with simpler analyses. In particular, we have the following remarks, whose proof follows 
immediately by applying Chebyshev's inequality and Bernstien's inequality along with the incoherence assumptions.

\begin{remark} \label{rem:rem3}
For $k=1$, the probability of error in \eqref{eq:ub0} is upper bounded by $\min \{\nu_1, \nu_2\}$, where 
$$\nu_1 \equiv \frac{1}{\delta^2}\frac{(\kappa\mu)^2}{dp} \text{\;, \;\; and \;\;\;\;\;}\nu_2 \equiv 2\exp\Big(\frac{-\delta^2}{2}\Big({\frac{(\kappa\mu)^2}{dp} + \delta\frac{(\kappa\mu)}{3 d p} }\Big)^{-1}\Big)\;.$$
\end{remark}

\begin{remark} \label{rem:rem2}
For $k=2$, the probability of error in \eqref{eq:ub0} is upper bounded by $\min \{\nu_1, \nu_2\}$, where 
$$\nu_1 \equiv \frac{1}{\delta^2}\frac{(\kappa\mu)^4}{d^2p}\big(2+\frac{r^2}{d}\big) \text{ \;, and \;\;\;\;} \nu_2 \equiv 2 \exp\Big(-\frac{\delta^2}{2}\Big({\frac{(\kappa\mu)^4}{d^2p}\big(2+\frac{r^2}{d}\big) + \delta\frac{(\kappa\mu)^2 r}{3 d^2 p} }\Big)^{-1}\Big)\;.$$
\end{remark}

When $k = 2$, for rank small $r \leq C \sqrt{d}$, only we only need $d^2p=O(1)$ samples for recovery up to any arbitrary small multiplicative error. When rank $r$ is large, our estimator requires  $d^2 p = O(d)$ for both $k \in \{1,2\}$.

\subsection{From Schatten norms to spectrum and spectral sum functions } 

Schatten norms by themselves are rarely of practical interest in real applications, 
but they provide a popular means to approximate 
 functions of singular values, which  are often of great practical interest \citep{DPS16,ZWJ15,KV16}. 
In this section, we consider two such applications using the first few Schatten norms explicitly: 
estimating the generalized rank in Section \ref{sec:genrank} 
and estimating the singular values in Section \ref{sec:hist}. 

\subsubsection{ Estimating the generalized rank }
	\label{sec:genrank}
	
For a matrix $M \in \reals^{d \times d}$ and a given constant $c \geq 0$, its 
{\em generalized rank} of order $c$ is given by
\begin{eqnarray}
{\rm rank}(M,c)  = \sum_{i = 1}^d \mathbb{I} \big[\sigma_i(M) > c \big]\,.
\end{eqnarray}
This  recovers the standard rank as a special case when $c=0$. 
Without loss of generality, we assume that $\sigma_{\max}(M)\leq 1$. 
For any given $0 \leq c_2 < c_1 \leq 1$, and $\delta \in [0,1)$, our goal is to get an estimate $\widehat r (\cP_\Omega(M))$ from sampled entries 
$\cP_\Omega(M)$ such that
\begin{eqnarray}
(1-\delta)\; {\rm rank}(M,c_1) \;\; \leq & \widehat r (\cP_\Omega(M)) & \leq \;\; (1+ \delta) \;{\rm rank}(M,c_2)\,.
\end{eqnarray}
The reason we take two different constants $c_1,c_2$ is to handle the ambiguous case when the matrix $M$ has many eigenvalues smaller but very close to $c_1$. If we were to set $c_2 = c_1$, then any estimator $\widehat r (M)$ would be strictly prohibited from counting these eigenvalues. However, since these eigenvalues are so close to the threshold, distinguishing them from other eigenvalues just above the threshold is difficult. Setting $c_2 < c_1$ allows us to avoid this difficulty and focus on the more fundamental challenges of the problem.

Consider the function $H_{c_1,c_2} : \reals \rightarrow [0,1]$ given by 
\begin{eqnarray}
H_{c_1,c_2}(x) = \begin{cases}
1 & \text{if } x > c_1\\
0 & \text{if } x < c_2\\
\frac{x-c_2}{c_1-c_2} & \text{otherwise.}
\end{cases} 
\end{eqnarray} 
It is a piecewise linear approximation of a step function and satisfies the following:
\begin{eqnarray}
	{\rm rank}(M,c_1) \;\; \leq & \sum_{i = 1}^d H_{c_1,c_2} (\sigma_i(M)) & \leq \;\; {\rm rank}(M,c_2)\,.
	\label{eq:genrankapproximation0}
\end{eqnarray}
We exploit this sandwich relation and estimate the generalized rank. 
Given a polynomial function $f : \reals \rightarrow \reals $ of finite degree $m$ 
such that $f (x)\approx H_{c_1,c_2}(x)$ for all $x$, such that  
$f(x) = a_0 + a_1x + \cdots + a_m x^{m}$, we immediately have the following relation, 
which extends to a function on the cone of PSD matrices in the standard way: 
\begin{eqnarray}
	\sum_{i = 1}^d f(\sigma_i(M)) & = & a_0 d + \sum_{k = 1}^m a_k\norm{M}_k^k\,.
	\label{eq:genrankapproximation}
\end{eqnarray}
Using this equality,  we propose 
the  estimator: 
\begin{eqnarray}\label{eq:grm}
\widehat r(\mathcal{P}_{\Omega}(M); c_1,c_2) & \equiv & a_0 d + \sum_{k = 1}^m a_k \hTheta_k(\mathcal{P}_{\Omega}(M))\,,
\end{eqnarray}
where we use the first several $\hTheta_k(\mathcal{P}_{\Omega}(M))$'s  obtained by the estimator \eqref{eq:estimate}. Note that function $f$ depends upon $c_1,c_2$. 
The remaining task is to obtain the coefficients of the polynomials in $f$ 
that is a suitable approximation of the function $H_{c_1,c_2}$. 
In a similar  context of estimating the generalized rank from approximate Schatten norms, 
\cite{ZWJ15} propose to use a composite function $f = q_s \circ q$ where $q$ is a finite-degree Chebyshev polynomial of the first kind such that $\sup_{x \in [0,1]} |q(x) - H_{c_1,c_2}(x) | \leq 0.1$, and $q_s$ is a polynomial of degree $2s+1$ given by 
\begin{eqnarray}
q_s(x) & = & \frac{1}{B(s+1,s+1)}\int_0^x t^s(1-t)^s dt\,, \qquad \text{where }B(\cdot,\cdot) \text{ is the Beta function.} 
\end{eqnarray}    
Note that, since $H_{c_1,c_2}$ is a continuous function with bounded variation, classical theory in \cite{mason2002chebyshev}, Theorem 5.7, guarantees existence of the Chebyshev polynomial $q$ of a finite constant degree, say $C_b$, that depends upon $c_1$ and $c_2$. 
Concretely, for a given choice of thresholds  $0\leq c_1<c_2 \leq 1$ and degree of the beta approximation $s$,  
the estimator $\widehat{r}(\cP_\Omega(M);c_1,c_2)$ in \eqref{eq:grm} can be computed as follows. 

\begin{algorithm}[ht]
\caption{Generalized rank estimator (a variation of \cite{ZWJ15})}
\label{algo:genrank}
  \begin{algorithmic}[1]
    \REQUIRE $\cP_\Omega(M)$, $c_1$, $c_2$, $s$  
	\ENSURE $\widehat{r}(\cP_\Omega(M);c_1,c_2)$    
    \STATE For given $c_1$ and $c_2$, find a Chebyshev polynomial of the first kind $q(x)$ satisfying \hfill [Appendix \ref{sec:cheby}]
    $$\sup_{x\in[0,1]} |q(x)-H_{c_1,c_2}(x)| <0.1$$
        \STATE Let  $C_b$ denote the degree of $q(x)$ 
    	\STATE Find the degree $(2s+1)C_b$ polynomial expansion of $q_s \circ q (x) = \sum_{k=0}^{(2s+1)C_b} a_k x^{k} $
	\STATE $\widehat{r}(\cP_\Omega(M);c_1,c_2)  \leftarrow  a_0 d + \sum_{k = 1}^{(2s+1)C_b} a_k \hTheta_k(\cP_\Omega(M)) $ 
	\hfill [Algorithm \ref{algo:est}]
  \end{algorithmic}
\end{algorithm}

The approximation of $H_{c_1,c_2}$ with $f = q_s \circ q$ and our upper bound on estimated Schatten norms $ \hTheta_k(\mathcal{P}_{\Omega}(M))$ translate into the following guarantee on generalized rank estimator $\widehat r(\mathcal{P}_{\Omega}(M); c_1,c_2)$ given in \eqref{eq:grm}.
\begin{coro}
Suppose $\norm{M}_2 \leq 1$. Under the hypotheses of Theorem \ref{thm:er_ub}, for any given $ 1 \geq c_1 > c_2 \geq 0 $, there exists a constant $C_b$, such that for any $s \geq 0$ and any $\gamma > 0$, the estimate in \eqref{eq:grm} with the choice of $f = q_s \circ q$ satisfies
\begin{eqnarray}
(1-\delta) ( {\rm rank}(M,c_1) - 2^{-s} d) \;\;\leq & \widehat r(\mathcal{P}_{\Omega}(M); c_1,c_2) & \leq \;\; (1+\delta)( {\rm rank}(M,c_2) + 2^{-s} d)\,,
\end{eqnarray} 
with probability at least $1 - \gamma C_b (2s+1)$, where $\delta \equiv  \max_{ 1 \leq k \leq C_b(2s+1)} \Big \{\sqrt{\frac{\rho^2}{\gamma} (\frac{\max\{1,r^{1-2/k}\}}{dp})^k} \Big\}$.
\label{coro:genrank}
\end{coro}
The proof follows immediately using Theorem \ref{thm:er_ub} and the following lemma which gives a uniform bound on the approximation error between $H_{c_1,c_2}$ and $f = q_s \circ q$. 
Lemma \ref{lem:genrank}, together with 
 Equations.~\eqref{eq:genrankapproximation0} and \eqref{eq:genrankapproximation}, 
provides a (deterministic) functional approximation guarantee of 
\begin{eqnarray}
	 {\rm rank}(M,c_1) - d \, 2^{-s}  \;\; \leq \;\;    \sum_{i = 1}^d f (\sigma_i(M))     \;\; \leq \;\; {\rm rank}(M,c_1) + d \, 2^{-s} \;,
\end{eqnarray}
for any $c_1<c_2$ and any choice of $s$, as long as $C_b$ is large enough to guarantee $0.1$ uniform error bound on 
the Chebyshev 
polynomial approximation.
Since we can achieve $1\pm\delta$ approximation on each polynomial in $f(\sigma_i(x))$, 
Theorem \ref{thm:er_ub} implies the desired Corollary  \ref{coro:genrank}. 
Note that using Remarks \ref{rem:rem3} and \ref{rem:rem2}, the bounds in \eqref{eq:ubChernoff1} hold for $k \in [1,\infty)$ with $r^{1-2/k}$ replaced by $\max\{1, r^{1-2/k}\}$.

\begin{figure}[h]
 \begin{center}
    \includegraphics[width=0.42\textwidth]{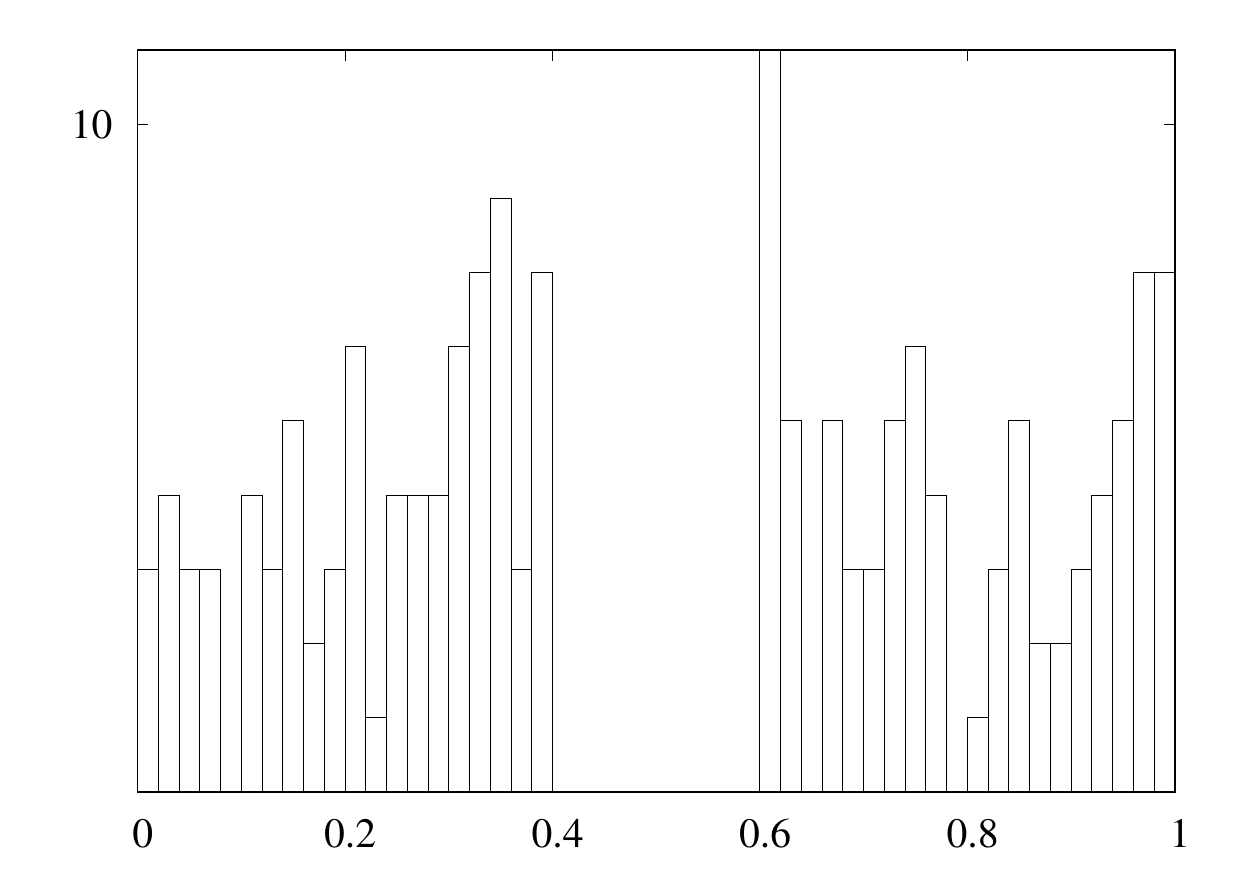} 
	\put(-120,-7){{\small singular values}}
    \put(-195,57){\rotatebox{90}{\parbox{4cm}{\small{count}}}}
    \hspace{2em}
    \includegraphics[width=0.42\textwidth]{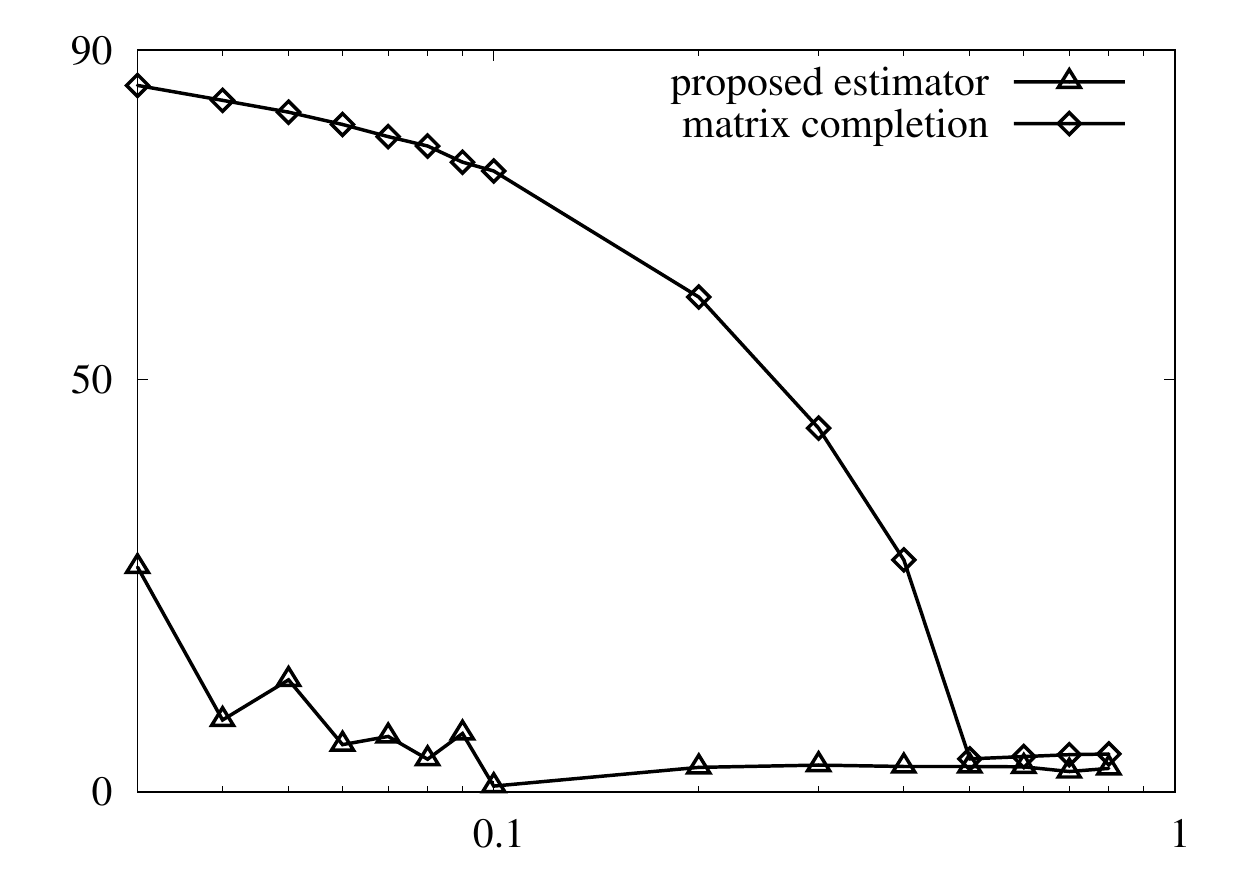} 
	\put(-140,-7){{\small sampling probability, $p$}}
    \put(-200,37){\rotatebox{90}{\parbox{4cm}{\small $|\widehat{r}(c_1,c_2) - r(c_1)|$}}}	
\caption{The left panel shows a histogram  of singular values of  $M$ chosen for the experiment. 
	The right panel compares absolute error in estimation $\widehat r(\mathcal{P}_{\Omega}(M); c_1=0.5,c_2=0.6)$ for two choices of the Schatten norm estimates $\widehat {\|M\|_k^k}$: first the proposed estimator $\hTheta_k(\mathcal{P}_{\Omega}(M))$ in \eqref{eq:estimate}, and second the Schatten norm of the completed matrix, $\widetilde M = {\rm AltMin}(\cP_{\Omega}(M))$ from \citep{JNS13}.}
	\label{fig:figGeneralizedRank}   
\end{center}	
\end{figure}

\begin{lemma}[\cite{ZWJ15}, Lemma 1]
\label{lem:genrank}
Consider the composite polynomial $f(x) = q_s(q(x))$. Then $f(x) \in [0,1]$ for all $x \in [0,1]$, and moreover
\begin{eqnarray}
|f(x) - H_{c_1,c_2}(x)| \leq 2^{-s}\,, \qquad \text{for all } x \in [0,c_2] \cup [c_1,1]\,.
\end{eqnarray}  
\end{lemma}
In Figure \ref{fig:figGeneralizedRank}, we evaluate the performance of estimator \eqref{eq:grm} numerically. We construct a symmetric matrix $M$ of size $d = 1000$ and rank $r = 200$. $\sigma_{i} \sim {\rm Uni}(0,0.4)$ for $ 1\leq i \leq r/2$, and $\sigma_{i} \sim {\rm Uni}(0.6,1)$ for $ r/2+1 \leq i \leq r$. We estimate $\widehat r(\mathcal{P}_{\Omega}(M); c_1,c_2)$ for Erd\"os-R\'enyi sampling $\Omega$, and a choice of $c_2 = 0.5$ and $c_1 = 0.6$, which is motivated by the distribution of $\sigma_i$. We use Chebyshev polynomial of degree $C_b = 2$, and $s = 1$ for $q_s$. That is function $f$ is of degree 6. Accuracy of the estimator can be improved by increasing $C_b$ and $s$, however that would require estimating higher Schatten norms.

 
 %
\subsubsection{Estimating the spectrum} 

	\label{sec:hist}

Given accurate estimates of first $K$ Schatten norms of a matrix $M$, we can estimate singular values of $M$ using a linear programming based algorithm given in \citep{KV16}. In particular, we get the following guarantees on the estimated singular values, 
whose proof follows directly using the analysis techniques  in the proof of \citep[Theorem 2]{KV16}. 
The main idea is that given the rank, the maximum support size of the true spectrum, and an estimate of its first $K$ moments, 
one can find $r$ singular values whose 
$K$ first moments are close to the estimated Schatten norms. 

\begin{algorithm}[ht]
\caption{Spectrum estimator (a variation of \cite{KV16})}
\label{algo:spectrum}
  \begin{algorithmic}[1]
    \REQUIRE $\cP_\Omega(M)$, $K$, $\epsilon$, target rank $r$,  lower bound $a$ and upper bound $b$ on the positive singular values
	\ENSURE estimated singular values $(\widehat{\sigma}_1,\widehat{\sigma}_2,\ldots,\widehat{\sigma}_r)$   
    \STATE  $L \in \reals^K : L_k =\hTheta_k(\cP_\Omega(M))$ for $k \in [K]$ 
	\hfill [Algorithm \ref{algo:est}]
	\STATE $t = \lceil (b-a)/\epsilon \rceil + 1$, $x \in \reals^t$: $x_i = a + \epsilon (i-1)$, for $i \in [t]$, 
	\STATE $V \in \reals^{K\times t}: V_{ij} = x_j^i$ for $i \in [K], j \in [t]$
\STATE $p^* \equiv  \{\min_{p \in \reals^t} |Vp - L|_1 : \ones_t^\top p = 1, p \geq 0\}$
\STATE $\widehat\sigma_i = \min\{x_j: \sum_{\ell \leq j }p^*_{\ell} \geq \frac{i}{r+1}\}$, $i$th $(r+1)$st-quantile of distribution corresponding to $p^*$
	  \end{algorithmic}
\end{algorithm}

Further, our upper bound on the first $K$ moments can be translated into 
an upper bound on the Wasserstein distance between those two distributions, which in turn gives 
the following bound on the singular values. 
With small enough $\epsilon$ and large enough $K$ and $r$, we need sample size 
$d^2p > C_{r,K,\epsilon,\gamma} d r^{1-2/k}$ to achieve arbitrary small error. 

\begin{coro}
Under the hypotheses of Theorem \ref{thm:er_ub}, given rank $r$, constants $ 0\leq a < b $ such that $\sigma_{\min} \geq a$, $\sigma_{\max} \leq b$, and estimates of the first $K$ Schatten norms of $M$, $\{\hTheta_k(\mathcal{P}_{\Omega}(M))\}_{k \in [K]}$ obtained by the estimator \eqref{eq:estimate}, for any $0 < \epsilon \ll (b-a)$, and $\gamma > 0$, Algorithm \ref{algo:spectrum} 
 runs in time ${\rm poly}(r,K,(b-a)/\epsilon)$ and returns  $\{\widehat \sigma_i\}_{i \in [r]}$ an estimate of $\{\sigma_i(M)\}_{i \in [r]}$ such that
\begin{eqnarray} \label{eq:spec_est}
\frac{1}{r} \;\sum_{i = 1}^r |\widehat \sigma_i - \sigma_i| & \leq & \frac{C(b-a)}{K} + \frac{b-a}{r} + g(K)(b-a)\Bigg(\epsilon K b^{K-1} + \sum_{k = 1}^K \sigma_{\max}^k\sqrt{\frac{\rho^2}{\gamma} \bigg(\frac{\max\{1, r^{1-2/k}\}}{dp}\bigg)^k}\Bigg)\,,\nonumber\\
\end{eqnarray} 
with probability at least $1 - \gamma K$, where $C$ is an absolute constant and $g(K)$ only depends on $K$.
\end{coro}

\begin{figure}[h]
 \begin{center}
    \includegraphics[width=0.48\textwidth]{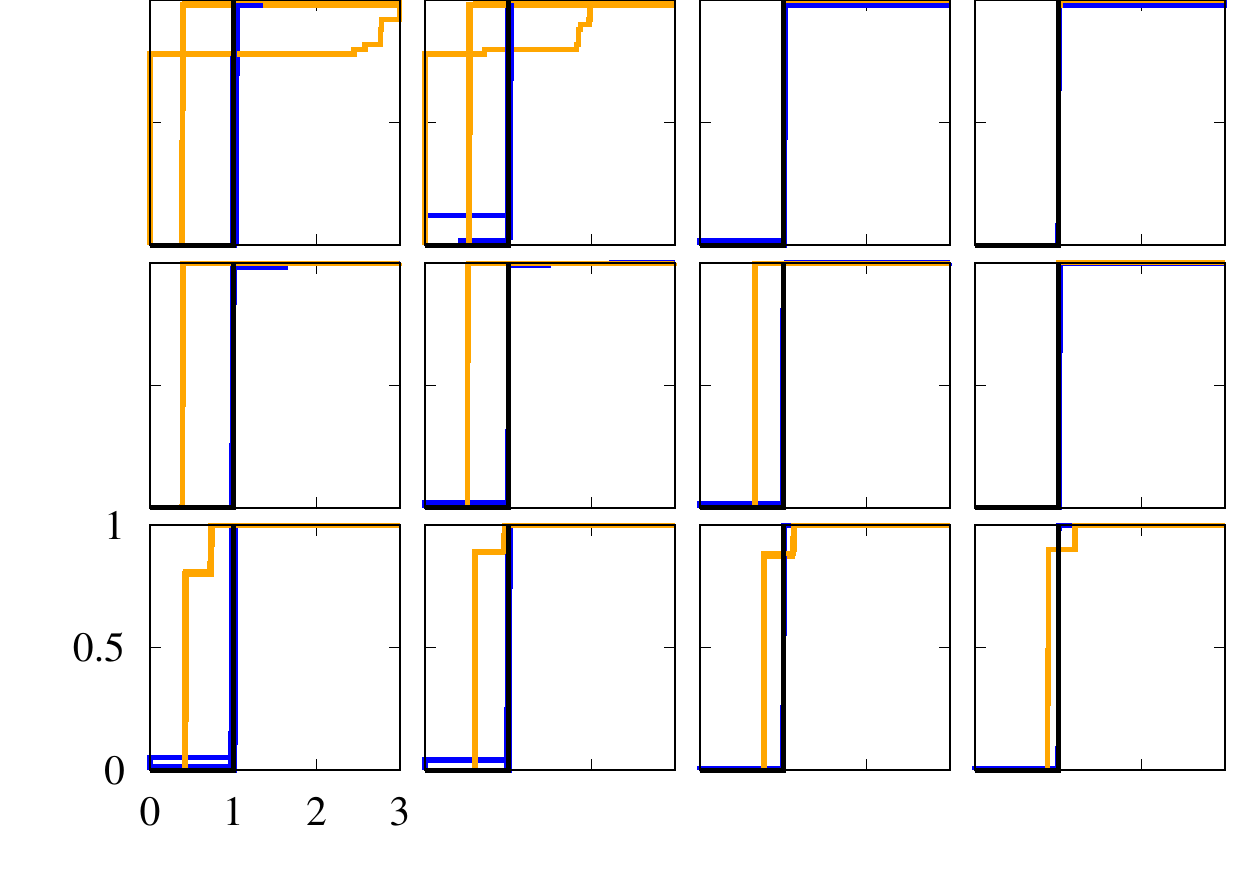} 
	\put(-202,1){{\small singular values}}
	\put(-188,162){{\small $\alpha = 3$}}
	\put(-138,162){{\small $\alpha = 5$}}
	\put(-88,162){{\small $\alpha = 8$}}
	\put(-42,162){{\small $\alpha = 10$}}	
   \put(-222,122){\rotatebox{90}{\parbox{4cm}{\small{$r = 50$}}}}     
   \put(-222,72){\rotatebox{90}{\parbox{4cm}{\small{$r = 200$}}}} 
   \put(-222,22){\rotatebox{90}{\parbox{4cm}{\small{$r = 500$}}}} 
    \includegraphics[width=0.48\textwidth]{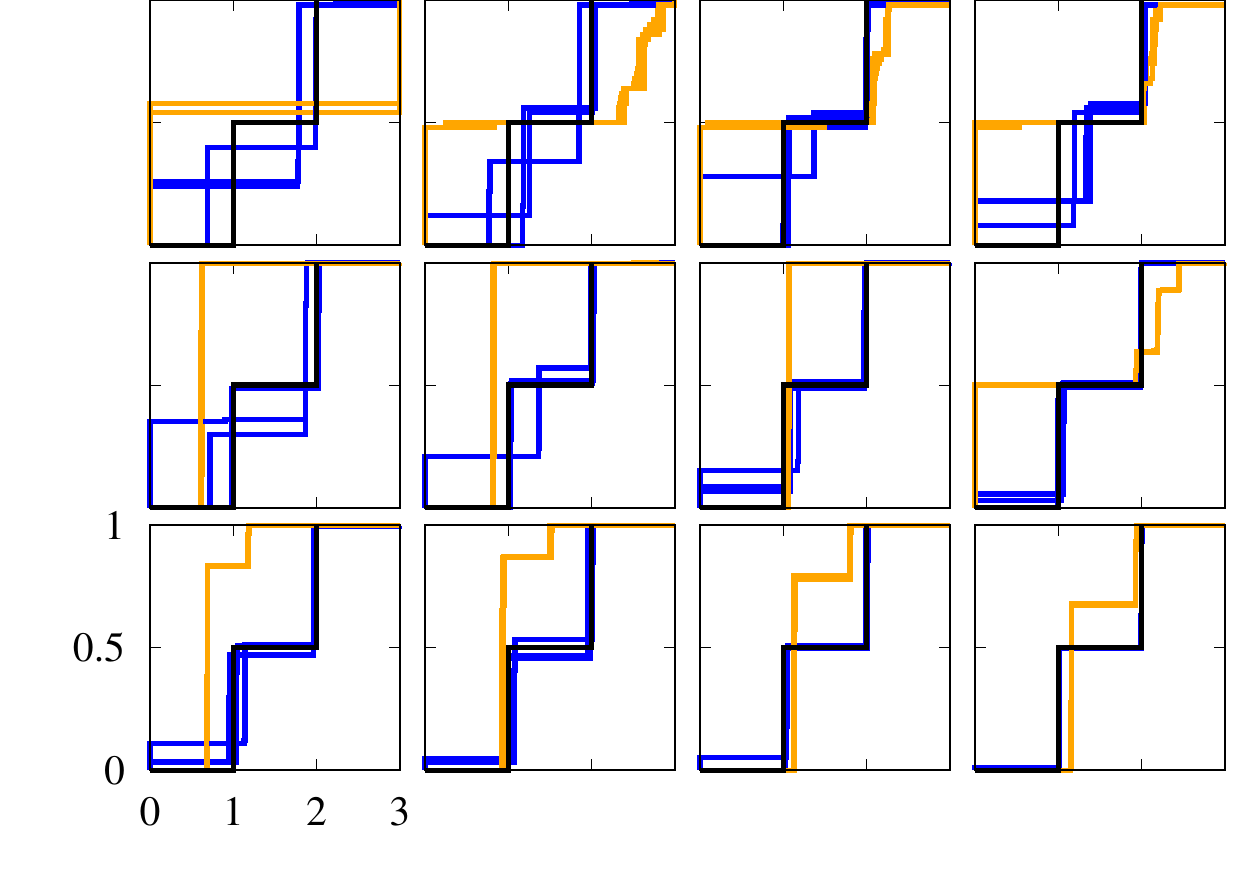}
    \put(-202,1){{\small singular values}}   
	\put(-188,162){{\small $\alpha = 3$}}
	\put(-138,162){{\small $\alpha = 5$}}
	\put(-88,162){{\small $\alpha = 8$}}
	\put(-42,162){{\small $\alpha = 10$}}	
	\caption{ The proposed estimator (in blue solid lines) outperforms matrix completion approaches (in orange solid lines) in estimating the ground truths empirical cumulative distribution function of the $r$ strictly positive singular values (in black solid line) for two examples: 
	one peak at $\sigma_i=1$ on the left and two peaks at $\sigma_i=1$ or $\sigma_i=2$ on the right.
	Both approaches achieve  better accuracy as sample size increases with $\alpha$, where $p = (\alpha/d) r^{1-2/7}$. 
	}
	\label{fig:figSP}   
\end{center}	
\end{figure}

In Figure \ref{fig:figSP}, we evaluate the performance of the proposed estimator \eqref{eq:estimate}, in recovering the true spectrum using 
Algorithm \ref{algo:spectrum}. We compare the results with the case when Schatten norms are estimated using matrix completion. We consider two distributions on singular values, one peak and two peaks. More general distributions of spectrum can be recovered accurately, however that would require estimating higher Schatten norms. 
For both cases, the proposed estimator outperforms matrix completion approaches, and 
achieves better accuracy as sample size increases with $\alpha$. 
In each graph, the black solid line depicts the empirical Cumulative Distribution Function (CDF) of the ground truths $\{\sigma_i\}_{i \in [r]}$ 
for those $r$ strictly positive singular values. On the left, there are $r$ singular values at one peak $\sigma_i=1$, and 
on the right there are $r/2$ singular values at each of the two peaks at $\sigma_i=1$ and $\sigma_i=2$.
Each blue line and the orange line depicts the empirical CDF of $\{\widehat \sigma_i\}_{i \in [d]}$ and $\{\widetilde \sigma_i\}_{i \in [d]}$ respectively for each trial, over three independent trials. $\widehat \sigma_i$'s are estimated using $\{\hTheta_k(\mathcal{P}_{\Omega}(M))\}_{k \in [K]}$ obtained by the estimator \eqref{eq:estimate}, and $\widetilde \sigma_i$'s are estimated using $\{\norm{\widetilde M}_k^k\}_{k \in [K]}$ where $\widetilde M = {\rm AltMin}(\cP_{\Omega}(M))$, along with Algorithm 2 in \citep{KV16}, for $K = 7$. $M$ is a symmetric matrix of size $d = 1000$ and rank $r \in \{50,200,500\}$ with singular values $\{\sigma_i\}_{i \in [d]}$. $\Omega$ is generated using  Erd\"os-R\'enyi sampling with probability $p = (\alpha/d) r^{1-2/7}$ for $\alpha \in \{3,5,8,10\}$.

\section{Graph sampling} 
\label{sec:gs}

Our framework for estimating the Schatten $k$-norm 
can be applied more generally to any random sampling, 
as long as the distribution is permutation invariant.  
In practice, we typically observe one instance of a sampled matrix 
and do not know how the samples were generated. 
Under a mild assumption that the probability of sampling an entry is independent of the value of that entry, 
the only information about the sampling model that we have is the {\em pattern}, 
i.e. an unlabelled graph $G=(V,E)$ capturing the pattern of sampled indices by the edges. 
This naturally suggests a  novel sampling scenario  
that we call {\em graph sampling} defined in Section \ref{sec:introsetup}. 
In this section, we provide an  estimator under graph sampling, 
and characterize the fundamental limit on the achievable error. 
This crucially depends on the original pattern $G$ via 
a  fundamental property $\lambda^*_{G,r}$, 
which is generally challenging to compute. 
However, we provide a  bound on $\lambda_{G,r}^*$ 
for two extreme cases of varying difficulty: 
a clique sampling that requires only $O(r^{2-4/k})$ samples 
and a clique-star sampling that requires as many as $O(dr^{1-4/k})$ samples. 
This is made formal by showing a  lower  bound on the minimax sample complexity. 
Comparing the two necessary conditions on sample complexity, 
$O(r^{2-4/k})$ for clique sampling and 
$O(dr^{1-4/k})$ for clique-star sampling, 
it follows that 
depending on the pattern of the samples, the sample complexity can vary drastically, especially for low-rank matrices where $r\ll d$.

Under the {\em graph sampling}, 
the probability $p(H)$ that we observe all edges in a walk with pattern $H$ is 
\begin{eqnarray}
	p(H) \;\; = \;\; \frac{\wgt_{\cP_\Omega(\ones_d \ones_d^T)}(H)}{ \wgt_{\ones_d \ones_d^T}(H)}  \;,
\end{eqnarray}
where $\ones_d \ones_d^T$ is the all ones matrix, and by permutation invariance, the probability is 
the ratio between total (unweighted) number of walks with $H(w)=H$ in the original pattern $\Omega$ and 
that of the complete graph $K_d$.
Note that although $\Omega$ is a random quantity, 
$\wgt_{\cP_\Omega(\ones \ones^T)}(H)$ only depends on the structure and not the labelling of the nodes and hence is a deterministic quantity. 
Plugging in this value of $p(H)$, which can be computed in time  $O(d^\alpha)$ for $k\leq 7$ 
as shown in Proposition \ref{pro:algorithm}
(and in general only increases the computational complexity of the estimate by a factor of two), 
into the estimator \eqref{eq:estimate}, we get an estimate customized for graph sampling. 

\subsection{Performance Guarantees}
\label{subsec:lb}

Recall the graph sampling defined in Section \ref{sec:introsetup}, 
where we relabel the nodes of a pattern graph $G(V,E)$ according to a random uniform permutation, 
and sample the entries of the matrix $M$ on the edges. 
We prove a fundamental lower bound on the sample complexity that crucially depends on 
the following property of the pattern $G$. Let $G_{\pi}(\widetilde V, \Omega)$ denote the graph after relabeling the nodes of $G = (V,E)$ with permutation $\pi: [d] \rightarrow [d]$.  
For independent Rademacher variables $u_i$ for $i\in[r]$
\begin{eqnarray}
	\label{eq:fgrlambda} 
	 f_{G,r}(\lambda) \; \equiv \; \max_{\pi}\Bigg\{\E_{u} \bigg[ \exp\bigg((5/d)^2\lambda^2 \sum_{(i,j) \in \cP^{(r)}(G_{\pi}) }  u_i u_j \bigg) \bigg]\Bigg\} \;, 
\end{eqnarray}
where $\cP^{(r)}(G_{\pi}) \subseteq [r]\times [r]$ 
is a projection of the edges $\Omega$ over $d$ nodes to a set of edges over $r$ nodes by 
mapping a node $i\in[d]$ to a node $1+(i-1  \text{ mod } r) \in[r]$. 
Precisely, $(i,j) \in \cP^{(r)}(G_{\pi})$ if there exists an edge $(i',j') \in \Omega$ such that $i=1+(i'-1  \text{ mod } r)$ and $j=1+(j'-1  \text{ mod } r)$. Observe that $f_{G,r}(\lambda)$ is a non-decreasing function of $\lambda$. It follows from the fact that for any positive $\lambda$ and random variable $x$ and any $\epsilon > 0$, we have $\E[e^{\lambda(1+\epsilon)x}] \geq \E[e^{\lambda x}](\E[e^{\lambda x}])^\epsilon \geq  \E[e^{\lambda x}]e^{\epsilon\lambda\E[x]} \geq  \E[e^{\lambda x}]$. The first and the second  inequalities use Jensen's inequality and the third one holds when $\E[x] \geq 0$. Note that $\E_{u}[\sum_{(i,j) \in \cP^{(r)}(G_{\pi}) }  u_i u_j ] \geq 0$, since $u_i$'s are i.i.d. Rademacher variables.

This function measures the distance between a particular low-rank matrix with Gaussian entries and its rank one perturbation, 
which is used in our constructive lower bound (see Eq.~\eqref{eq:lba1}). 
Intuitively, smaller $f_{G,r}(\lambda)$ implies that two rank-$r$ 
matrices with separated Schatten norms look similar after graph sampling w.r.t. $G$. 
Hence, when this function is small, say less than $26/25$, then 
it is hard to distinguish which of the two (distributions of) matrices we are observing. 
This is captured by  the largest value of $\lambda$ that still maintains 
$f_{G,r}(\lambda)$ sufficiently small:  
\begin{eqnarray}
	\label{eq:deflambda}
	\lambda_{G,r}^* \;\; \equiv\;\;   \max_{\{\lambda>0:f_{G,r}(\lambda) \leq 26/25\}}  \lambda   \;.
\end{eqnarray}
One can choose any number not necessarily  $26/25$ as long as it is strictly larger than one and strictly smaller than two,
 and this will only 
change the probability upper bound in \eqref{eq:lb1}. 
If we sample from a graph $G$ with large $\lambda_{G,r}^*$, 
then we cannot distinguish two distributions even if they have a large Schatten norm separation. 
We do not have enough samples and/or our pattern is not sample efficient. 
The dependence of the fundamental lower bound  on the graph $G$ is captured by this property $\lambda_{G,r}^*$, which is made precise in the following theorem. 
We provide a lower bound 
that captures how sample complexity depends on the pattern $G$ and also on  the underlying matrix, 
by providing analysis customized for each family of matrices $\cM_{r,\mu}$ 
parametrized by its rank and incoherence: 
	$$\cM_{r,\mu} \equiv \big\{ M \in \reals^{d \times d}: M  = M^\top, 
	{\rm rank}(M) \leq r\,,\, \mu(M) \leq \mu\big\}.$$

\begin{thm}[General lower bound under graph sampling]
	\label{thm:lb}
	For any finite $k \in [3,\infty)$ 
	suppose we observe samples 
	under  the graph sampling defined in Section \ref{sec:introsetup} with respect to a pattern graph $G = (V,E)$.  
	 Then there exist  universal constants $C>0$, $C'>0$ and $C'' > 0$   such that 
	 for any $r \geq  e^{C''k}$ and $\mu \geq C' \sqrt{\log r}$,  
	  if  $\lambda_{G,r}^* \geq  C d r^{1/k-1/2}$ then
	\begin{eqnarray}
	\label{eq:lb1} 
	 \inf_{M \in \cM_{r,\mu}} \; \sup_{\widetilde\Theta} \;\;   \mathbb{P}\bigg( \frac{1}{2} \norm{M}_k \leq \widetilde \Theta(\cP_{\Omega(M)}) \leq 2 \norm{M}_k \bigg) & \leq & \frac{3}{4} \,,
	\end{eqnarray} 
	where the supremum is over any measurable function of  $\cP_{\Omega(M)}$ and 
	the probability is with respect to the random sampling $\Omega$.	
\end{thm}
A proof of Theorem \ref{thm:lb} is given in Section \ref{sec:proof_lb}. It is in general challenging to evaluate $\lambda^*_{G,r}$ for a given graph. 
For a special case of {\em clique sampling} where the pattern $G(V,E)$ is a clique over a subset of $\ell$ nodes among $d$, 
we provide a sharp upper bound on $\lambda^*_{G,r}$. 
\begin{lemma}[Lower bound for clique sampling]
	\label{lem:lb}
	If the pattern graph $G(V,E)$ is a clique over a subset of $\ell$ nodes, 
	then $\lambda_{G,r}^* \leq 2^{-4}d{(\min\{\ell,r\})}^{-1/2}$. 
\end{lemma}
Together with Theorem \ref{thm:lb}, this implies that if $\ell\leq 2^{-8}C^{-2} r^{1-2/k}$ 
(such that $\lambda_{G,r}^* \geq  C d r^{1/k-1/2}$), then 
 with probability at least $1/4$ any estimator makes an multiplicative error larger than two.
 Hence, sample size of $\ell(\ell+1)/2= O( r^{2-4/k})$ is necessary to achieve multiplicative error of two with high probability. 
We show that our estimator is optimal, by providing a matching upper bound on the sample complexity   when $k=3$. 
For any positive semidefinite matrix $M\in\reals^{d\times d}$ of rank $r$ with 
incoherence  $\mu(M)=\mu$,  $\kappa=\sigma_{\rm max}(M)/\sigma_{\rm min}(M)$, 
and some function $g(k)=O(k!)$, 
we define
\begin{eqnarray*}
	\tilde{\rho}^2 \;\;\equiv\;\;  (\kappa \mu)^{2k}g(k)  \max \Bigg\{1\,,\, \frac{\ell^{k-1}}{r^{k-2}} \,,\, \frac{\ell}{r} \,,\, \frac{r^{1/2}\ell^k  }{d} \Bigg\}\,,
\end{eqnarray*}
such that the variance of our estimator is bounded by 
${\rm Var}( \hTheta(\cP_\Omega(M))/\|M\|_k^k ) \leq \rho^2 (r^{1-2/k}/\ell)^k$ as we show for $k=3$ 
in the proof of Theorem \ref{thm:graph_ub} 
in Section \ref{sec:proof_graph_ub}. 
Here, $g(k)=O(k!)$ is a function of $k$ only.

\begin{thm}[Upper bound for clique sampling]\label{thm:graph_ub}
For $k =3$, any $\delta >0$, and any rank-$r$  matrix $M\succeq 0$, 
the proposed estimator \eqref{eq:estimate} achieves a multiplicative error $\delta$ 
with probability of error bounded by
\begin{eqnarray}\label{eq:ubg1}
	\P\Bigg( \frac{\big|\hTheta_k(\cP_\Omega(M)) - \norm{M}_k^k \big|}{\norm{M}_k^k} & \geq & \delta  \Bigg) \;\; \leq \;\; 
	\frac{\tilde{\rho}^{2}}{\delta^2 }  \Big( \frac{r^{1-2/k}}{\ell}\Big)^k \,,
\end{eqnarray}
under the graph sampling with the pattern graph $G$ that is a clique over $\ell$ nodes.
\end{thm}

\begin{figure}[h]
 \begin{center}
	\includegraphics[width=0.45\textwidth]{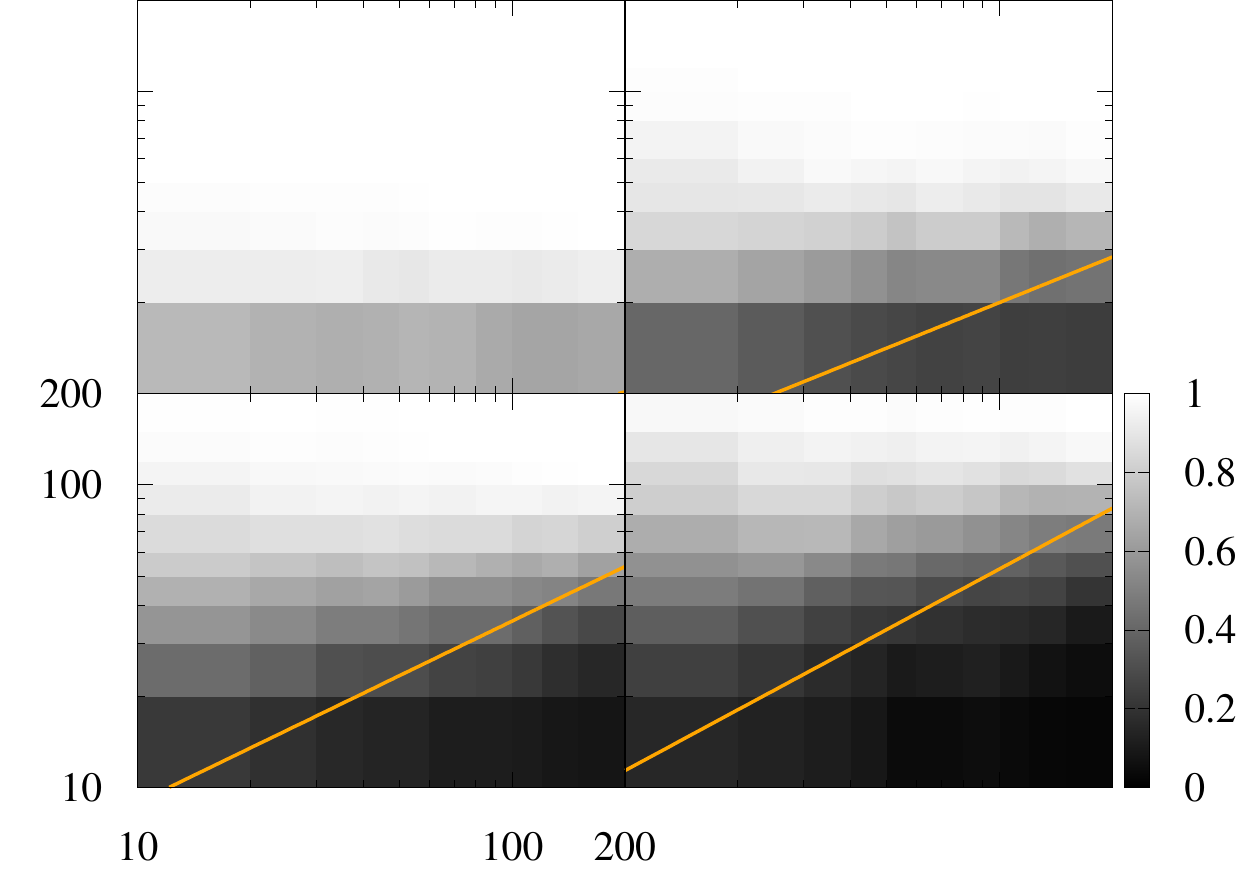} 
	\put(-170,-3){\small{rank $r$}}
	\put(-216,20){\rotatebox{90}{\parbox{4cm}{\small{$\;$clique size $\ell$}}}}
	\put(-182,138){\small{$k = 3$}}
	\put(-100,138){\small{$k = 4$}}
	\put(-182,70){\small{$k = 5$}}
	\put(-100,70){\small{$k = 6$}}
	\includegraphics[width=0.45\textwidth]{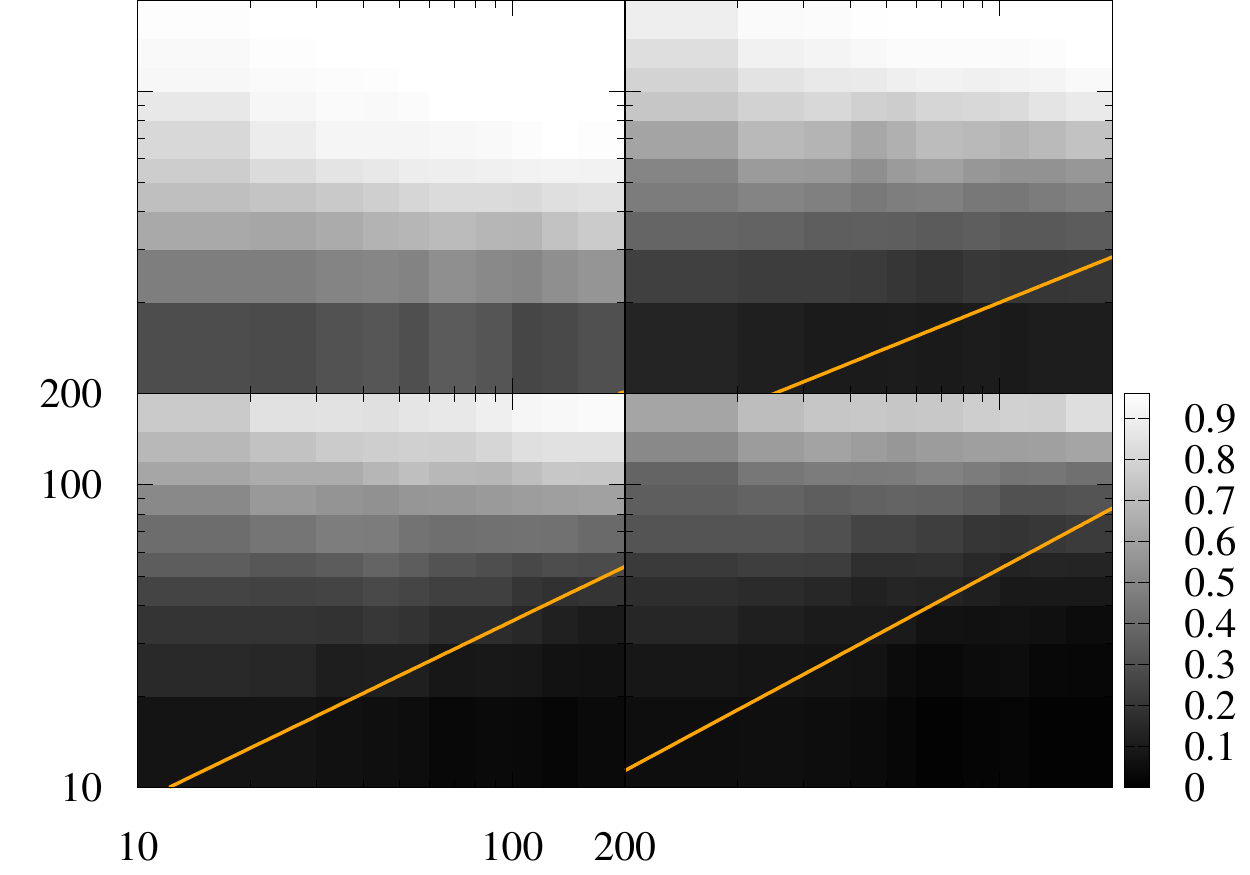} 	
	\put(-182,138){\small{$k = 3$}}
	\put(-100,138){\small{$k = 4$}}
	\put(-182,70){\small{$k = 5$}}
	\put(-100,70){\small{$k = 6$}}
	\caption{Each colormap in each block for $k \in \{3,4,5,6\}$ show empirical probability of the event $\big\{\big|\|M\|_k^k - \widehat \Theta_k(\mathcal{P}_{\Omega}(M))\big|/\|M\|_k^k \leq \delta\big\}$, for $\delta = 0.5$ (left panel) and $\delta = 0.2$ (right panel). 
		$\Omega$ is generated by clique sampling of matrix $M$ with a clique of size $\ell$ (vertical axis). 
		$M$ is a  positive semi-definite matrix of size $d = 1000$. The solid lines correspond to our theoretical prediction $\ell = \sqrt{k}r^{1-2/k}$.}
	\label{fig:figClique}   
\end{center}	
\end{figure}

For a typical scenario with finite $\mu$ and  $\kappa$, this upper bound shows that 
sample size of $\ell(\ell+1)/2=O(r^{2-4/k})$ is sufficient to achieve any arbitrarily small multiplicative error 
for $k=3$ and sufficiently small rank $r\leq d^{2k/(3k-2)}$ and  $\ell \leq r^{(k-2)/(k-1)}$,  to ensure that the first term dominates in $\tilde{\rho}^2$. However, the numerical experiments suggest that our analysis holds more generally for all regimes of the rank $r$. 
This matches the previous lower bound, proving optimality of the proposed estimator. 
Although the current analysis holds only for $k=3$, 
we are intentionally writing the guarantee in general form as we 
 expect the bound to hold more generally. In particular, we believe that Lemma \ref{conj:conj1} holds for all $k \geq 3$, and thereby Theorem \ref{thm:graph_ub} holds for any fixed integer $k \in [3,\infty).$
In the  numerical experiments in Figure \ref{fig:figClique}, $M$ is generated using settings similar to that of Figure \ref{fig:compare}. Empirical probabilities are computed by averaging over $100$ instances.

Although our analysis does not give a tight lower bound for Erd\"os-R\'enyi sampling,  
there exists graph patterns such that sample complexity is large, i.e.~scales linearly in $d$. 
Consider a {\em clique-star sampling} where 
the pattern graph $G(V,E)$ has a clique on a small subset of nodes $V_1$, $|V_1| = \ell$, and the remaining nodes $V \setminus V_1$ are disconnected among themselves and are fully connected with the clique in $V_1$.
Precisely, $G=(V,E)$ with 
$(i,j)\in E$ if $i \in V_1$ or $j\in V_1$.

\begin{lemma}[Lower bound for clique-star sampling]
	\label{lem:lb2}
	Under the clique-star sampling over a clique of size $\ell$,  there exists 
	an absolute constant $c$ such that 
	$\lambda_{G,r}^* \leq c d (r(\min\{\ell,r\}))^{-1/4}$.
\end{lemma}
Together with Theorem \ref{thm:lb}, this implies that if $\ell\leq c^{4}C^{-4} r^{1-4/k}$, then 
 with probability at least $1/4$ any estimator makes an multiplicative error larger than two.
This implies that the total number of edges in the pattern graph should be $O(dr^{1-4/k})$ for accurate estimation.
Together with the upper bound on clique sampling in Theorem \ref{thm:graph_ub}, this shows that 
the sample complexity can drastically change based on the pattern of your sampling model. 
Clique sampling requires only $O(r^{2-4/k})$ samples (for $k=3$) whereas 
clique-star sampling requires at least $O(dr^{1-4/k})$. A proof of Lemma \ref{lem:lb} and Lemma \ref{lem:lb2} is given in Section \ref{sec:proof_lb_lem} and \ref{sec:proof_lb_lem2} respectively.

\section{Discussion}
\label{sec:discussion}

We list some observations and future research directions. 



\noindent
{\bf Complexity of the estimator beyond $k=7$.} 
For $k\geq8$, 
our approach of using matrix operations to count (the weights of) walks for each pattern $H\in \cH_k$ 
can potentially be extended. 
However, the complexity of the problem fundamentally changes for $k\geq 8$. 
As our estimator is at least as hard as counting small structures in a simple (unweighted) graph, 
we can borrow known complexity results to get a lower bound. 
For instance, for $k\geq 8$, we need to count $K_4$ in a graph, which 
 the best known run time is $O(d^{\alpha+1})$ for general graphs  \citep{KKM00}. 
For general $k$, under standard hardness assumptions,  \cite{FG04} show that 
 there is no  algorithm with run time $O(f(k) d^c)$ for counting cycles of length $k$, for any function $f(k)$ 
 and a constant $c$ that does not depend on $k$. 
In comparison, finding {\em one cycle} of length $k$ can be done in time $2^{O(k)} d^\alpha$  \citep{AYZ97}.
This implies that the complexity should scale as $O(d^{f(k)})$, and we believe 
$f(k)$ should be larger than  $(\alpha \sqrt{2k}/3)$.
The reason is that for 
$k\geq {\ell \choose 2}$ for an odd $\ell$, our estimator needs to count the number of cliques $K_\ell$ 
of size $\ell$. 
Similarly, for $k\geq (1/2) \ell^2$ for an even $\ell$, we require counting $K_\ell$. 
The best known algorithm for counting $K_\ell$ 
takes time $O(\min\{ d^{1+\alpha\lceil(\ell-1)/3\rceil} , d^{2+\alpha\lceil(\ell-2)/3\rceil}  \})$ for general graphs 
\citep[Theorem 6.4]{AYZ97}. 
Putting these bounds together, we believe that the estimator take time at least $d^{\alpha \sqrt{2k}/3}$.

\bigskip 
\noindent
{\bf Graph sampling.} 
Typical guarantees known for matrix completion assumes the Erd\"os-R\'enyi sampling. 
One exception is the deterministic 
sampling studied by  \cite{BJ14}, 
but such generalization in sampling comes at a price of requiring more strict assumptions on the matrix $M$. 
We propose graph sampling, which can potentially capture how estimation guarantees 
depends explicitly on the pattern $G$, and still remain analytically tractable. 
We give such examples for special  graphs in Section \ref{sec:gs}, and 
graph sampling model can potentially be used to bridge the gap in sampling models between 
theory and practice.

\bigskip 
\noindent
{\bf (Standard) rank estimation.}
As several popular matrix completion approaches require the knowledge of the rank of the original matrix, 
it is of great practical interest to estimate the standard rank of a matrix from sampled entries. 
Our framework in Section \ref{sec:genrank} provides a way to estimate the standard rank from samples. 
However, there are a few parameters that needs to be tuned, such as the thresholds $c_1$ and $c_2$,  
and the degree of the polynomial approximation and the degree of the Schatten norm. 
For rank estimation, 
\cite{keshavan2009gradient} give an estimator that 
is provably correct in the regime where matrix completion works, 
justifying the requirement that popular matrix completion algorithms \citep{KMO10IT,JNS13} 
need to know the underlying rank. 
However, in the regime of our interest, which is below the standard matrix completion threshold, the algorithm fails miserably and 
there are no guarantees. 
In a more recent work, \cite{saade2015matrix} propose a novel rank estimator of 
counting the negative eigenvalues of Bethe Hessian matrix.  
It is an interesting future direction to build upon our framework to 
provide a guideline for choosing the parameters for standard rank estimation, and compare its performance to existing methods.

\begin{figure}
	\begin{center}
	\includegraphics[width=0.5\textwidth]{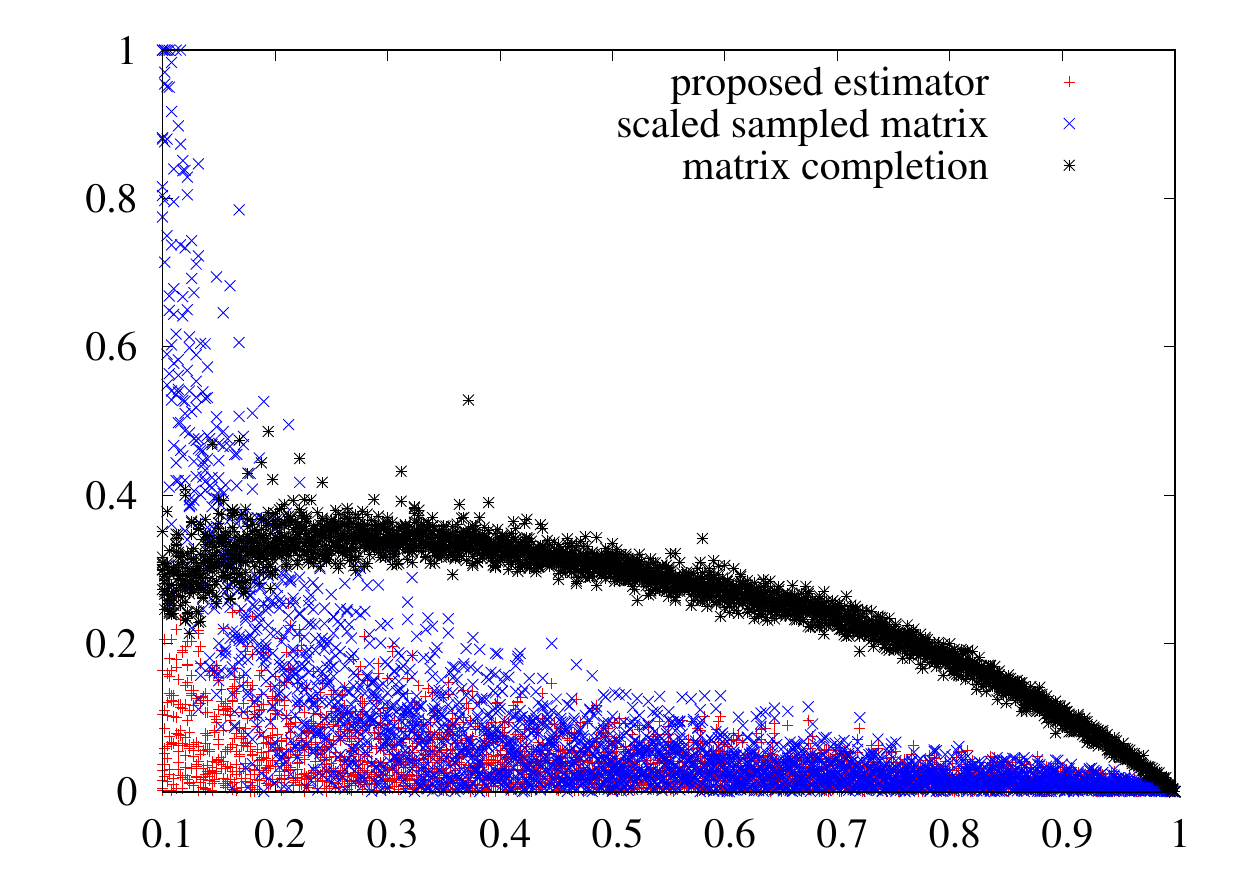}
	\put(-160,-3){\small{$\;$sampling probability, $p$}}
	\put(-230,47){\rotatebox{90}{\parbox{4cm}{\small{relative error}}}} 
	\put(-150,163){\large{$d=500$, $r =500$}}
	\end{center}
	\caption{For a matrix with a very small effective rank, the gap between the proposed estimator and 
		the simple scaled sampled matrix approach is smaller. }
	\label{fig:discussion}
\end{figure}

\bigskip 
\noindent
{\bf The effect of the effective rank.} 
One property of the Schatten norm is that  
as $k$ gets large and as the singular values have small effective rank (meaning that they decay fast), 
the summation is dominated by the largest few singular values. 
In such scenarios, in the estimation problem, 
any algorithm that tracks the first few singular values correctly would achieve small error. 
Hence, the gap get smaller as effective rank gets smaller, 
between the proposed estimator and the simple Schatten $k$-norm of the rescaled sampled matrix, as depicted in Figure \ref{fig:discussion}.   
We are using the same setting as those in Figure \ref{fig:compare} with 
a full rank matrix $M$ with $r=d=500$, but the effective rank is relatively small as 
the singular values are decaying as  $\sigma_i = 1/i^2$. 
For the current choice of $k=5$, notice that the contribution 
in $\|M\|_k^k$ of the $2$nd singular value is a factor of $2^{10}$ smaller than the top singular value, making it 
effectively a rank one matrix. 

\bigskip 
\noindent
{\bf Technical challenges.}  
The technical challenge in proving bounds on the necessary number of samples needed to estimate Schatten $k$-norms lies in getting tight bounds on the variance of the estimator. 
Variance is a function of weighted counts of each pseudograph of $2k$-closed walks, in the complete matrix. 
As the weight of each walk can be positive or negative, 
significant cancellation occurs when we sum all the weights. 
However, this stochastic cancellation is hard to capture in the analysis and we assume the worst case when all the weights are positive, which cannot occur for  incoherent and well-conditioned matrices. 
This weakness of the analysis leads to the requirement of rank being sufficiently small in the case of Erd\"os-R\'enyi sampling and $k$ small in the case of clique sampling. We believe these bounds can be tightened and the same is reflected in the numerical simulations which show the same scaling holds for all small values of $k$ and rank close to the dimension of the matrix. 

\section*{Acknowledgement}
This work is supported by NSF award  CNS-1527754 and CCF-1553452, and 
Google Faculty Research Award. 
The authors thank Prateek Jain for stimulating discussions on estimating Schatten norms from counts. 

\bibliography{schatten}

\newpage
\appendix
\section*{Appendix}
\section{$k$-cyclic pseudographs}
	\label{sec:graphs}
	\input{cycles4.tex}

	\input{cycles5.tex}
	\input{cycles6.tex}

	\input{cycles7.tex}

\newpage
\section{Efficient computation of $\wgt_M(H)$ for $k\in\{4,5,6,7\}$ }
	\label{sec:matrix}
	\input{computation}

\newpage

\section{Algorithm for estimating Schatten $k$-norm}
\label{sec:kgreaterthan7}

\begin{algorithm}[ht]
\caption{Schatten $k$-norm estimator}
\label{algo:kgreaterthan7}
  \begin{algorithmic}[1]
    \REQUIRE $\cP_\Omega(M)$, $k$, $\cH_k$, $p(H)$ for all $H \in \cH_k$
	\ENSURE $\hTheta_k(\cP_\Omega(M))$    
    	\STATE $\hTheta_k(\cP_\Omega(M)) \leftarrow 0$ 
		\STATE For all $H \in \cH_k$, let ${\rm simple}(H)$ be $H$ where multiple edges (self loops) are condensed into one.
		\STATE Let $\{\cH_{k,i}^{\rm iso}\}_{1 \leq i \leq r}$ be a partition of the collection $\cH_k$ such that $H,H' \in \cH_{k,i}^{\rm iso}$ if ${\rm simple(H)}$ is isomorphic to ${\rm simple(H')}$, for all $1 \leq i \leq r $.  Let $F_{k,i} \equiv {\rm simple(H)}$, and  $p(F_{k,i}) \equiv p(H)$ for any $H \in \cH_{k,i}^{\rm iso}$.
		\STATE Let $W_{k,i}$ be a collection of all possible $k$-closed walks on $F_{k,i}$ that uses each edge at least once. 	
		\FOR {$1 \leq \ell \leq k$}
		\STATE Enumerate a list, $\mathcal{L}_{\ell}$, of all connected $\ell$-vertex induced subgraphs (possibly with loops) of the graph $G([d],\Omega)$
		\FORALL {$g \in \mathcal{L}_{\ell}$}
		\STATE Enumerate a list, $\mathcal{S}_{g,\ell}$ of all connected $\ell$-vertex subgraphs of the graph $g$ by removing one or more edges
		\FORALL {$h \in \mathcal{S}_{g,\ell}$} 
		\FOR {$1 \leq i \leq r$}
		\IF{$h$ is isomorphic to $F_{k,i}$}
		\STATE $\hTheta_k(\cP_\Omega(M)) \leftarrow \hTheta_k(\cP_\Omega(M)) + \frac{1}{p(F_{k,i})} \sum_{w \in W_{k,i}}\wgt_{\cP_{\Omega}(M)}(w)$ 
		\ENDIF
		\ENDFOR
		\ENDFOR
		\ENDFOR 
		\ENDFOR
      \end{algorithmic}
\end{algorithm}

\section{Algorithm for computing the Chebyshev polynomial}
\label{sec:cheby}

\begin{algorithm}[ht]
\caption{Chebyshev polynomial of the first kind approximating  $H_{c_1,c_2}(x)$ }
\label{algo:cheby}
  \begin{algorithmic}[1]
    \REQUIRE $H_{c_1,c_2}$, $c_1$, $c_2$, and target accuracy $\delta=0.1$
	\ENSURE  Chebyshev polynomial $q(x)$ of first kind 
	\STATE $g(x) \equiv \frac{x-c_2}{c_1 - c_2}$
	\STATE $T_0(x) \equiv 1$, $T_1(x) \equiv  x$   
    	\STATE $q(x) \leftarrow \frac{1}{\pi} \int_{c_2}^{c_1} (1-x^2)^{-1/2} g(x)T_0(x)dx + \frac{1}{\pi} \int_{c_1}^{1} (1-x^2)^{-1/2}T_0(x)dx$
    	\STATE $i = 1$
    	\WHILE{$ \sup_{x \in [0,1]} |q(x) - H_{c_1,c_2}(x)| \geq \delta$}
    	\STATE $q(x) \leftarrow  q(x) + \frac{2T_i(x)}{\pi}\int_{c_2}^{c_1} (1-x^2)^{-1/2} g(x)T_{i}(x)dx + \frac{2T_i(x)}{\pi}\int_{c_1}^{1} (1-x^2)^{-1/2} T_{i}(x)dx $ 
    	\STATE $ i \leftarrow i +1$ 
    	\STATE $T_i(x) \equiv 2x T_{i-1}(x) - T_{i-2}(x) $
    	\ENDWHILE
      \end{algorithmic}
\end{algorithm}

\newpage
\section{Proofs}
We provide proofs for main results and technical lemmas. 

\subsection{Proof of Theorem \ref{thm:er_ub}}
\label{sec:proof_er_ub} 
Consider $\widetilde W$ to be the collection of all length $k$ closed walks on a complete graph of $d$ vertices. 
Here we slightly overload the notion of complete graph to refer to an undirected graph with 
not only all the ${d(d-1)/2}$ simple edges but also with $d$ self loops as well. 
Construct the largest possible collection $W$ from $\widetilde W$ wherein each walk has distinct weights 
that is $\wgt(w)\neq\wgt(w')$ for all $w,w'\in W$. 
We partition $W$ 
according to the pattern among $k$-cyclic pseudographs, 
which are further partitioned into four groups. 
The estimator \eqref{eq:estimate} can be re-written as 
\begin{eqnarray}
	\label{eq:estimate1}
	   \hTheta_k(\cP_\Omega(M)) &  = & \sum_{w \in W } \frac{ c(H(w))}{p(H(w))} \,\wgt_{\cP_{\Omega}(M)}(w)   \nonumber \\
	   		& = & \sum_{H \in \cH_k}  \Big\{ \frac{c(H)}{p(H)} \, \sum_{w: H(w)=H} \wgt_{M}(w)\, {\mathbb I}(w\subseteq \Omega) \Big\}   \\
			& = & \sum_{ i=1}^4 \sum_{H \in \cH_{k,i}}  \Big\{ \frac{c(H)}{p(H)} \, \sum_{w: H(w)=H} \wgt_{M}(w) \,{\mathbb I}(w\subseteq \Omega) \Big\}   \;, 
\end{eqnarray}
where we write $w\subseteq \Omega$ to denote 
the event that all the edges in the walk $w$ are sampled, 
and we define 
\begin{itemize}
\item $\cH_{k,1} \equiv \{C_k\}$ is just a (set of a) simple cycle of length $k$ and there are total $|\{w \in W :H(w)\in \cH_{k,1}\}|={d \choose k}( k!/2k) \leq (d^k/2k)$ corresponding walks to this set, and  $c(C_k)=2k$. 
\item $\cH_{k,2} \equiv \{ H(V_H,E_H) \in \cH_k: |V_H| \leq k-1 \text{ and no self loops}\}$, and there are total 
	$|\{w\in W:H(w)\in \cH_{k,2}  | \leq d^{k-1}$ corresponding walks to this set. 
\item $\cH_{k,3} \equiv \bigcup_{s = 1}^{k-1} \cH_{k,3,s} \text{ where } \cH_{k,3,s} = \{H \in \cH_k \text{ with $s$ self loops} \}$, and 
	there are total $|\{w\in W:H(w)\in \cH_{k,3}\}| \leq d^{k-s}$ corresponding walks in this set. 
\item $\cH_{k,4} \equiv \{H(V_H,E_H) \in \cH_k: |V_H| =1 \}$ is a (set of a) graph with $k$ self loops and there are total 
	$|\{w\in W:H(w)\in \cH_{k,4}\}|  = d$ corresponding walks to  this set. 
\end{itemize}
Given this unbiased estimator, we provide an upper bound on the variance of each of the partitions to prove concentration with Chebyshev's inequality. 
For any walk $w \in  W$, let $|w|$ denote the number of unique edges (including self loops) that the walk $w$ traverses. Let $|w \cap w'|$ denote the number of unique overlapping edges (including self loops) of walks $w$ and $w'$. We have,  
\begin{eqnarray}
	   {\rm Var}\big(\hTheta_k(\cP_\Omega(M))) & = & 2\sum_{\ell=1}^{k-1}\sum_{\substack{w\neq w' \in \widetilde W\\ |w \cap w'| = \ell}}
	   {\rm Covar} \Bigg(\frac{\mathbb{I}(w \subseteq \Omega) \wgt_{M}(w) c(H(w))}{p(H(w))}  \,,   \frac{\mathbb{I}(w' \subseteq \Omega) \wgt_{M}(w') c(H(w'))}{p(H(w'))}  \Bigg)\nonumber\\  
	   && + \;\;\sum_{ i=1}^4 \sum_{H \in \cH_{k,i}}  \Big\{ \frac{c(H)^2}{p(H)^2} \, \sum_{w: H(w)=H} \wgt_{M}(w)^2 {\rm Var} \Big( {\mathbb I}(w\subseteq \Omega) \Big) \Big\} \\
	   & < &  4\sum_{\ell=1}^{k-1}\sum_{\substack{w\neq w' \in  W\\ |w \cap w'| = \ell}}
	    \E\Big[ \mathbb{I}(w \subseteq \Omega) \mathbb{I}(w' \subseteq \Omega)\Big] \Bigg(\frac{ \big| \wgt_{M}(w) \,\wgt_{M}(w')\big| c(H(w)) c(H(w'))}{p(H(w))\,p(H(w'))}  \Bigg)\nonumber\\  
	   && + \;\;  \sum_{ i=1}^4 \sum_{H \in \cH_{k,i}}    \sum_{w: H(w)=H} \frac{c(H)^2 \wgt_{M}(w)^2}{p(H)^2} \E\Big[  {\mathbb I}(w\subseteq \Omega) \Big]   \,\label{eq:ub13}.
\end{eqnarray}
Recall from the definition of incoherence that 
$|M_{ii}|\leq \sigma_1(M) \mu r / d$ and $|M_{ij}|=\sigma_1(M) \mu r^{1/2} / d$, and let 
$\alpha=\sigma_1(M) \mu r^{1/2} / d$ denote the maximum off-diagonal entry, such that 
$|M_{ij}|\leq \alpha$ and $|M_{ii}|\leq \alpha \sqrt{r}$ for all $i,j\in[d]$.
Let $A_{p,k,\alpha,d}=d^k\alpha^{2k}/p^k$ denote the target scaling of the variance, then 
\[ \sum_{H \in \cH_{k,i}}    \sum_{w: H(w)=H} \frac{c(H)^2\,\wgt_{M}(w)^2}{p(H)^2} \E\Big[  {\mathbb I}(w\subseteq \Omega) \Big]   \;  \leq \; \]
\begin{numcases}{}
\frac{d^k}{2k}\frac{(2k)^2\alpha^{2k}}{p^k} = 2k A_{p,k,\alpha,d} \,,  & \text{for }$i=1$  \label{eq:ub1} \;,\\
d^{k-1} \frac{f(k)^2 \alpha^{2k}}{p^{k}} = \frac{f(k)^2}{ d}  A_{p,k,\alpha,d} \,, & \text{for } $i=2$\label{eq:ub3} \;,\\
d\frac{r^k\alpha^{2k}}{p} = \frac{r^k p^{k-1}}{d^{k-1}}  A_{p,k,\alpha,d} \,, & \text{for } $i = 4$\;,\label{eq:ub5}
\end{numcases}
and for $i=3$ and for $1 \leq s \leq k-1$, we have 
\begin{eqnarray}
	\sum_{H \in \cH_{k,3,s}}    \sum_{w: H(w)=H} \frac{c(H)^2\,\wgt_{M}(w)^2}{p(H)^2} \E\Big[  {\mathbb I}(w\subseteq \Omega) \Big] 
	\; \leq \; d^{k-s}\frac{f(k)^2 \alpha^{2k}r^s}{p^{k}} = \frac{f(k)^2  r^s}{ d^s}  A_{p,k,\alpha,d}\;, 
	\label{eq:ub4}
\end{eqnarray}
where $c(H)$ is defined as the multiplicity of walks with the same weight satisfying  $c(H)\leq f(k)$. 
For $w \neq w'$ and $|w \cap w'| = \ell$, where the range of $\ell$ varies across equations depending upon the set to which $w,w'$ belongs, we have the following: 
\[\sum_{\substack{w \neq w' \in W\\ |w\cap w'| = \ell, H(w) \in \cH_{k,i,s}, H(w')\in\cH_{k,i',s'} }}\E \big[\mathbb{I}(w \in \Omega)\mathbb{I}(w' \in \Omega)\big]\frac{\big|\wgt_{M}(H(w))\wgt_{M}(H(w')) \big|\, c(H(w))c(H(w'))}{p(H(w))p(H(w'))} \;\leq \]
\begin{numcases}{}
\frac{d^k d^{k-(\ell+1)}}{2k} \frac{ \alpha^{2k}(2k)^2}{p^{\ell}}  \;=\; \frac{(dp)^{k-{\ell}}}{d}\ 2k   A_{p,k,\alpha,d},  & \text{for } $i=i'=1$ \label{eq:ub2}\\
\frac{f(k)^2d^{k-1}d^{k-1-(\ell+1)}\alpha^{2k}}{p^{\ell}} \; \leq \; \frac{f(k)^2 (dp)^{k-\ell}}{d^3} A_{p,k,\alpha,d}& \text{for }$i=i'= 2$\label{eq:ub7}\\
\frac{f(k)^2 d^{k-s} d^{k-s'-\ell}\alpha^{2k-s-s'}(\alpha\sqrt{r})^{s+s'}}{p^{\ell}} \;\leq \;  \frac{f(k)^2 (dp)^{k-\ell}}{(d/\sqrt{r}  )^{s+s'}} A_{p,k,\alpha,d}\,, & \text{for }  $ i=i'=3$\label{eq:ub6}\\
\frac{f(k)^2 d^{k}d^{k-1-(\ell+1)}\alpha^{2k}}{p^{\ell}} \; \leq \; \frac{f(k)^2 (dp)^{k-\ell}}{d^2} A_{p,k,\alpha,d}& \text{for }$i=1, i' =2$\label{eq:ub8}\\
\frac{f(k)^2 d^{k}d^{k-s-(\ell+1)}\alpha^{2k-s}(\alpha\sqrt{r})^s}{p^{\ell}} \; \leq \; \frac{f(k)^2 (dp)^{k-\ell}}{d(d/\sqrt{r})^s}A_{p,k,\alpha,d}& \text{for }$ i=1, i' =3$\label{eq:ub9}\\
\frac{f(k)^2 d^{k-1}d^{k-s-(\ell+1)}\alpha^{2k-s}(\alpha\sqrt{r})^s}{p^{\ell}} \; \leq \; \frac{f(k)^2 (dp)^{k-\ell}}{d^2(d/\sqrt{r})^s} A_{p,k,\alpha,d}& \text{for }$ i=2, i' =3 $\label{eq:ub10}\\
\frac{f(k)^2 d d^{k-s-\ell}\alpha^{k-s}(\alpha\sqrt{r})^{k+s}}{p^{\ell}} \; \leq \; \frac{f(k)^2 (dp)^{k-\ell}}{d^{k-1}(d/\sqrt{r})^{k+s}} A_{p,k,\alpha,d}& \text{for }$ i=3, i' =4$\label{eq:ub11}\,,
\end{numcases}
where \eqref{eq:ub11} is valid only for $\ell=1$. Note that for any $w$ with $H(w)\in \cH_{k,1} \bigcup \cH_{k,2}$, 
it has no overlap with $w'$ such that $H(w')\in \cH_{k,4}$.

Observe that $ {\rm Var}\big(\hTheta_k(\cP_\Omega(M)))$ as bounded in \eqref{eq:ub13} is upper bounded by the sum of quantities in \eqref{eq:ub1}-\eqref{eq:ub11},  summating over all possible values of $1 \leq \ell \leq k-1$, and $1 \leq s,s' \leq k-1$. 
Let $h(k) \equiv f(k)^2 A_{p,k,\alpha,d}$. Observe that quantities in \eqref{eq:ub1},\eqref{eq:ub3}, and \eqref{eq:ub4} are upper bounded by $h(k)$. Quantities in \eqref{eq:ub2}-\eqref{eq:ub11} are upper bounded by 
$h_1(k) \equiv h(k){(dp)^{k-1}}/{d}$. Quantity in \eqref{eq:ub5} is upper bounded by $h_2(k) \equiv h(k) r^kp^{k-1}/d^{k-1}$. 

Given   $\norm{M}_k^{k} \geq r(\sigma_{\min})^k$, 
recall a bound on off diagonals of matrix $M$ by $|M_{ij}| \leq \alpha = \mu \sigma_{\max}\sqrt{r}/d$ 
and $A_{p,k,\alpha,d}=d^k\alpha^{2k}/p^k$. This gives  
\begin{align}\label{eq:ub15}
	\frac{A_{p,k,\alpha,d}}{\norm{M}_k^{2k}}  \;\leq\; \frac{\kappa^{2k} \mu^{2k}r^{k-2}}{ d^k p^k} \;.
\end{align}
Using Chebyshev's inequality and collecting all terms in the upper bound on the variance, we have for sufficiently large $d$, the following bound:
\begin{eqnarray}\label{eq:ub00}
	\P\Bigg( \frac{\big|\hTheta_k(\cP_\Omega(M)) - \norm{M}_k^k \big|}{\norm{M}_k^k} \;\;\geq\;\; \delta  \Bigg) & \leq & 
	\frac{(\kappa\mu)^{2k}f(k)^2r^{k-2}}{\delta^2 (dp)^k}\max \Bigg\{1, \frac{(dp)^{k-1}}{d}, \frac{r^{k}p^{k-1}}{d^{k-1}}\Bigg\}\,,
\end{eqnarray}
where the second and the third term in the $\max$ expression follow by evaluating $h_1(k)$ and $h_2(k)$. If sampling probability $p$ is small enough such that $dp \leq Cd^{1/(k-1)}$ for some constant $C$, then the second and the third terms are smaller than the first term. Hence, the desired result in Theorem \ref{thm:er_ub} follows.

\subsection{Proof of Theorem \ref{thm:er_ub2}}
\label{sec:proof_er_ub2}

We can prove a Bernstien-type bound on accuracy of the estimator. The estimator \eqref{eq:estimate} can be re-written as a multi-linear polynomial function of $d(d+1)/2$ i.i.d. Bernoulli($p$) random variables.
\begin{eqnarray}
	\label{eq:estimate1}
	   \hTheta_k(\cP_\Omega(M)) &  = & \sum_{w \in W } \bigg\{\frac{ c(H(w))}{p(H(w))}\,\wgt_{M}(w) \,\prod_{ (i,j) \in {\rm unique}(w)} {\mathbb I}((i,j)\in \Omega)\bigg\}\,,
\end{eqnarray}
where ${\mathbb I}((i,j)\subseteq \Omega)$ is a random variable that takes value $1$ if the $(i,j)_{\rm th }$ entry of the matrix $M$ is sampled, and ${\rm unique}(w)$ denotes the set of the unique edges (and self loops) that the walk $w$ traverses.  Let $q$ denote the power of the polynomial function that is the maximum number of unique edges in the walk $w$, that is $q=k$. 

We use the following Bernstien-type concentration results of \cite{schudy2011bernstein} for the polynomials of independent random variables.
\begin{lemma}[\cite{schudy2011bernstein},Theorem 1.3]\label{lem:berns}
We are given $d(d+1)/2$ independent central moment bounded random variables $\{{\mathbb I}((i,j)\in \Omega)\}_{1\leq i \leq j\leq d}$ with same parameter $L$. We are given a multilinear polynomial $\hTheta_k(\cP_\Omega(M))$ of power $q$, then 
\begin{eqnarray}\label{eq:ub21}
\P\Big[\Big|\hTheta_k(\cP_\Omega(M)) - \E\big[\hTheta_k(\cP_\Omega(M))\big]\Big| \geq \lambda\Big] & \leq & e^2 \max\bigg\{e^{\frac{-\lambda^2}{{\rm Var}[\hTheta_k(\cP_\Omega(M))]R^{q}}}, \max_{t \in [q]} e^{-(\frac{\lambda}{\mu_t L^t R^q})^{1/t}} \bigg\}\,,
\end{eqnarray} 
where $R$ is some absolute constant and $\mu_t$ is defined as follows:
\begin{eqnarray} \label{eq:ub22}
\mu_t &=& \max_{\substack{S \subseteq \{(i,j): i,j \in [d]\}\\|S| = t}} \Bigg( \sum_{w \in W| w \supseteq S} \,\,\frac{ c(H(w))}{p(H(w))}\,|\wgt_{M}(w)| \prod_{(i,j) \in {\rm unique}(w) \setminus S} \E[{\mathbb I}((i,j)\in \Omega)]\Bigg)\,,
\end{eqnarray}
where $ w \supseteq S$ denotes that the walk $w$ comprises edges(and self loops) contained in the set $S$. $L$ is defined as follows:
A random variable $Z$ is called central moment bounded with real parameter $L > 0$, if for any integer $i \geq 1$ we have
\begin{eqnarray}
E\big[|Z - \E[Z]|^i\big] & \leq & i \, L \, \E[|Z - \E[Z]|^{i-1}]\,.
\end{eqnarray}
\end{lemma}
For Bernoulli random variables $L \in [1/4, 1]$. In the following, we show that $\mu_t \leq (\mu\sigma_{\max})^k g(k)r^k/(d (dp)^t) $, for $t \in [k]$. Using Lemma \ref{lem:berns}, along with $\norm{M}_k^k \geq r(\sigma_{\min})^k $, the bound in \eqref{eq:ubChernoff1} follows immediately. 

To compute $\mu_t$, define a set of walks $W_{\ell,s,\hat s}$ such that $w \in W_{\ell,s,\hat s}$ has $0 \leq \ell \leq k$ unique edges and $0 \leq s \leq k$ unique self loops, and $\hat s$ total self loops with $ \ell +\hat s \leq k$. For the set $S$ as required in \eqref{eq:ub22}, let $S_{\tilde \ell,\tilde s}$ be a set of $\tilde \ell$ unique edges and $\tilde s$ unique self loops, with $|S_{\tilde\ell,\tilde s}| = \tilde \ell+ \tilde s$ where $1 \leq  \tilde \ell+\tilde s \leq k$.
Therefore, we have
\begin{eqnarray*} \label{eq:ub23}
\mu_t &=& \max_{\substack{S_{\tilde \ell, \tilde s}\\: \tilde \ell + \tilde s = t}} \Bigg( \sum_{\substack{0 \leq s\leq \hat s \leq k\\ \ell \in [k]: \ell+ \hat s \leq k} }\sum_{\substack{w \in W_{\ell,s,\hat s}\\: w \supseteq S_{\tilde \ell,\tilde s}}} \,\,\frac{ c(H(w))}{p(H(w))}\,|\wgt_{M}(w)| \prod_{(i,j) \in {\rm unique}(w) \setminus S_{\tilde \ell,\tilde s}} \E[{\mathbb I}((i,j)\subseteq \Omega)]\Bigg)\,\\
&\leq & \max_{\substack{S_{\tilde \ell, \tilde s}\\: \tilde \ell + \tilde s = t}} \Bigg( \sum_{\substack{0 \leq s\leq \hat s \leq k\\ \ell \in [k]: \ell+ \hat s \leq k} }\sum_{\substack{w \in W_{\ell,s,\hat s}\\: w \supseteq S_{\tilde \ell,\tilde s}}} \,\,\frac{f(k)}{p^{\ell+s}}\,{\alpha^k r^{\hat s/2}} p^{\ell+s -(\tilde \ell+\tilde s)}\Bigg)\,\\
&\leq & \max_{\substack{S_{\tilde \ell, \tilde s}\\: \tilde \ell + \tilde s = t}} \Bigg( \sum_{\substack{0 \leq s\leq \hat s \leq k\\ \ell \in [k]: \ell+ \hat s \leq k, \tilde s \leq s} }  \,\,\frac{d^{\ell -(1+\tilde \ell)}f(k)}{p^{\ell+s}}\,\frac{(\mu \sigma_{\max})^k r^{(k+\hat s)/2}}{d^k} p^{\ell+s -(\tilde \ell+\tilde s)}\Bigg)\,\\
&= & \max_{\substack{S_{\tilde \ell, \tilde s}\\: \tilde \ell + \tilde s = t}} \Bigg( \sum_{\substack{0 \leq s\leq \hat s \leq k\\ \ell \in [k]: \ell+ \hat s \leq k, \tilde s \leq s} }  \,\,\frac{f(k)(\mu \sigma_{\max})^k r^{(k+\hat s)/2}}{dd^{(k-\ell-\tilde s)}(dp)^{(\tilde \ell+\tilde s)}}\Bigg)\,\\
&\leq & \max_{\substack{S_{\tilde \ell, \tilde s}\\: \tilde \ell + \tilde s = t}} \Bigg(\frac{k^3f(k)(\mu \sigma_{\max})^k r^{(k+\hat s)/2}}{dd^{(k-\ell-\tilde s)}(dp)^{(\tilde \ell+\tilde s)}}\Bigg) \;\; \leq \;\; \frac{(\mu \sigma_{\max})^k g(k)r^k}{d (dp)^t}\,.
\end{eqnarray*}

\subsection{Proof of Theorem \ref{thm:lb}}
\label{sec:proof_lb}
The proof technique is a generalization to a rank $r$ symmetric matrix of the proof given by \cite{LNW14} for deriving lower bound on the size of a random bi-linear sketch needed for approximating Schatten norm of any matrix. It also draws on the techniques used in \cite{andoni2013tight} for proving a lower bound on the size of the linear sketches of moments. 

We prove Theorem \ref{thm:lb} for an arbitrary fixed relabeling permutation $\pi$ of the graph nodes. Indeed, by Yao's minimax principle, it suffices to give two distributions on matrix $M \in \cM_r$ for which the $\norm{M}_k$ values differ by a constant factor with high probability, but for any relabeling permutation $\pi$ of the nodes of the pattern graph $G$, the induced distributions on the sampled entries $\cP_{\Omega}(M)$ corresponding to the relabeled graph $G_{\pi}(\widetilde V, \Omega)$, have low total variation distance.

For positive $C>0$ to be specified later, define $\lambda \equiv Cdr^{1/k-1/2}$.
We construct distributions $\cD_1$ and $\cD_2$ for $M \in \cM_{r,\mu}$ with $\mu = C'\sqrt{\log r}$, for some absolute constant $C'$, such that the following holds: 
\begin{enumerate}
\item $\norm{M}_k \leq \lambda $ on the entire support of $\cD_1$, and $\norm{M}_k \geq 4\lambda$ on the entire support of $\cD_2$.
\item Let $\cE_1$ and $\cE_2$ denote the distribution of the sampled matrix  $\cP_{\Omega}(M)$ when $M$ is drawn from $\cD_1$ and $\cD_2$ respectively. 
Recall that $\Omega$ is the set of edges of the relabeled graph $G_{\pi}(\widetilde V, \Omega)$ as defined in Section \ref{subsec:lb}.
If $\lambda_{G,r}^*\geq \lambda$ then, 
  the total variation distance between $\cE_1$ and $\cE_2$ is 
bounded by  ${\rm{TV}}( \cE_1 , \cE_2) \leq 1/2$.
\end{enumerate}
The desired result \eqref{eq:lb1} follows from the above claims 
and the following relationship between statistical tests and estimators: 
\begin{eqnarray}
&&\underset{M \sim \frac{1}{2}(\cD_1 + \cD_2)}{\mathbb{P}} \bigg( \frac{1}{2} \norm{M}_k \leq \widetilde \Theta(\cP_{\Omega(M)}) \leq 2 \norm{M}_k \bigg) \nonumber\\
&& \leq \frac{1}{2}\underset{M \sim \cD_2}{\mathbb{P}} \bigg(\widetilde \Theta(\cP_{\Omega(M)}) \geq  2 \lambda \bigg) + \frac{1}{2}\underset{M \sim \cD_1}{\mathbb{P}} \bigg(\widetilde \Theta(\cP_{\Omega(M)}) \leq  2 \lambda \bigg) \\
&& \leq \frac{1}{2}\big(1 + {\rm{TV}}(\cE_1,\cE_2) \big) \;\;  \leq \;\; \frac{3}{4}\,, 
\end{eqnarray}
where the last inequality follows from the following characterization of the total variation distance ${\rm{TV}}(\cE_1,\cE_2) \equiv \sup_{A} |\cE_1(A) - \cE_2(A)|$.

To prove the two claims, 
we construct one of the desired rank-$r$ random matrix via tiling, i.e. 
covering the matrix with copies of a single $r\times r$ sub-matrix 
from the {\em{Gaussian Wigner Ensemble}}, where diagonals and off-diagonals(upper triangle) are both distributed as i.i.d. standard Gaussians. 
Another one is constructed by adding a  rank one perturbation. 
Precisely, we define a random matrix drawn from $\cD_1$ as follows.

A random $r \times r$ matrix $Z$ chosen from Gaussian Wigner Ensemble, $\mathcal{G}(r,r)$, is a symmetric matrix whose entries $Z_{i,i}$ and $Z_{i,j}$ for $i<j$ are independent with $N(0,1)$ distribution.
Define $B \equiv \ones_{\lceil d/r \rceil} \ones_{\lceil d/r\rceil}^\top$ to be an all-ones matrix of  size $\lceil d/r\rceil \times \lceil d/r \rceil$. 
Let $\bar \cD_1$ denote the distribution of  $M_1 = Y \otimes B$ where $Y \sim \mathcal{G}(r,r)$, 
and  $\otimes$ denotes the standard Kronecker product of two matrices. 
Note that the matrix norm of $M_1$ and $Y$ are related by $\norm{M_1}_k = \lceil d/r \rceil \norm{Y}_k$. 
Since the Schatten norm of $Y \sim \mathcal{G}(r,r)$ takes value on the entire $\reals_{+}$, we need to truncate it. 
We set $\cD_1$ to be $\bar \cD_1$ conditioned on the event $S_1 = \{M_1:\norm{M_1}_k \leq  \lambda, \textcolor{black}{\mu(M_1) \leq C'\sqrt{\log r}} \}$, i.e. $\cD_1(A) = \bar \cD_1 (A \cap S_1)/ \bar \cD_1 (S_1)$.

We define $\bar \cD_2$ by adding a rank one perturbation. 
Precisely, let $M_2=M_1 + (5/d)\lambda U$, 
 where $M_1 \sim \bar \cD_1$ and $U = uu^{\top} \otimes B$. 
 Here a random vector $u \in {\{\pm 1\}}^r$ is a vector of i.i.d. Rademacher random variables. Note that $U$ is a rank one matrix and $\norm{U}_k = \lceil d/r \rceil \norm{uu^\top}_k = d$. 
We set $\cD_2$ to be $\bar \cD_2$ conditioned on the event $S_2 = \{M_2:\norm{M_2}_k \geq  4\lambda,\textcolor{black}{\mu(M_2) \leq C'\sqrt{\log r}} \}$. 
Observe that $M_1 \sim \bar \cD_1$ and $M_2 \sim \bar \cD_2$ belong to $\reals^{d\times d}$, are symmetric and both are rank  at most $r+1$. 

Let $\bar \cE_1$ and $\bar \cE_2$ denote the distribution of $\cP_{\Omega}(M)$ when $M$ is drawn from $\bar \cD_1$ and $\bar \cD_2$ respectively. 
We first show that their total variation distance is not too large. 
Using the triangle inequality, we have
\begin{eqnarray}
\tv(\cE_1,\cE_2) &\leq&  \tv(\bar \cE_1, \bar \cE_2) + \tv(\bar \cE_1, \cE_1) + \tv(\bar \cE_2, \cE_2)\nonumber\\
 & \leq &    \tv(\bar \cE_1, \bar \cE_2) + \tv(\bar \cD_1, \cD_1) + \tv(\bar \cD_2, \cD_2) \label{eq:lb2}\\
&=& \tv(\bar \cE_1, \bar \cE_2) + \underset{M_1 \sim D_1}{\mathbb{P}} \big((\norm{M_1}_k \geq  \lambda) \cup (\textcolor{black}{\mu(M_1) \geq C'\sqrt{\log r}}) \big) \nonumber\\
&& \;\; + \underset{M_2 \sim \cD_2}{\mathbb{P}} \big((\norm{M_2}_k \leq  4\lambda) \cup (\textcolor{black}{\mu(M_2) \geq C'\sqrt{\log r}}) \big)\,, \label{eq:lb3}
\end{eqnarray}
where \eqref{eq:lb2} follows from the data processing inequality and \eqref{eq:lb3} follows from ${\rm{TV}}(\cE_1,\cE_2) \equiv \sup_{A} |\cE_1(A) - \cE_2(A)|$. 
 We next show that the three terms in \eqref{eq:lb3} are sufficiently small.

We first provide an upper bound on $\tv(\bar \cE_1, \bar \cE_2)$. 
 As per our construction, only the upper triangular (including diagonals) of the upper-left submatrix of size $r \times r$ of $M_1 \sim \cD_1$ and $M_2 \sim \cD_2$ has unique entries and the rest are copies of these. 
Observe that the set of unique entries of $M_1$(or $M_2$) corresponding to any pattern graph $G(V,E)$ are precisely the following entries of the projection graph $\cP^{(r)}(G)$ that is defined in Section \ref{subsec:lb}:
\begin{eqnarray}\label{eq:lb_uniequentries}
E(\cP^{(r)}(G)) & \equiv & \Big\{(i,j): i \leq j \in [r], (i, j) \in \cP^{(r)}(G(V, E))\Big\}\,.
\end{eqnarray}


For the purpose of computing the total variation distance $\tv(\bar \cE_1, \bar \cE_2)$, 
it is sufficient to consider only $E(\cP^{(r)}(G_{\pi}))$ entries of $M_1$ distributed as 
 i.i.d. standard Gaussians $N(0,I_{\ell_1\times \ell_1})$, and the entries of $M_2$ distributed as  
 $N(W,I_{\ell_1\times \ell_1}))$, where $\ell_1 = |E(\cP^{(r)}(G_{\pi}))|$. 
 The random vector $W$ represents the rank one perturbation  and is distributed as 
\begin{eqnarray}
	W_{i,j} = (5/d) \lambda \,u_i u_j \,,\qquad (i,j) \in E(\cP^{(r)}(G_{\pi}))\,.
\end{eqnarray}   

To bound total variation distance between $\bar \cE_1$ and $\bar \cE_2$, we use the following lemma and the fact that for any two distributions $\mu$ and $\nu$, $\tv(\mu,\nu) \leq \sqrt{\mathcal{X}^2(\mu \,\|\, \nu)}$. Let $\mu \ast \nu$ denote the convolution of the density (or equivalently addition of the two random variables). 

\begin{lemma}[\cite{IS12}, p97] \label{lem:chisquare}
It holds that $\mathcal{X}^2(N(0,I_n)\ast \mu \,\| \,N(0,I_n)) \leq \E \exp({\ip{z}{z'}}) -1$, where $z,z' \sim \mu$ are independent. 
\end{lemma}

It follows that
\begin{eqnarray*}
	\tv(\bar \cE_1,\bar \cE_2) &\leq  & \sqrt{\E e^{\ip{W }{ W'}} - 1} \;\leq\; 1/5\,,
\end{eqnarray*}
for $\lambda^*_G \geq \lambda$ 
where the expectation is taken over independent $W $ and $W' $ which are identically distributed. 
We show that if  $\lambda^*_G \geq \lambda$ the last inequality holds, as following: 
 \begin{eqnarray}
&&\E_{ W ,  W' } \exp \big(\langle  W ,  W' \rangle \big)\nonumber\\ 
	&=&\E_{u,u'} \exp\bigg((5/d)^2\lambda^2 \sum_{(i,j) \in E(\cP^{(r)}(G_{\pi}))} u_i u'_i  u_j u'_j \bigg)\nonumber\\
 	&=&\E_{u} \exp\bigg((5/d)^2\lambda^2 \sum_{(i,j) \in E(\cP^{(r)}(G_{\pi}))} u_i  u_j \bigg)\label{eq:lba5}\\
 	&=&\E_{u} \bigg[\exp\bigg((5/d)^2\lambda^2 \sum_{\substack{ (i,j) \in E(\cP^{(r)}(G_{\pi}))\\:i \neq j }} u_i  u_j \bigg)\bigg]\exp\bigg((5/d)^2\lambda^2 \sum_{\substack{ (i,j) \in E(\cP^{(r)}(G_{\pi}))\\:i = j }}  u_i  u_j \bigg)\nonumber\\	
 	&\leq &\E_{ u} \bigg[\exp\bigg((5/d)^2\lambda^2 \sum_{\substack{ (i,j) \in E(\cP^{(r)}(G_{\pi}))\\:i \neq j }} 2 u_i u_j \bigg)\bigg]\exp\bigg((5/d)^2\lambda^2 \sum_{\substack{ (i,j) \in E(\cP^{(r)}(G_{\pi}))\\:i =j }}  u_i  u_j \bigg)\label{eq:lba2}\\
 	&=&\E_{ u} \bigg[\exp\bigg((5/d)^2\lambda^2 \sum_{\substack{ (i,j) \in \cP^{(r)}(G_{\pi})\\:i \neq j }}  u_i  u_j \bigg)\bigg]\exp\bigg((5/d)^2\lambda^2 \sum_{\substack{ (i,j) \in \cP^{(r)}(G_{\pi})\\:i= j }}  u_i  u_j \bigg)\label{eq:lba3}\\ 	
 	&=&\E_{ u} \bigg[\exp\bigg((5/d)^2\lambda^2 \sum_{(i,j) \in \cP^{(r)}(G_{\pi})}  u_i  u_j \bigg)\bigg]\nonumber\\
	&\leq& 1 +1/25\,, \label{eq:lba1}
\end{eqnarray}
where \eqref{eq:lba5} follows from the fact that $u,u'$ are i.i.d. Rademacher variables, \eqref{eq:lba2} follows from the fact that $f_{G,r}(\lambda)$ defined in \eqref{eq:fgrlambda} is non-decreasing in $\lambda$, \eqref{eq:lba3} follows from the definition of $E(\cP^{(r)}(G_{\pi}))$ in \eqref{eq:lb_uniequentries},and \eqref{eq:lba1} follows from the definition of $\lambda^*_G$ in \eqref{eq:deflambda}.

To bound the other two terms in \eqref{eq:lb3}, we use Wigner's semicircular law and its rate of convergence for Gaussian Wigner Ensemble, $\mathcal{G}(r,r)$ as defined above.  Consider the empirical spectral distribution of $Z \in \reals^{r \times r}$ as
\begin{eqnarray}
F_Z(x) = \frac{1}{r}|\{i: \lambda_i(Z) \leq x\}|.
\end{eqnarray} 
\begin{lemma}[\cite{wigner1955}] \label{lem:wigner}
Define $Z = (1/\sqrt{r})Y$ for $Y \sim \mathcal{G}(r,r)$. Then as $r \rightarrow \infty$ the empirical distribution $F_Z(x)$ of $Z$ converges weakly to the distribution $G(x)$ with density 
\begin{eqnarray}
g(t) = \frac{\sqrt{4-t^2}}{2\pi} \;\;t \in [-2,2]\,. 
\end{eqnarray}
\end{lemma}

\begin{lemma}[\cite{gotzerate}] \label{lem:wigner_convg}
	For any positive constant $\alpha > 0$, let $\ell_{r,\alpha} = \log r (\log \log r )^\alpha$.  There exists an absolute positive constant $C$ and $c$ such that for $r$ large enough,
	\begin{eqnarray}
	\P\bigg \{ \sup_{x} \big|F_{Z}(x) - G(x)\big| \geq r^{-1} \log r \ell_{r,\alpha}^6 \bigg\} & \leq &  C \exp\big\{-c\ell_{r,\alpha} \big\}\,.
	\end{eqnarray}
\end{lemma}
		
To bound the schatten norm of a matrix $Y \sim \mathcal{G}(r,r)$, along with Lemma \ref{lem:wigner} and Lemma \ref{lem:wigner_convg} we use the following. If $F(x)$ and $G(x)$ are cumulative distribution functions of densities $\mu, \nu$ then for any continuous and bounded function $f$, we have
\begin{eqnarray}
\bigg| \int fd\mu - \int f d\nu \bigg| & \leq & \norm{f}_{\infty} \sup_{x} \big|F(x) - G(x)\big|\,.
\end{eqnarray}
Choosing $f(x) = x ^{k}$ for $x \in [-2,2]$, we can see that for $k = O(\log r)$ there exists a constant $C>2$ such that with probability $1 - 1/80$ it holds that
\begin{eqnarray}\label{eq:lbYconc}
\norm{(1/\sqrt{r})Y}_k^k = \Bigg(\int_{-2}^{2} x^k \frac{\sqrt{4-x^2}}{2\pi} dx + o(1) \Bigg) r \;\leq \; (2^k+o(1))r \;\leq \; C^kr \,.
\end{eqnarray} 
Hence $\norm{Y}_k \leq Cr^{(1/k+1/2)}$. By construction of distribution $\bar \cD_1$, for $M_1 \sim \bar \cD_1$, $\norm{M_1}_k = (d/r)\norm{Y}_k \leq Cdr^{(1/k-1/2)} = \lambda$. Also, by construction $M_2 \sim \bar \cD_2$ is $M_2 = M_1 + (5/d)\lambda U$ where $\norm{U}_k = d$. Using triangle inequality, we have
\begin{eqnarray*}
\norm{M_2}_k &\geq& \norm{(5/d)\lambda U}_k - \norm{M_1}_k \nonumber \\ 
 	               &\geq& 5\lambda  - Cdr^{1/k-1/2} \;=\; 4\lambda\;,  
\end{eqnarray*}
Recall that, incoherence parameter $\mu(M)$ is defined as \textcolor{black}{$\mu(M) = \max_{i \neq j \in [d]} M_{i,j}/(|\sigma_{\max}(M)|\sqrt{r}/d)$}.
From \eqref{eq:lbYconc}, there exists a constant $0 <  C' < 1 $ such that with probability $1 - 1/160$ it holds that $\norm{Y}_2 \geq C'r$. 
\textcolor{black}{The integral evaluates to 1 for $k =2$.}
Therefore, the largest singular value of $M_1$ is lower bounded: $|\sigma_{\max} (M_1)| \geq C'd/\sqrt{r}$.   
Using the fact that there exists a constant $C''$ such that $\max_{i,j\in [r]} \{Y_{i,j}\} \leq C''\sqrt{\log r}$ with probability at least $1 - 1/160$, we have, $\mu(M_1) \leq (C''/C') \sqrt{\log r}$. The same $\mu(M_1)$ satisfies the upper bound on diagonals as well. 
Therefore, using union bound, the second and the third term in \eqref{eq:lb3} are upper bounded by $1/40$.	

\subsection{Proof of Lemma \ref{lem:lb}} 
\label{sec:proof_lb_lem}
Observe that for any given permutation $\pi$, $\cP^{(r)}(G_{\pi})$ as defined in Section \ref{subsec:lb} is a clique over a subset of nodes $\widetilde V_{\pi}$, where $|\widetilde V_{\pi}| \leq \min\{\ell,r\}$.
From the definition of $f_{G,r}(\lambda)$, \eqref{eq:fgrlambda}, we have the following: 
	\begin{eqnarray}
 f_{G,r}(\lambda) &=& \max_{\pi} \bigg\{\E_{u} \exp\bigg( (5/d)^2 \lambda^2 \sum_{ (i,j) \in \cP^{(r)}(G_{\pi}) } u_i u_j \bigg)\bigg\}
	 = \max_{\pi} \bigg\{\E_{ u} \exp\bigg((5/d)^2 \lambda^2 \big(\sum_{i \in \widetilde V_{\pi}} u_i\big)^2 \bigg)\bigg\}\nonumber\\
	 &=&   \max_{\pi} \bigg\{\sum_{t=0}^{\infty}\frac{(5/d)^{2t} \lambda^{2t} \E_{u}\big[\big(\sum_{i \in \widetilde V_{\pi}} u_i\big)^{2t}\big]}{t!}	\bigg\}
	 \;\leq \; \max_{\pi} \bigg\{ \bigg(1+ 2\sum_{t=1}^{\infty}\big((5/d)^{2} \lambda^{2}|\widetilde V_{\pi}|\big)^t\bigg) \bigg\}, \nonumber
	\end{eqnarray}
	where the inequality follows from the bound in \eqref{eq:lb6}. Therefore, from the definition of $\lambda^*_{G,r}$, we have that $\lambda^*_{G,r}$ is upper bounded by $2^{-4}d{(\min\{\ell,r\})}^{-1/2}$.
	
To bound $\E(\sum_{i \in \widetilde V_{\pi}} u _i)^{2t}$, for $t \in [1,\infty)$, using Hoeffding bound we have that
	\begin{eqnarray}
	\E\bigg| \sum_{i \in \widetilde V_{\pi}}  u _i \bigg|^{2t} & = & \int_{0}^{{|\widetilde V_{\pi}|}^{2t}} \P \bigg( \bigg| \sum_{i \in \widetilde V_{\pi}}  u _i \bigg|^{2t} \geq z\bigg)dz \;	\leq\;  2  \int_{0}^{|\widetilde V_{\pi}|^{2t}} \exp\bigg(\frac{-z^{1/t}}{2|\widetilde V_{\pi}|}\bigg) dz \;\leq\; 2(2|\widetilde V_{\pi}|)^t t!\,,\label{eq:lb6}
	\end{eqnarray}
	where the integral is evaluated by variable substitution. 	

\subsection{Proof of Lemma \ref{lem:lb2}} 
\label{sec:proof_lb_lem2}
For the given pattern graph $G$ and any given permutation $\pi$, let $\widetilde A_{\pi} \in \{0,1\}^{r \times r}$ be the adjacency matrix of the graph $\cP^{(r)}(G_{\pi})$ that is defined in Section \ref{subsec:lb}. Observe that for a permutation $\pi$, $\ell_\pi$ rows of $\widetilde A_{\pi}$ are all-ones and the remaining are all-zeros, where $\ell_\pi \leq \min\{\ell,r\}$. Let $A_{\pi}$ be a copy of $\widetilde A_{\pi}$ where all the diagonal entries are replaced with zero. Note that $\E_{u}(u^\top A_{\pi}u)^{2t+1} = 0$ for all $t\geq 0$, where $u_i$'s are i.i.d. Rademacher random variables. Define $C_{\pi} \equiv \exp((5/d)^2 \lambda^2\ell_\pi)$.

From the definition of $f_{G,r}(\lambda)$, \eqref{eq:fgrlambda}, we have the following: 
	\begin{eqnarray}
 f_{G,r}(\lambda) &=& \max_{\pi} \bigg\{\E_{u} \exp\bigg( (5/d)^2 \lambda^2 \sum_{ (i,j) \in \cP^{(r)}(G_{\pi}) } u_i u_j \bigg)\bigg\}
	 = \max_{\pi} \bigg\{C_{\pi}\E_{ u} \exp\bigg((5/d)^2 \lambda^2 (u^\top A_{\pi}u) \bigg)\bigg\}\nonumber\\
	 &=&   \max_{\pi} \bigg\{C_{\pi}\sum_{t=0}^{\infty}\frac{(5/d)^{4t} \lambda^{4t} \E_{u}\big[(u^\top A_{\pi} u)^{2t}\big]}{(2t)!}\bigg\}	
	 \;\leq \; \max_{\pi}\bigg\{C_{\pi}\bigg(1+ 4\sum_{t=1}^{\infty}\big(2c(5/d)^{2} \lambda^{2}\sqrt{\ell_{\pi} r}\big)^{2t}\bigg)\bigg\}, \nonumber
	\end{eqnarray}
	where the inequality follows from the bound in \eqref{eq:lb61}, and $c$ is some absolute constant. Therefore, from the definition of $\lambda^*_{G,r}$, we have that $\lambda^*_{G,r}$ is upper bounded by $cd({(\min\{\ell,r\})r)}^{-1/4}$.
	
To bound $ \E_{u}\big[(u^\top A_{\pi} u)^{2t}\big]$, for $t \in [1,\infty)$, we use Hanson-Wright Inequality. Observe that $\norm{A_{\pi}}_2 \leq \sqrt{\ell_{\pi}r}$, and $\norm{A_{\pi}}_{\rm F}^2 = (r -1) \ell_{\pi} < \ell_\pi r$.
	\begin{eqnarray}
	\E_{u}\big[(u^\top A_{\pi} u)^{2t}\big] & =&  \int_{0}^{(2\sqrt{r\ell_\pi})^{2t}} \P \big( (u^\top A_{\pi} u)^{2t} \geq z\big)dz \;+\; \int_{(2\sqrt{r\ell_\pi})^{2t}}^{(\ell_\pi r)^{2t}} \P \big( (u^\top A_{\pi} u)^{2t} \geq z\big)dz\nonumber\\
	& \leq &  \int_{0}^{(2\sqrt{r\ell_\pi})^{2t}} \exp\bigg(\frac{-c z^{1/t}}{4\ell_\pi r} \bigg)dz + \int_{(2\sqrt{r\ell_\pi})^{2t}}^{(\ell_\pi r)^{2t}} \exp\bigg(\frac{-cz^{1/(2t)}}{2\sqrt{\ell_\pi r}} \bigg)dz \nonumber\\
	& \leq & 2 (4 \ell_\pi r/c)^t t! + 2  (2\sqrt{\ell_\pi r}/c )^{2t} (2t)! \; \leq \;  4 (2 \sqrt{\ell_\pi r}/c )^{2t} (2t)! \,,\label{eq:lb61}
	\end{eqnarray}
	where the integral is evaluated by variable substitution.

\subsection{Proof of Theorem \ref{thm:graph_ub}}
\label{sec:proof_graph_ub} 
For a clique of size $m$ selected uniformly at random, we derive an upper bound on variance of our estimator. Following the notations defined in the proof of Theorem \ref{thm:er_ub}, we have the following bound on the variance.
\begin{eqnarray}
	   {\rm Var}\big(\hTheta_k(\cP_\Omega(M))) & = & 2\sum_{\ell=0}^{\textcolor{black}{k}}\sum_{\substack{w\neq w' \in \widetilde W\\ |w \cap w'| = \ell}}
	   {\rm Covar} \Bigg(\frac{\mathbb{I}(w \subseteq \Omega) \wgt_{M}(w) c(H(w))}{p(H(w))}  \,,   \frac{\mathbb{I}(w' \subseteq \Omega) \wgt_{M}(w') c(H(w'))}{p(H(w'))}  \Bigg)\nonumber\\  
	   && + \;\;\sum_{ i=1}^4 \sum_{H \in \cH_{k,i}}  \Big\{ \frac{c(H)^2}{p(H)^2} \, \sum_{w: H(w)=H} \wgt_{M}(w)^2 {\rm Var} \Big( {\mathbb I}(w\subseteq \Omega) \Big) \Big\} \\
	   & < &  2\sum_{\ell=0}^{\textcolor{black}{k}}\sum_{\substack{w\neq w' \in  W\\ |w \cap w'| = \ell}}
	    \E\Big[ \mathbb{I}(w \subseteq \Omega) \mathbb{I}(w' \subseteq \Omega)\Big] \Bigg(\frac{ \wgt_{M}(w) \,\wgt_{M}(w') c(H(w)) c(H(w'))}{p(H(w))\,p(H(w'))}  \Bigg)\nonumber\\  
&& - 
2\sum_{\ell=0}^{\textcolor{black}{k}}\sum_{\substack{w\neq w' \in  W\\ |w \cap w'| = \ell}}
\E\Big[ \mathbb{I}(w \subseteq \Omega)\Big]\E\Big[ \mathbb{I}(w' \subseteq \Omega)\Big] \Bigg(\frac{ \wgt_{M}(w) \,\wgt_{M}(w') c(H(w)) c(H(w'))}{p(H(w))\,p(H(w'))}  \Bigg)\nonumber\\
	   && + \;\;  \sum_{ i=1}^4 \sum_{H \in \cH_{k,i}}    \sum_{w: H(w)=H} \frac{c(H)^2 \wgt_{M}(w)^2}{p(H)^2} \E\Big[  {\mathbb I}(w\subseteq \Omega) \Big]   \,\label{eq:ub31}.
\end{eqnarray}
where we abuse the earlier defined notation $|w \cap w'|$ to denote the number of overlapping nodes in the two walks $w,w' \in W$ instead of number of overlapping edges. Note that in pattern sampling, covariance term for two walks that do not have any overlapping node is not zero. As earlier, we provide bound on each of the terms in \eqref{eq:ub31}.

Probability of any walk $w$ being sampled is $\P[w \in \Omega] =  {m \choose \ell} / {d \choose \ell} \leq f(\ell)m^\ell/d^{\ell}$, where $\ell$ is the number of unique nodes that the walk traverses and $f(\ell)$ is an exponential function in $\ell$.  
Recall that off diagonals of matrix $M$ are bounded by $|M_{ij}| \leq \alpha = \mu \sigma_{\max}\sqrt{r}/d$ and the diagonals are bounded by  $|M_{ii}| \leq \mu \sigma_{\max}r/d$.
We have, 
\[ \sum_{H \in \cH_{k,i}}    \sum_{w: H(w)=H} \frac{c(H)^2\,\wgt_{M}(w)^2}{p(H)^2} \E\Big[  {\mathbb I}(w\subseteq \Omega) \Big]   \;  \leq \; \]
\begin{numcases}{}
\frac{d^k}{2k}\frac{f(k)^2\alpha^{2k}d^k}{m^k} \leq \frac{f(k)^2(\mu \sigma_{\max})^{2k} r^k}{m^k} \,,  & \text{for }$i=1$  \label{eq:ub1} \;,\\
\bigg(\frac{d^{2}}{m}\bigg)^{k-1} f(k)^2 \alpha^{2k} = \frac{ m}{ d^2}  \frac{f(k)^2(\mu \sigma_{\max})^{2k} r^k}{m^k} \,, & \text{for } $i=2$\label{eq:ub3} \;,\\
\frac{d^2}{m}r^k\alpha^{2k} = \frac{r^k m^{k-1}}{d^{2k-2}}  \frac{f(k)^2(\mu \sigma_{\max})^{2k} r^k}{m^k} \,, & \text{for } $i = 4$\;,\label{eq:ub5}
\end{numcases}
and for $i=3$ and for $1 \leq s \leq k-1$, we have 
\begin{eqnarray}
	\sum_{H \in \cH_{k,3,s}}    \sum_{w: H(w)=H} \frac{c(H)^2\,\wgt_{M}(w)^2}{p(H)^2} \E\Big[  {\mathbb I}(w\subseteq \Omega) \Big] 
	\; \leq \;\bigg(\frac{d^{2}}{m}\bigg)^{k-s} f(k)^2 \alpha^{2k}r^s = \frac{ m^s r^s}{ d^{2s}} \frac{f(k)^2(\mu \sigma_{\max})^{2k} r^k}{m^k}\;, 
	\label{eq:ub4}
\end{eqnarray}

For any two walks $w,w'$ with $\ell \geq 0$ overlapping nodes, $ \P[w,w' \in \Omega]/(\P[w \in \Omega] \P[w' \in \Omega]) \leq f(k)d^{\ell}/m^{\ell}$.
For $w \neq w'$ and $|w \cap w'| = \ell$, where the range of $\ell$ varies across equations depending upon the set to which $w,w'$ belongs, we have the following: 
\[\sum_{\substack{w \neq w' \in W\\ |w\cap w'| = \ell \\H(w) \in \cH_{k,i,s}\\ H(w')\in\cH_{k,i',s'} }}\bigg(\E\Big[ \mathbb{I}(w \subseteq \Omega) \mathbb{I}(w' \subseteq \Omega)\Big] - \E\Big[ \mathbb{I}(w \subseteq \Omega)\Big]\E\Big[ \mathbb{I}(w' \subseteq \Omega)\Big] \bigg) \Bigg(\frac{ \wgt_{M}(w) \,\wgt_{M}(w') c(H(w)) c(H(w'))}{p(H(w))\,p(H(w'))}  \Bigg) \;\leq \;\]
\begin{numcases}{}
\frac{f(k)^2d^\ell}{m^\ell} \frac{(\mu \sigma_{\max})^{2k} r^2}{d^{\ell}} \;=\; \frac{f(k)^2(\mu \sigma_{\max})^{2k} \max\{r^2,r^{\ell}\}}{m^{\ell}},  & \text{for } $i=i'=1, \ell \geq 1$ \label{eq:ub42}\\
\frac{m^{2k-1}}{d^{2k}}\frac{d^{2k}f(k)^2(\mu \sigma_{\max})^{2k} r^2}{m^{2k}} \;=\; \frac{f(k)^2(\mu \sigma_{\max})^{2k} r^2}{m},  & \text{for } $i=i'=1, \ell = 0$ \label{eq:ub43}\\
\frac{f(k)^2d^\ell d^{2k-2-\ell}\alpha^{2k}}{m^{\ell}} \; \leq \; \frac{f(k)^2(\mu \sigma_{\max})^{2k} r^k}{m^\ell d^2},  & \text{for }$i=i'= 2$\label{eq:ub47}\\
\frac{f(k)^2d^\ell d^{2k-s-s'-\ell}\alpha^{2k}(\sqrt{r})^{s+s'}}{m^{\ell}} \; \leq \; \frac{f(k)^2(\mu \sigma_{\max})^{2k} r^k}{m^\ell d}, & \text{for }  $ i=i'=3$\label{eq:ub46}\\
f(k)^2d^{2}\alpha^{2k}(\sqrt{r})^{2k} \; \leq \; \frac{f(k)^2(\mu \sigma_{\max})^{2k} r^k}{d^{2k-2}/r^k}, & \text{for }  $ i=i'=4$\label{eq:ub414}\\
\frac{f(k)^2d^\ell d^{2k-1-\ell}\alpha^{2k}}{m^{\ell}} \; \leq \; \frac{f(k)^2(\mu \sigma_{\max})^{2k} r^k}{m^\ell d}, & \text{for }$i=1, i' =2$\label{eq:ub48}\\
\frac{f(k)^2d^\ell d^{2k-s-\ell}\alpha^{2k}(\sqrt{r})^{s}}{m^{\ell}} \; \leq \; \frac{f(k)^2(\mu \sigma_{\max})^{2k} r^k}{m^\ell d/\sqrt{r}},& \text{for }$ i=1, i' =3$\label{eq:ub49}\\
\frac{f(k)^2d^\ell d^{k+1-\ell}\alpha^{2k}(\sqrt{r})^{k}}{m^{\ell}} \; \leq \; \frac{f(k)^2(\mu \sigma_{\max})^{2k} r^k}{m^\ell d^{k-1}/(\sqrt{r})^k},& \text{for }$ i=1, i' =4$\label{eq:ub412}\,,\\
\frac{f(k)^2d^\ell d^{2k-1-s-\ell}\alpha^{2k}(\sqrt{r})^{s}}{m^{\ell}} \; \leq \; \frac{f(k)^2(\mu \sigma_{\max})^{2k} r^k}{m^\ell d^2/\sqrt{r}},& \text{for }$ i=2, i' =3 $\label{eq:ub410}\\
\frac{f(k)^2d^\ell d^{k-\ell}\alpha^{2k}(\sqrt{r})^{k}}{m^{\ell}} \; \leq \; \frac{f(k)^2(\mu \sigma_{\max})^{2k} r^k}{m^\ell d^k (\sqrt{r})^k},& \text{for }$ i=2, i' =4$\label{eq:ub413}\,,\\
\frac{f(k)^2d^\ell d^{k+1-s-\ell}\alpha^{2k}(\sqrt{r})^{s+k}}{m^{\ell}} \; \leq \; \frac{f(k)^2(\mu \sigma_{\max})^{2k} r^k}{m^\ell d^{k-1}/(\sqrt{r})^{k}},& \text{for }$ i=3, i' =4$\label{eq:ub411}\,,
\end{numcases}
Where \eqref{eq:ub42} and \eqref{eq:ub43} both use \eqref{eq:conj1}, and \eqref{eq:ub43} also uses \eqref{eq:ub415}. 
Note that $\ell$ is zero in \eqref{eq:ub414}. 
Collecting all the terms, and using Chebyshev's inequality, along with $\norm{M}_k^k \geq r(\sigma_{\min})^k $, we get the desired result.

For any two disjoint simple cycles $w\neq w' \in \cH_{k,1}$ with $|w \cap w'| =0$, we have the following
\begin{eqnarray}
\P\big[w \in \Omega\big] - \P\big[w \in \Omega \;\big|\; w' \in \Omega\big] & = & \frac{{m \choose k}}{{d \choose k}} - \frac{{m - k \choose k}}{{d-k \choose k}}
\;\leq \; \frac{m^k}{(d-k+1)^k} - \frac{(m-2k+1)^k}{(d-k)^k} \leq  \frac{f(k)m^{k-1}}{d^k}
\,,\nonumber\\
&& \label{eq:ub415}
\end{eqnarray}
where the last inequality assumes that $k < d/2$.
\begin{lemma}\label{conj:conj1}
For $k = 3$, and any $0 \leq \ell \leq \textcolor{black}{k}$
\begin{eqnarray}\label{eq:conj1}
\sum_{w \neq w' \in \cH_{k,1} : |w \cap w'| =\ell} \wgt_{M}(w) \wgt_{M}(w') & \leq & \frac{f(k)(\mu \sigma_{\max})^{2k}\max\{r^2,r^{\ell}\}}{d^\ell}\,.
\end{eqnarray}
\end{lemma}
Although we give a proof for $k=3$ only, we are intentionally writing the lemma for general $k$ as we expect the lemma holds for all $k \geq 3$. The joint walk $w \neq w' \in \cH_{k,1} : |w \cap w'| =\ell$ corresponds to $H(w) = D_{27}$, for $\ell =1$; and $H(w) = D_{23}$, for $\ell =2$ in Figure \ref{fig:6cyclic}. Define $\tilde M \equiv M - {\rm diag}(M)$,
and let $\odot$ denote the Hadamard product of two matrices. We have, 
\begin{eqnarray}\label{eq:conj2}
\sum_{w \neq w' \in \cH_{k,1} : |w \cap w'| =2} \wgt_{M}(w) \wgt_{M}(w') & = & (1/4) \sum_{i,j \in [d]}\Big(\big(\tilde M^2\odot \tilde M^2  - (\tilde M \odot \tilde M)^2 \big)\odot(\tilde M \odot \tilde M)\Big)_{i,j}\,.
\end{eqnarray}
Let's denote the quantity in \eqref{eq:conj2} by $C_1$, we have,
\begin{eqnarray}\label{eq:conj3}
\sum_{w \neq w' \in \cH_{k,1} : |w \cap w'| =1} \wgt_{M}(w) \wgt_{M}(w') & = & 
(1/8)\sum_{i \in [d]}\Big({\rm diag}( \tilde M^3) \odot {\rm diag}( \tilde M^3)  - 2{\rm diag}((\tilde M \odot \tilde M)^3)\Big)_i - 2 C_1\,.\nonumber\\
&&
\end{eqnarray}
It is easy to verify Equation \eqref{eq:conj1} for $k=3$ and $\ell \in \{1,2\}$ using the fact that $M$ is a $\mu$ incoherent symmetric matrix with its off-diagonals bounded by $\mu \sigma_{\max}(\sqrt{r}/d)$. 
For $\ell=0$, quantity in \eqref{eq:conj1} is the sum of each pair of disjoint triangles. For sum of all triangles, we have, 
\begin{eqnarray}\label{eq:conj4}
\sum_{w \in \cH_{k,1}} \wgt_{M}(w) = (1/6)\sum_{i \in [d]} \Big({\rm diag}(\tilde M^3)\Big)_i \; \leq \;(\mu \sigma_{\max})^3 r\,.
\end{eqnarray} 
Using Equations \eqref{eq:conj2}, \eqref{eq:conj3} and \eqref{eq:conj4}, bound for $\ell=0$ follows immediately. Bound for $\ell = k$, follows by using the fact that $M_{i,j} \leq \mu \sigma_{\max}(\sqrt{r}/d)$ for $i \neq j \in [d]$.

\end{document}

%% file: cycles3.tex
\tikzset{vertex/.style = {shape=circle,fill=black,inner sep=3pt,draw}}
\tikzset{every loop/.style={min distance=10mm,looseness=10}}
\tikzset{me/.style={to path={
\pgfextra{%
 \pgfmathsetmacro{\startf}{-(#1-1)/2}  
 \pgfmathsetmacro{\endf}{-\startf} 
 \pgfmathsetmacro{\stepf}{\startf+1}}
 \ifnum 1=#1 -- (\tikztotarget)  \else
\foreach \i in {\startf,\stepf,...,\endf}
    {%
     (\tikztostart)        parabola[bend pos=0.5, thin] bend +(0,0.3*\i)  (\tikztotarget)
      }
      \fi   
     \tikztonodes
      }}} 

\captionsetup[subfloat]{labelformat=empty}

\begin{figure}[h]
\centering
\begin{tikzpicture}
\def\fixsize{0.7}
\matrix [name=m,
cells={anchor=south},
column sep=0.1cm,row sep = 0.1cm]
{
\setcounter{subfigure}{7}
\node[scale=\fixsize] (a1) {\subcaptionbox{\large{$A_1$}}{\input {Tikz/A/e1.tikz}} }; &
\node[scale=\fixsize] (a2) {\subcaptionbox{\large{$A_2$}}{\input {Tikz/A/e2.tikz}} }; &
\node[scale=\fixsize] (a3) {\subcaptionbox{\large{$A_3$}}{\input {Tikz/A/e3.tikz}} }; \\
};
\end{tikzpicture}
\put(-170,-8){$c(A_1)=1$}
\put(-110,-8){$c(A_2)=3$}
\put(-50,-8){$c(A_3)=6$}
\caption{The 3-cyclic pseudographs $\cH_3=\{A_1,A_2,A_3\}$.}
\label{fig:3cyclic}
\end{figure}

%% file: cycles4.tex
\tikzset{vertex/.style = {shape=circle,fill=black,inner sep=3pt,draw}}
\tikzset{every loop/.style={min distance=10mm,looseness=10}}
\tikzset{me/.style={to path={
\pgfextra{%
 \pgfmathsetmacro{\startf}{-(#1-1)/2}  
 \pgfmathsetmacro{\endf}{-\startf} 
 \pgfmathsetmacro{\stepf}{\startf+1}}
 \ifnum 1=#1 -- (\tikztotarget)  \else
\foreach \i in {\startf,\stepf,...,\endf}
    {%
     (\tikztostart)        parabola[bend pos=0.5, thin] bend +(0,0.3*\i)  (\tikztotarget)
      }
      \fi   
     \tikztonodes
      }}} 

\captionsetup[subfloat]{labelformat=empty}

%

\begin{figure}
\centering
\begin{tikzpicture}
\def\fixsize{0.7}
\matrix [name=m,
cells={anchor=south},
column sep=0.1cm,row sep = 0.1cm]
{
\setcounter{subfigure}{7}
\node[scale=\fixsize] (b1) {\subcaptionbox{\large{$B_1$}}{\input {Tikz/B/e1.tikz}} }; &
\node[scale=\fixsize] (b2) {\subcaptionbox{\large{$B_2$}}{\input {Tikz/B/e2.tikz}} }; &
\node[scale=\fixsize] (b3) {\subcaptionbox{\large{$B_3$}}{\input {Tikz/B/e3.tikz}} }; &
\node[scale=\fixsize] (b4) {\subcaptionbox{\large{$B_4$}}{\input {Tikz/B/e4.tikz}} }; &
\node[scale=\fixsize] (b5) {\subcaptionbox{\large{$B_5$}}{\input {Tikz/B/e5.tikz}} }; &
\node[scale=\fixsize] (b6) {\subcaptionbox{\large{$B_6$}}{\input {Tikz/B/e6.tikz}} }; &
\node[scale=\fixsize] (b7) {\subcaptionbox{\large{$B_7$}}{\input {Tikz/B/e7.tikz}} };\\
};
\end{tikzpicture}

\caption{The 4-cyclic pseudographs $\cH_4$.}
\label{fig:4cyclic}
\end{figure}

%% file: cycles5.tex
\tikzset{vertex/.style = {shape=circle,fill=black,inner sep=3pt,draw}}
\tikzset{every loop/.style={min distance=10mm,looseness=10}}
\tikzset{me/.style={to path={
\pgfextra{%
 \pgfmathsetmacro{\startf}{-(#1-1)/2}  
 \pgfmathsetmacro{\endf}{-\startf} 
 \pgfmathsetmacro{\stepf}{\startf+1}}
 \ifnum 1=#1 -- (\tikztotarget)  \else
\foreach \i in {\startf,\stepf,...,\endf}
    {%
     (\tikztostart)        parabola[bend pos=0.5, thin] bend +(0,0.3*\i)  (\tikztotarget)
      }
      \fi   
     \tikztonodes
      }}} 

\captionsetup[subfloat]{labelformat=empty}

%

%

\begin{figure}
\centering
\begin{tikzpicture}
\def\fixsize{0.7}
\matrix [name=m,
cells={anchor=south},
column sep=0.1cm,row sep = 0.1cm]
{
\setcounter{subfigure}{12}
\node[scale=\fixsize] (c1) {\subcaptionbox{\large{$C_1$}}{\input {Tikz/C/e1.tikz}} }; &
\node[scale=\fixsize] (c2) {\subcaptionbox{\large{$C_2$}}{\input {Tikz/C/e2.tikz}} }; &
\node[scale=\fixsize] (c3) {\subcaptionbox{\large{$C_3$}}{\input {Tikz/C/e3.tikz}} }; &
\node[scale=\fixsize] (c4) {\subcaptionbox{\large{$C_4$}}{\input {Tikz/C/e4.tikz}} }; \\
\node[scale=\fixsize] (c5) {\subcaptionbox{\large{$C_5$}}{\input {Tikz/C/e5.tikz}} }; &
\node[scale=\fixsize] (c6) {\subcaptionbox{\large{$C_6$}}{\input {Tikz/C/e6.tikz}} }; &
\node[scale=\fixsize] (c7) {\subcaptionbox{\large{$C_7$}}{\input {Tikz/C/e7.tikz}} }; &
\node[scale=\fixsize] (c8) {\subcaptionbox{\large{$C_8$}}{\input {Tikz/C/e8.tikz}} };  \\
\node[scale=\fixsize] (c9) {\subcaptionbox{\large{$C_9$}}{\input {Tikz/C/e9.tikz}} }; &
\node[scale=\fixsize] (c10) {\subcaptionbox{\large{$C_{10}$}}{\input {Tikz/C/e10.tikz}} }; &
\node[scale=\fixsize] (c11) {\subcaptionbox{\large{$C_{11}$}}{\input {Tikz/C/e11.tikz}} }; &
\node[scale=\fixsize] (c12) {\subcaptionbox{\large{$C_{12}$}}{\input {Tikz/C/e12.tikz}} };  \\
};
\end{tikzpicture}

\caption{The 5-cyclic pseudographs $\cH_5$.}
\label{fig:5cyclic}
\end{figure}

%% file: cycles6.tex
\tikzset{vertex/.style = {shape=circle,fill=black,inner sep=3pt,draw}}
\tikzset{every loop/.style={min distance=10mm,looseness=10}}
\tikzset{me/.style={to path={
\pgfextra{%
 \pgfmathsetmacro{\startf}{-(#1-1)/2}  
 \pgfmathsetmacro{\endf}{-\startf} 
 \pgfmathsetmacro{\stepf}{\startf+1}}
 \ifnum 1=#1 -- (\tikztotarget)  \else
\foreach \i in {\startf,\stepf,...,\endf}
    {%
     (\tikztostart)        parabola[bend pos=0.5, thin] bend +(0,0.3*\i)  (\tikztotarget)
      }
      \fi   
     \tikztonodes
      }}} 

\captionsetup[subfloat]{labelformat=empty}

\begin{figure}
\centering
\begin{tikzpicture}
\def\fixsize{0.7}
\matrix [name=m,
cells={anchor=south},
column sep=0.1cm,row sep = 0.1cm]
{
\setcounter{subfigure}{36}
\node[scale=\fixsize] (d1) {\subcaptionbox{\large{$D_1$}}{\input {Tikz/D/e1.tikz}} }; &
\node[scale=\fixsize] (d2) {\subcaptionbox{\large{$D_2$}}{\input {Tikz/D/e2.tikz}} }; &
\node[scale=\fixsize] (d3) {\subcaptionbox{\large{$D_3$}}{\input {Tikz/D/e3.tikz}} }; &
\node[scale=\fixsize] (d4) {\subcaptionbox{\large{$D_4$}}{\input {Tikz/D/e4.tikz}} }; \\
\node[scale=\fixsize] (d5) {\subcaptionbox{\large{$D_5$}}{\input {Tikz/D/e5.tikz}} }; &
\node[scale=\fixsize] (d6) {\subcaptionbox{\large{$D_6$}}{\input {Tikz/D/e6.tikz}} }; &
\node[scale=\fixsize] (d7) {\subcaptionbox{\large{$D_7$}}{\input {Tikz/D/e7.tikz}} }; &
\node[scale=\fixsize] (d8) {\subcaptionbox{\large{$D_8$}}{\input {Tikz/D/e8.tikz}} };  \\
\node[scale=\fixsize] (d9) {\subcaptionbox{\large{$D_9$}}{\input {Tikz/D/e9.tikz}} }; &
\node[scale=\fixsize] (d10) {\subcaptionbox{\large{$D_{10}$}}{\input {Tikz/D/e10.tikz}} }; &
\node[scale=\fixsize] (d11) {\subcaptionbox{\large{$D_{11}$}}{\input {Tikz/D/e11.tikz}} }; &
\node[scale=\fixsize] (d12) {\subcaptionbox{\large{$D_{12}$}}{\input {Tikz/D/e12.tikz}} };  \\
\node[scale=\fixsize] (d13) {\subcaptionbox{\large{$D_{13}$}}{\input {Tikz/D/e13.tikz}} }; &
\node[scale=\fixsize] (d14) {\subcaptionbox{\large{$D_{14}$}}{\input {Tikz/D/e14.tikz}} }; &
\node[scale=\fixsize] (d15) {\subcaptionbox{\large{$D_{15}$}}{\input {Tikz/D/e15.tikz}} }; &
\node[scale=\fixsize] (d16) {\subcaptionbox{\large{$D_{16}$}}{\input {Tikz/D/e16.tikz}} }; \\
\node[scale=\fixsize] (d17) {\subcaptionbox{\large{$D_{17}$}}{\input {Tikz/D/e17.tikz}} }; &
\node[scale=\fixsize] (d18) {\subcaptionbox{\large{$D_{18}$}}{\input {Tikz/D/e18.tikz}} }; &
\node[scale=\fixsize] (d19) {\subcaptionbox{\large{$D_{19}$}}{\input {Tikz/D/e19.tikz}} }; &
\node[scale=\fixsize] (d20) {\subcaptionbox{\large{$D_{20}$}}{\input {Tikz/D/e20.tikz}} };  \\
\node[scale=\fixsize] (d21) {\subcaptionbox{\large{$D_{21}$}}{\input {Tikz/D/e21.tikz}} }; &
\node[scale=\fixsize] (d22) {\subcaptionbox{\large{$D_{22}$}}{\input {Tikz/D/e22.tikz}} }; &
\node[scale=\fixsize] (d23) {\subcaptionbox{\large{$D_{23}$}}{\input {Tikz/D/e23.tikz}} }; &
\node[scale=\fixsize] (d24) {\subcaptionbox{\large{$D_{24}$}}{\input {Tikz/D/e24.tikz}} };  \\
\node[scale=\fixsize] (d25) {\subcaptionbox{\large{$D_{25}$}}{\input {Tikz/D/e25.tikz}} }; &
\node[scale=\fixsize] (d26) {\subcaptionbox{\large{$D_{26}$}}{\input {Tikz/D/e26.tikz}} }; &
\node[scale=\fixsize] (d27) {\subcaptionbox{\large{$D_{27}$}}{\input {Tikz/D/e27.tikz}} }; &
\node[scale=\fixsize] (d28) {\subcaptionbox{\large{$D_{28}$}}{\input {Tikz/D/e28.tikz}} };  \\
\node[scale=\fixsize] (d29) {\subcaptionbox{\large{$D_{29}$}}{\input {Tikz/D/e29.tikz}} }; &
\node[scale=\fixsize] (d30) {\subcaptionbox{\large{$D_{30}$}}{\input {Tikz/D/e30.tikz}} }; &
\node[scale=\fixsize] (d31) {\subcaptionbox{\large{$D_{31}$}}{\input {Tikz/D/e31.tikz}} }; &
\node[scale=\fixsize] (d32) {\subcaptionbox{\large{$D_{32}$}}{\input {Tikz/D/e32.tikz}} };  \\
};
\end{tikzpicture}

\caption{The 6-cyclic pseudographs $\cH_6$.}
\label{fig:6cyclic}
\end{figure}

%% file: cycles7.tex
\tikzset{vertex/.style = {shape=circle,fill=black,inner sep=3pt,draw}}
\tikzset{every loop/.style={min distance=10mm,looseness=10}}
\tikzset{me/.style={to path={
\pgfextra{%
 \pgfmathsetmacro{\startf}{-(#1-1)/2}  
 \pgfmathsetmacro{\endf}{-\startf} 
 \pgfmathsetmacro{\stepf}{\startf+1}}
 \ifnum 1=#1 -- (\tikztotarget)  \else
\foreach \i in {\startf,\stepf,...,\endf}
    {%
     (\tikztostart)        parabola[bend pos=0.5, thin] bend +(0,0.3*\i)  (\tikztotarget)
      }
      \fi   
     \tikztonodes
      }}} 

\captionsetup[subfloat]{labelformat=empty}

\begin{figure}
\centering
\begin{tikzpicture}
\def\fixsize{0.7}
\matrix [name=m,
cells={anchor=south},
column sep=0.1cm,row sep = 0.1cm]
{
\setcounter{subfigure}{36}
\node[scale=\fixsize] (e1) {\subcaptionbox{\large{$E_1$}}{\input {Tikz/E/e1.tikz}} }; &
\node[scale=\fixsize] (e2) {\subcaptionbox{\large{$E_2$}}{\input {Tikz/E/e2.tikz}} }; &
\node[scale=\fixsize] (e3) {\subcaptionbox{\large{$E_3$}}{\input {Tikz/E/e3.tikz}} }; &
\node[scale=\fixsize] (e4) {\subcaptionbox{\large{$E_4$}}{\input {Tikz/E/e4.tikz}} }; \\
\node[scale=\fixsize] (e5) {\subcaptionbox{\large{$E_5$}}{\input {Tikz/E/e5.tikz}} }; &
\node[scale=\fixsize] (e6) {\subcaptionbox{\large{$E_6$}}{\input {Tikz/E/e6.tikz}} }; &
\node[scale=\fixsize] (e7) {\subcaptionbox{\large{$E_7$}}{\input {Tikz/E/e7.tikz}} }; &
\node[scale=\fixsize] (e8) {\subcaptionbox{\large{$E_8$}}{\input {Tikz/E/e8.tikz}} };  \\
\node[scale=\fixsize] (e9) {\subcaptionbox{\large{$E_9$}}{\input {Tikz/E/e9.tikz}} }; &
\node[scale=\fixsize] (e10) {\subcaptionbox{\large{$E_{10}$}}{\input {Tikz/E/e10.tikz}} }; &
\node[scale=\fixsize] (e11) {\subcaptionbox{\large{$E_{11}$}}{\input {Tikz/E/e11.tikz}} }; &
\node[scale=\fixsize] (e12) {\subcaptionbox{\large{$E_{12}$}}{\input {Tikz/E/e12.tikz}} };  \\
\node[scale=\fixsize] (e13) {\subcaptionbox{\large{$E_{13}$}}{\input {Tikz/E/e13.tikz}} }; &
\node[scale=\fixsize] (e14) {\subcaptionbox{\large{$E_{14}$}}{\input {Tikz/E/e14.tikz}} }; &
\node[scale=\fixsize] (e15) {\subcaptionbox{\large{$E_{15}$}}{\input {Tikz/E/e15.tikz}} }; &
\node[scale=\fixsize] (e16) {\subcaptionbox{\large{$E_{16}$}}{\input {Tikz/E/e16.tikz}} }; \\
\node[scale=\fixsize] (e17) {\subcaptionbox{\large{$E_{17}$}}{\input {Tikz/E/e17.tikz}} }; &
\node[scale=\fixsize] (e18) {\subcaptionbox{\large{$E_{18}$}}{\input {Tikz/E/e18.tikz}} }; &
\node[scale=\fixsize] (e19) {\subcaptionbox{\large{$E_{19}$}}{\input {Tikz/E/e19.tikz}} }; &
\node[scale=\fixsize] (e20) {\subcaptionbox{\large{$E_{20}$}}{\input {Tikz/E/e20.tikz}} };  \\
\node[scale=\fixsize] (e21) {\subcaptionbox{\large{$E_{21}$}}{\input {Tikz/E/e21.tikz}} }; &
\node[scale=\fixsize] (e22) {\subcaptionbox{\large{$E_{22}$}}{\input {Tikz/E/e22.tikz}} }; &
\node[scale=\fixsize] (e23) {\subcaptionbox{\large{$E_{23}$}}{\input {Tikz/E/e23.tikz}} }; &
\node[scale=\fixsize] (e24) {\subcaptionbox{\large{$E_{24}$}}{\input {Tikz/E/e24.tikz}} };  \\
\node[scale=\fixsize] (e25) {\subcaptionbox{\large{$E_{25}$}}{\input {Tikz/E/e25.tikz}} }; &
\node[scale=\fixsize] (e26) {\subcaptionbox{\large{$E_{26}$}}{\input {Tikz/E/e26.tikz}} }; &
\node[scale=\fixsize] (e27) {\subcaptionbox{\large{$E_{27}$}}{\input {Tikz/E/e27.tikz}} }; &
\node[scale=\fixsize] (e28) {\subcaptionbox{\large{$E_{28}$}}{\input {Tikz/E/e28.tikz}} };  \\
\node[scale=\fixsize] (e29) {\subcaptionbox{\large{$E_{29}$}}{\input {Tikz/E/e29.tikz}} }; &
\node[scale=\fixsize] (e30) {\subcaptionbox{\large{$E_{30}$}}{\input {Tikz/E/e30.tikz}} }; &
\node[scale=\fixsize] (e31) {\subcaptionbox{\large{$E_{31}$}}{\input {Tikz/E/e31.tikz}} }; &
\node[scale=\fixsize] (e32) {\subcaptionbox{\large{$E_{32}$}}{\input {Tikz/E/e32.tikz}} };  \\
};
\end{tikzpicture}

\caption{The 7-cyclic pseudographs $\cH_7$}
\end{figure}

\begin{figure}
\centering
\begin{tikzpicture}
\def\fixsize{0.7}
\matrix [name=m,
cells={anchor=south},
column sep=0.1cm,row sep = 0.1cm]
{
\setcounter{subfigure}{36}
\node[scale=\fixsize] (e33) {\subcaptionbox{\large{$E_{33}$}}{\input {Tikz/E/e33.tikz}} }; &
\node[scale=\fixsize] (e34) {\subcaptionbox{\large{$E_{34}$}}{\input {Tikz/E/e34.tikz}} }; &
\node[scale=\fixsize] (e35) {\subcaptionbox{\large{$E_{35}$}}{\input {Tikz/E/e35.tikz}} }; &
\node[scale=\fixsize] (e36) {\subcaptionbox{\large{$E_{36}$}}{\input {Tikz/E/e36.tikz}} };  \\
\node[scale=\fixsize] (e37) {\subcaptionbox{\large{$E_{37}$}}{\input {Tikz/E/e37.tikz}} }; &
\node[scale=\fixsize] (e38) {\subcaptionbox{\large{$E_{38}$}}{\input {Tikz/E/e38.tikz}} }; &
\node[scale=\fixsize] (e39) {\subcaptionbox{\large{$E_{39}$}}{\input {Tikz/E/e39.tikz}} }; &
\node[scale=\fixsize] (e40) {\subcaptionbox{\large{$E_{40}$}}{\input {Tikz/E/e40.tikz}} };  \\
\node[scale=\fixsize] (e41) {\subcaptionbox{\large{$E_{41}$}}{\input {Tikz/E/e41.tikz}} }; &
\node[scale=\fixsize] (e42) {\subcaptionbox{\large{$E_{42}$}}{\input {Tikz/E/e42.tikz}} }; &
\node[scale=\fixsize] (e43) {\subcaptionbox{\large{$E_{43}$}}{\input {Tikz/E/e43.tikz}} }; &
\node[scale=\fixsize] (e44) {\subcaptionbox{\large{$E_{44}$}}{\input {Tikz/E/e44.tikz}} };  \\
\node[scale=\fixsize] (e45) {\subcaptionbox{\large{$E_{45}$}}{\input {Tikz/E/e45.tikz}} }; &
\node[scale=\fixsize] (e46) {\subcaptionbox{\large{$E_{46}$}}{\input {Tikz/E/e46.tikz}} }; &
\node[scale=\fixsize] (e47) {\subcaptionbox{\large{$E_{47}$}}{\input {Tikz/E/e47.tikz}} }; &
\node[scale=\fixsize] (e48) {\subcaptionbox{\large{$E_{48}$}}{\input {Tikz/E/e48.tikz}} };  \\
\node[scale=\fixsize] (e49) {\subcaptionbox{\large{$E_{49}$}}{\input {Tikz/E/e49.tikz}} }; &
\node[scale=\fixsize] (e50) {\subcaptionbox{\large{$E_{50}$}}{\input {Tikz/E/e50.tikz}} }; &
\node[scale=\fixsize] (e51) {\subcaptionbox{\large{$E_{51}$}}{\input {Tikz/E/e51.tikz}} }; &
\node[scale=\fixsize] (e52) {\subcaptionbox{\large{$E_{52}$}}{\input {Tikz/E/e52.tikz}} };  \\
\node[scale=\fixsize] (e53) {\subcaptionbox{\large{$E_{53}$}}{\input {Tikz/E/e53.tikz}} }; &
\node[scale=\fixsize] (e54) {\subcaptionbox{\large{$E_{54}$}}{\input {Tikz/E/e54.tikz}} }; &
\node[scale=\fixsize] (e55) {\subcaptionbox{\large{$E_{55}$}}{\input {Tikz/E/e55.tikz}} }; &
\node[scale=\fixsize] (e56) {\subcaptionbox{\large{$E_{56}$}}{\input {Tikz/E/e56.tikz}} };  \\
\node[scale=\fixsize] (e57) {\subcaptionbox{\large{$E_{57}$}}{\input {Tikz/E/e57.tikz}} }; &
\node[scale=\fixsize] (e58) {\subcaptionbox{\large{$E_{58}$}}{\input {Tikz/E/e58.tikz}} }; &
\node[scale=\fixsize] (e59) {\subcaptionbox{\large{$E_{59}$}}{\input {Tikz/E/e59.tikz}} }; &
\node[scale=\fixsize] (e60) {\subcaptionbox{\large{$E_{60}$}}{\input {Tikz/E/e60.tikz}} };  \\
};
\end{tikzpicture}

\caption{The 7-cyclic pseudographs $\cH_7$}
\end{figure}

\begin{figure}
\centering
\begin{tikzpicture}
\def\fixsize{0.7}
\matrix [name=m,
cells={anchor=south},
column sep=0.1cm,row sep = 0.1cm]
{
\setcounter{subfigure}{36}
\node[scale=\fixsize] (e61) {\subcaptionbox{\large{${E_{61}}$}}{\input {Tikz/E/e61.tikz}} }; &
\node[scale=\fixsize] (e62) {\subcaptionbox{\large{$E_{62}$}}{\input {Tikz/E/e62.tikz}} }; &
\node[scale=\fixsize] (e63) {\subcaptionbox{\large{$E_{63}$}}{\input {Tikz/E/e63.tikz}} }; &
\node[scale=\fixsize] (e64) {\subcaptionbox{\large{$E_{64}$}}{\input {Tikz/E/e64.tikz}} }; &
\node[scale=\fixsize] (e65) {\subcaptionbox{\large{$E_{65}$}}{\input {Tikz/E/e65.tikz}} }; \\
\node[scale=\fixsize] (e66) {\subcaptionbox{\large{$E_{66}$}}{\input {Tikz/E/e66.tikz}} }; &
\node[scale=\fixsize] (e67) {\subcaptionbox{\large{$E_{67}$}}{\input {Tikz/E/e67.tikz}} }; &
\node[scale=\fixsize] (e68) {\subcaptionbox{\large{$E_{68}$}}{\input {Tikz/E/e68.tikz}} }; &
\node[scale=\fixsize] (e69) {\subcaptionbox{\large{$E_{69}$}}{\input {Tikz/E/e69.tikz}} };  \\
};
\end{tikzpicture}

\caption{The 7-cyclic pseudographs $\cH_7$.}
\label{fig:7cyclic}
\end{figure}


%

%% file: computation.tex
In this section we provide the complete matrix oeprations for copmuting $\gam_M(H)$'s. 
Equations \eqref{eq:matrix4start} - \eqref{eq:matrix4end} give expressions to compute $\gam_M(H)$ for $H \in \cH_4$ as labeled in Figure \ref{fig:4cyclic}.
Equations \eqref{eq:matrix5start} - \eqref{eq:matrix5end} give expressions to compute $\gam_M(H)$ for $H \in \cH_5$ as labeled in Figure \ref{fig:5cyclic}.
Equations \eqref{eq:matrix61start} - \eqref{eq:matrix62end} give expressions to compute $\gam_M(H)$ for $H \in \cH_6$ as labeled in Figure \ref{fig:6cyclic}.
Equations \eqref{eq:matrix71start} - \eqref{eq:matrix74end} give expressions to compute $\gam_M(H)$ for $H \in \cH_7$ as labeled in Figure \ref{fig:7cyclic}.

For brevity of notations and readability, we define the following additional notations. 
Let $A \hd B$ denote the Hadamard product. For $A \in \mathbb{R}^{d\times d}$, let $\sm(A)$ denote a vector $v \in \mathbb{R}^d$ such that $v_i = \sum_{j \in [d]} A_{i,j}$. With a slight abuse of notation, for $v \in \mathbb{R}^d$, let $\sm(v)$ denote sum of all elements of $v$ that is $\sm(v) = \sum_{i \in [d]} v_i$. Let $\sm(\gam_M(H_{i}): \gam_M(H_{j})) \equiv \sum_{i' = i}^j \gam_M(H_{i'})$.
Define $R \equiv \ones_{d\times d} - {\diag}(\ones_{d\times d} )$, that is $R$ is an all-ones matrix except on diagonals which are zeros. Further, for brevity, we omit the subscript $M$ from the notations $\gam_M(H), O_M$ and $D_M$.

\begin{eqnarray}
\val{B_1} &=& \sm(\sm(D \hd D \hd D \hd D)) \label{eq:matrix4start}\\
\val{B_2} &=& \sm(\sm(O \hd O \hd O \hd O))\\
\val{B_3} &=& 4\; \trace(O \mm O \mm D \mm D) \\
\val{B_4} &=& 2 \; \sm(\sm((O \hd O) \mm (O \hd O)\hd R))\\
\val{B_5} &=& 2\; \trace(O \mm D \mm O \mm D)\\
\val{B_6} &=& \trace(O\mm O \mm O \mm O) - \sm( \val{B_2}:\val{B_4}) \\ 
\val{B_7} &=& \trace(M \mm M \mm M \mm M) - \sm(\val{B_1}:\val{B_6}) 
\label{eq:matrix4end}
\end{eqnarray}
\begin{eqnarray}
        \val{C_1}  & = &\trace(D \hd D \hd D \hd D \hd D ) \label{eq:matrix5start}\\
        \val{C_2}  & = & 5\; \sm(\sm(D\mm O \hd O \hd O \hd O))\\
       \val{C_3}  & = & 5\; \sm(\sm((D\hd D \hd D) \mm (O\hd O)))\\
        \val{C_4}  & = & 5 \; \trace((O\hd O \hd O)\mm O \mm O)\\
        \val{C_5}  & = & 5\; \sm(\sm(D\mm (O \hd O) \mm (D \hd D)))\\
       \val{C_6}  & = &5 \;  \sm(\sm(((O\hd O ) \mm D\mm(O\hd O))\hd R))\\
        \val{C_7}  & = & 5 \; \sm(\sm((D \mm (O \hd O ) \mm (O \hd O))\hd R))\\
       \val{C_8}  & = & 5\; \trace(O\mm O \mm O\mm (D \hd D))\\
       \val{C_9}  & = &5 \; \sm(\diag(O \hd O \hd O) \hd \sm(O \hd O)) -10\; \trace((O\hd O \hd O) \mm O \mm O))\\
       \val{C_{10}}  & = & \trace(O\mm O \mm O \mm O \mm O) - \val{C_4} - \val{C_9}\\ 
       \val{C_{11}}  & = & 5\; \trace(O \mm D \mm O \mm D \mm O)\\
       \val{C_{12}}  & = & \trace(M \mm M \mm M \mm M \mm M) - \sm(\val{C_1}:\val{C_{11}}) \label{eq:matrix5end}
\end{eqnarray}
\begin{eqnarray} 
        \val{D_{1}} & = &  \sm(\sm(D \hd D \hd D \hd D \hd D \hd D))\label{eq:matrix61start}\\
        \val{D_{2}} & = &  \sm(\sm(O \hd O \hd O\hd O\hd O\hd O))\\
        \val{D_{3}} & = &  6\;  \sm(\sm(((O \hd O) \mm (O \hd O \hd O \hd O)) \hd R ))\\
        \val{D_{4}} & = &  6 \; \sm(\sm(((O \hd O ) \mm (D \hd D\hd D \hd D)) \hd R ))\\
        \val{D_{5}} & = &  9\;  \sm(\sm(((D \hd D) \mm (O \hd O \hd O \hd O)) \hd R ))\\
        \val{D_{6}} & = &  3\;  \sm(\sm(((D \hd D) \mm (O \hd O) \mm (D \hd D)) \hd R))\\
        \val{D_{7}} & = &  6\;  \sm(\sm(((D \hd D) \mm (O \hd O) \mm (O \hd O)) \hd R ))\\
        \val{D_{8}} & = &  9\;  \sm(\sm(((O \hd O) \mm (D \hd D) \mm (O\hd O)) \hd R))\\
        \val{D_{9}} & = &  6\;  \sm(\sm(((D \hd D \hd D) \mm (O \hd O) \mm D) \hd R))\\
        \val{D_{10}} & = &  6\;  \sm(\sm((D \mm (O \hd O \hd O \hd O) \mm D) \hd R ))\\
        \val{D_{11}} & = &  3\;  \sm\Big(\big(\sm(((O \hd O) \mm (O \hd O)) \hd R)\big) \hd \big(\sm(O \hd O)\big) - \sm\Big(\big((O \hd O \hd O \hd O) \mm (O \hd O)\big) \hd R\Big)\nonumber\\
        &&\;\; - \diag((O \hd O) \mm (O \hd O) \mm (O \hd O))\Big)\\
        \val{D_{12}} & = &  4\; \trace((O \hd O) \mm (O \hd O) \mm (O \hd O))\\
        \val{D_{13}} & = &  2\;  \sm\Big( (\sm(O \hd O))\hd (\sm(O \hd O))\hd (\sm(O \hd O)) - \sm((O \hd O \hd O \hd O \hd O \hd O)) \nonumber\\
        && \;\; - 3 \; \big( (\sm(O \hd O \hd O \hd O)) \hd (\sm(O \hd O))- (\sm(O \hd O \hd O \hd O \hd O \hd O) ) \big)\Big)\\
        \val{D_{14}} & = &  3\;  \sm(\sm((D \mm (O \hd O) \mm (O \hd O) \mm D) \hd R))\\
        \val{D_{15}} & = &  12\;  \sm(\sm((D \mm (O \hd O) \mm D \mm (O \hd O)) \hd R))\\
        \val{D_{16}} & = &  6\;  \sm\Big(\sm(((O \hd O \hd O) \mm O) \hd R \hd (O \mm O)) - \sm(((O \hd O \hd O \hd O) \mm (O \hd O)) \hd R)\Big)\\
        \val{D_{17}} & = &  6\; \trace((D \hd D \hd D) \mm O \mm O \mm O)\\
        \val{D_{18}} & = &  24\; \trace(D \mm (O \hd O \hd O) \mm O \mm O)\\
        \val{D_{19}} & = &  6 \; \trace(D \mm O \mm (O \hd O \hd O) \mm O)\\
        \val{D_{20}} & = &  6 \;\Big(\sm(\sm( (O\mm O) \hd ((O \mm (D \hd D) \mm O) \hd R )))  - \sm(\sm(((O \hd O) \mm (D \hd D) \mm (O \hd O)) \hd R))\Big)\nonumber\\ 
        &&       \\
        \val{D_{21}} & = &  12\; \trace(O \mm (D \hd D) \mm O \mm D \mm O)\\
        \val{D_{22}} & = &  6 \;\Big(\sm\Big(\sm\big(((O \mm O) \hd R \hd (O \mm O) - ((O \hd O) \mm (O \hd O)) \hd R)\big) \hd \sm(O \hd O) \Big)\nonumber\\
           && - 2\;  \sm\Big(\sm((((O \hd O \hd O) \mm O) \hd R \hd (O \mm O) - ((O \hd O \hd O \hd O) \mm (O \hd O)) \hd R))\Big)\nonumber\\
            &&- \sm\Big(\sm\Big( \big( (O \mm O) \hd R \hd (O \mm O) - ((O \hd O) \mm (O \hd O)) \hd R\big) \hd (O \hd O)\big)\Big)\Big)\Big)\\
        \val{D_{23}} & = &  9\;  \sm(\sm(((O \mm O) \hd R \hd (O \mm O) - ((O \hd O) \mm (O \hd O)) \hd R) \hd ((O \hd O))))\\ 
        \val{D_{24}} & = &  12\;  \sm(\diag(O \mm D \mm O \mm O) \hd \sm((O \hd O)) - \diag((O \hd O \hd O) \mm D \mm O \mm O)\nonumber\\
        && - \diag((O \hd O \hd O) \mm O \mm D \mm O))\\
        \val{D_{25}} & = &  6\;  \sm(\diag(O \mm O \mm O) \hd \sm((O \hd O) \mm D) - 2 \; \diag((O \hd O \hd O) \mm D \mm O \mm O))\\
        \val{D_{26}} & = &  12\;  \sm(\diag(O \mm O \mm O) \hd \diag(D) \hd \sm((O \hd O)) -  \diag((O \hd O \hd O) \mm O \mm O) \hd \diag(D))\\
        \val{D_{27}} & = &  3\;  \sm\Big(\diag(O \mm O \mm O)\hd \diag(O \mm O \mm O)  - 2\; \diag((O \hd O) \mm (O \hd O) \mm (O \hd O)) \Big) \nonumber\\
        && \;\; - (4/3) \val{D_{23}} \label{eq:matrix61end}
        \end{eqnarray}
        \begin{eqnarray}        
        \val{D_{28}} & = &  \trace(O \mm O \mm O \mm O \mm O \mm O) - \val{D_{2}} - \val{D_{3}}- \val{D_{11}} - \val{D_{12}} - \val{D_{13}} \nonumber\\
        && \;\; - \val{D_{16}} - \val{D_{22}} - \val{D_{23}} - \val{D_{27}} \label{eq:matrix62start}\\
        \val{D_{29}} & = &  2\; \trace(D \mm O \mm D \mm O \mm D \mm O)\\
        \val{D_{30}} & = &  3\;  \sm(\sm((O \mm D \mm O) \hd R \hd (O \mm D \mm O)) - \sm(((O \hd O) \mm (D \hd D) \mm (O \hd O)) \hd R))\nonumber\\
&&\\
        \val{D_{31}} & = &  6\;  \sm(\sm((O \mm D \mm O \mm D) \hd R \hd (O \mm O)) - \sm(((O \hd O) \mm D \mm (O \hd O) \mm D) \hd R))\\
        \val{D_{32}} & = &  \trace(M \mm M \mm M \mm M \mm M \mm M) - \trace(O \mm O \mm O \mm O \mm O \mm O) - \sm(\val{D_{1}}: \val{D_{26}}) + \val{D_{2}} + \val{D_{3}}+ \nonumber\\
        && \;\;\val{D_{11}} + \val{D_{12}} + \val{D_{13}}  + \val{D_{16}} + \val{D_{22}} + \val{D_{23}}  - \val{D_{29}} - \val{D_{30}} - \val{D_{31}} \label{eq:matrix62end}
\end{eqnarray}

\begin{eqnarray}    
      \val{E_{1}} & = & \sm(\diag((D \hd D \hd D \hd D \hd D \hd D \hd D)))\label{eq:matrix71start}\\
       \val{E_{2}} & = & 7\; \sm(\sm((O\hd O)\mm(D \hd D \hd D \hd D \hd D)))\\
       \val{E_{3}} & = & 7\; \sm(\sm(((D \hd D)\mm(O\hd O)\mm(D \hd D \hd D))\hd R))\\        
       \val{E_{4}} & = & 14\; \sm(\sm((O\hd O \hd O \hd O)\mm(D \hd D \hd D)))\\
       \val{E_{5}} & = & 7\; \sm(\sm((O\hd O \hd O \hd O \hd O \hd O)\mm D))\\
       \val{E_{6}} & = & 7\; \sm(\sm((D\mm(O\hd O)\mm(D \hd D \hd D \hd D))\hd R))\\
       \val{E_{7}} & = & 21\; \sm(\sm((D\mm(O\hd O \hd O \hd O)\mm(D \hd D))\hd R))\\
       \val{E_{8}} & = & 7\; \sm(\sm(((O\hd O)\mm(O\hd O)\mm(D \hd D \hd D))\hd R))\\
       \val{E_{9}} & = & 14\; \sm(\sm(((O\hd O)\mm(D \hd D \hd D)\mm(O\hd O))\hd R))\\
       \val{E_{10}} & = & 7\; \sm(\sm(((O\hd O \hd O \hd O)\mm(O\hd O)\mm D)\hd R))\\
       \val{E_{11}} & = & 21\; \sm(\sm(((O\hd O \hd O \hd O)\mm D\mm (O\hd O))\hd R))\\
       \val{E_{12}} & = & 14\; \sm(\sm((D\mm (O\hd O \hd O \hd O)\mm (O\hd O))\hd R))\\
       \val{E_{13}} & = & 7 \;\trace((O\hd O \hd O \hd O \hd O)\mm O\mm O)\\  
       \val{E_{14}} & = & 14\; \trace((O\hd O \hd O)\mm O\mm (O\hd O \hd O))\\ 
       \val{E_{15}} & = & 7\; \sm(\sm(((O\hd O)\mm (O\hd O))\hd R)\hd\sm((O\hd O)\mm D) - \sm(((O\hd O \hd O \hd O)\mm D\mm (O\hd O))\hd R) \nonumber\\
       && \;\; - \diag(((O\hd O)\mm D\mm (O\hd O)\mm (O\hd O))))\\
       \val{E_{16}} & = & 14\; \sm((\sm(((O\hd O)\mm (O\hd O))\hd R)\hd\sm((O\hd O)) - \sm(((O\hd O \hd O \hd O)\mm (O\hd O))\hd R) \nonumber\\
       && \;\; - \diag(((O\hd O)\mm (O\hd O)\mm (O\hd O))))\hd\diag(D))\\
       \val{E_{17}} & = & 7\; \sm(((\sm(O\hd O)\hd \sm(O\hd O) \hd \sm(O\hd O)) - \sm((O\hd O \hd O \hd O \hd O \hd O))\nonumber\\
       && \;\; - 3\;(\sm((O\hd O \hd O \hd O))\hd\sm((O\hd O))  -\sm((O\hd O \hd O \hd O \hd O \hd O))))\hd\diag(D))\\
        Z_{1} &\equiv & 0.5 \; ((\sm(O\hd O)\hd \sm(O\hd O)) - \sm((O\hd O \hd O \hd O)))\nonumber\\ 
       \val{E_{18}} & = & 14\; \sm(\sm((O\hd O)\mm D)\hd Z_{1} - \sm((O\hd O \hd O \hd O)\mm D)\hd\sm((O\hd O))  \nonumber\\
       && \;\;+ \; \sm((O\hd O \hd O \hd O \hd O \hd O)\mm D))\\
       \val{E_{19}} & = & 28\; \sm(\diag((O\hd O)\mm (O\hd O)\mm (O\hd O))\hd\diag(D))\\
       \val{E_{20}} & = & 21\; \sm(\sm((D\mm (O\hd O)\mm (D \hd D)\mm (O\hd O))\hd R))\\
       \val{E_{21}} & = & 14\; \sm(\sm(((D \hd D)\mm (O\hd O)\mm D\mm (O\hd O))\hd R))\\
       \val{E_{22}} & = & 7\; \sm(\sm((D\mm (O\hd O)\mm (O\hd O)\mm (D \hd D))\hd R))\\
       \val{E_{23}} & = & 7\; \sm(\diag(O\mm O\mm O)\hd\diag((D \hd D \hd D \hd D)))\\
       \val{E_{24}} & = & 28\; \sm(\diag((O\hd O \hd O)\mm O\mm O)\hd\sm((O\hd O)) - \diag((O\hd O \hd O \hd O \hd O)\mm O\mm O) \nonumber\\
       && \;\;  -\diag((O\hd O \hd O)\mm O\mm (O\hd O \hd O)))\\
       \val{E_{25}} & = & 7\; \sm(\diag(O\mm (O\hd O \hd O)\mm O)\hd\sm((O\hd O)) - 2\;\diag((O\hd O \hd O)\mm (O\hd O \hd O)\mm O))\\
       \val{E_{26}} & = & 7\; \sm(\diag(O\mm (O\hd O \hd O)\mm O)\hd\diag((D \hd D)))\\
       \val{E_{27}} & = & 42\; \sm(\diag((O\hd O \hd O)\mm O\mm O)\hd\diag((D \hd D)))\\
       \val{E_{28}} & = & 7\; \sm(\diag(O\mm O\mm O)\hd\sm((O\hd O \hd O \hd O)) - 2\;\diag((O\hd O \hd O \hd O \hd O)\mm O\mm O))\\
       \val{E_{29}} & = & 7\; \sm(\sm((D\mm (O\hd O)\mm D\mm (O\hd O)\mm D)\hd R))\\
       \val{E_{30}} & = & 28\; \sm(\diag(O\mm D\mm (O\hd O \hd O)\mm O)\hd\diag(D))\label{eq:matrix71end} 
       \end{eqnarray}       
       \begin{eqnarray} 
       \val{E_{31}} & = & 28\;\trace(O\mm D\mm (O\hd O \hd O)\mm D\mm O) \label{eq:matrix72start}\\ 
       \val{E_{32}} & = & 14\; \sm(\diag(O\mm (D \hd D)\mm O\mm O)\hd\sm((O\hd O)) - \diag((O\hd O \hd O)\mm O\mm (D \hd D)\mm O) \nonumber\\
       && \;\;  - \diag((O\hd O \hd O)\mm (D \hd D)\mm O\mm O))\\      
       \val{E_{33}} & = & 14\; \sm(\diag(O\mm D\mm O\mm O)\hd\diag((D \hd D \hd D)))\\ 
       \val{E_{34}} & = & 7 \trace(O\mm (D \hd D)\mm O\mm (D \hd D)\mm O)\\ 
       \val{E_{35}} & = & 7 (\sm(\sm((((O \mm O)\hd R )\hd ((O\mm (D \hd D \hd D)\mm O)\hd R )))) \nonumber\\
       && \;\; - \sm(\sm(((O\hd O)\mm (D \hd D \hd D)\mm (O\hd O))\hd R )))\\ 
       \val{E_{36}} & = & 14\; \sm(\sm(((O\hd O \hd O)\mm O)\hd R \hd (O\mm D\mm O))\nonumber\\
       && \;\;  - \sm(((O\hd O \hd O \hd O)\mm D\mm (O\hd O))\hd R))\\ 
       \val{E_{37}} & = & 28\; \sm(\sm(((O\hd O \hd O)\mm D\mm O)\hd R \hd (O\mm O)) \nonumber\\
       && \;\; - \sm(((O\hd O \hd O \hd O)\mm D\mm (O\hd O))\hd R))\\ 
        Z_{2} & \equiv & (((O\mm O)\hd R )\mm O - O\hd (\ones_{d\times 1}\mm (\sm((O\hd O)^\top))^\top- (O\hd O)))\hd R\\ 
        Z_{3} & \equiv & (O\hd ((O\mm O)\hd R ))\hd R \\ 
        Z_{4} & \equiv & (O\hd (((O\hd O \hd O \hd O \hd O)\mm O)\hd R ))\hd R\\
        Z_{6} & \equiv & ((O\hd O \hd O)\hd ((O\mm O)\hd R ))\hd R\\ 
        Z_{7} & \equiv & (O\hd (((O\hd O \hd O)\mm (O\hd O \hd O))\hd R ))\hd R\\
       \val{E_{38}} & = & 7\; \sm(\sm((((O\hd O \hd O) \mm O)\hd R \hd Z_{2} - (((O\hd O \hd O \hd O) \mm Z_{3})\hd R  - Z_{4}) \nonumber\\
       && \;\;  - ((Z_{6}\mm (O\hd O))\hd R  - Z_{7}))))\\ 
        Z_{7} & \equiv & 0.5\; \sm(\sm(O\hd (((O\hd O)\mm (O\hd O))\hd R )\hd ((O\mm O)\hd R ) \nonumber\\
       && \;\;  - O\hd (((O\hd O \hd O)\mm (O\hd O \hd O))\hd R )))\\
       \val{E_{39}} & = & 7\;(\sm(\sm((O\hd ((O\mm O)\hd R )\hd (\sm((O\hd O))\mm \ones_{1\times d}\nonumber\\
       && \;\;  - (O\hd O))\hd (\ones_{d\times 1}\mm (\sm((O\hd O)^\top))^\top - (O\hd O)))))
             \nonumber\\
       && \;\;- \sm(\sm((O\hd (((O\hd O \hd O)\mm O)\hd R )\hd (\ones_{d\times 1}\mm (\sm((O\hd O)^\top))^\top   - (O\hd O)))))
           \nonumber\\
       && \;\; - \sm(\sm((O\hd ((O\mm (O\hd O \hd O))\hd R )\hd (\sm((O\hd O))\mm \ones_{1\times d} - (O\hd O)))))
         \nonumber\\
       && \;\;   + \sm(\sm((O\hd (((O\hd O \hd O)\mm (O\hd O \hd O))\hd R ))))) -14\;Z_{7}\\
       \val{E_{40}} & = & 21\; \sm(\diag((D \hd D)\mm O\mm O\mm O)\hd \sm((O\hd O)) - 2\;\diag((D \hd D)\mm (O\hd O \hd O)\mm O\mm O))\\
       \val{E_{41}} & = & 7\; \sm(\diag(O\mm O\mm O)\hd \sm((O\hd O)\mm (D \hd D)) - 2\;\diag((O\hd O \hd O)\mm (D \hd D)\mm O\mm O))\\
       \val{E_{42}} & = & 7\;(\sm(\diag(O\mm O\mm O)\hd \sm(((O\hd O)\mm (O\hd O))\hd R) - 2\;\diag((O\hd O \hd O)\mm (O\hd O \hd O)\mm O))\nonumber\\
       && \;\;
            - 2\; \sm(\diag((O\hd O \hd O)\mm O\mm O)\hd \sm((O\hd O)) - \diag((O\hd O \hd O \hd O \hd O)\mm O\mm O) \nonumber\\
       && \;\;- \diag((O\hd O \hd O)\mm O\mm (O\hd O \hd O)))) - 28\;Z_{7}\\
       \val{E_{43}} & = & 14\; \sm(\diag(O\mm O\mm O)\hd Z_{1} - 2\;(\diag((O\hd O \hd O)\mm O\mm O)\hd \sm((O\hd O)) \nonumber\\
       && \;\; - \diag((O\hd O \hd O \hd O \hd O)\mm O\mm O) - 0.5\;\diag((O\hd O \hd O)\mm O\mm (O\hd O \hd O))))\\ 
       \val{E_{44}} & = & 56\;Z_{7}\\
        Z_{8} & \equiv & (O\hd (((O\hd O \hd O)\mm O)\hd R ))\hd R\\
         Z_{9} & \equiv & (O \hd ((O\mm O)\hd R ))\hd R \\
          Z_{10} & \equiv & (O\hd ((O \mm (O\hd O \hd O))\hd R ))\hd R \\
        Z_{11} & \equiv & ((O\mm O)\hd R \hd Z_{2} - (((O\hd O)\mm Z_{3})\hd R  - Z_{8}) - ((Z_{9}\mm (O\hd O))\hd R  - Z_{10}))\\ 
       \val{E_{45}} & = & 14\;(\sm(0.5\; \sm(Z_{11})\hd \sm((O\hd O)))  - (1/7)\;\val{E_{38}} - \sm(\sm(((O\hd O))\hd Z_{11})))\label{eq:matrix72end} 
              \end{eqnarray}
              \begin{eqnarray}
                     \val{E_{46}} & = & 21\; \sm(\sm(((O\hd O))\hd Z_{11})) \label{eq:matrix73start}\\ 
       \val{E_{47}} & = & 7\; \sm(\sm(Z_{11})\hd \diag((D \hd D)))\\ 
       \val{E_{48}} & = & 7\;\trace((D \hd D)\mm O\mm D\mm O\mm D\mm O)\\ 
       \val{E_{49}} & = & 14\; \sm(\diag(D\mm O\mm O\mm O)\hd \sm((O\hd O)\mm D) - 2\;\diag(D\mm (O\hd O \hd O)\mm D\mm O\mm O))\\
       \val{E_{50}} & = & 14\; \sm(\diag(O\mm O\mm D\mm O)\hd \sm((O\hd O)\mm D) - \diag((O\hd O \hd O)\mm D\mm O\mm D\mm O) \nonumber\\
       && \;\;- \diag((O\hd O \hd O)\mm (D \hd D)\mm O\mm O))\\
       \val{E_{51}} & = & 28\; \sm(\diag(D\mm O\mm D\mm O\mm O)\hd \sm((O\hd O)) - \diag(D\mm (O\hd O \hd O)\mm D\mm O\mm O) \nonumber\\
       && \;\;- \diag(D\mm (O\hd O \hd O)\mm O\mm D\mm O))\\
       \val{E_{52}} & = & 7\; \sm(\diag(O\mm D\mm O\mm D\mm O)\hd \sm((O\hd O)) - 2\;\diag((O\hd O \hd O)\mm D\mm O\mm D\mm O))\\
       \val{E_{53}} & = & 14\; \sm((\sm(((((O \mm O)\hd R )\hd ((O\mm D\mm O)\hd R ))  \nonumber\\
       && \;\; - ((O\hd O)\mm D\mm (O\hd O))\hd R )))\hd \diag((D \hd D)))\\ 
       \val{E_{54}} & = & 7\;\sm(\sm(((((O\mm D\mm O)\hd R )\hd ((O\mm (D \hd D)\mm O)\hd R ))  \nonumber\\
       && \;\; - ((O\hd O)\mm (D \hd D \hd D)\mm (O\hd O))\hd R )))\\ 
        Z_{12} & \equiv & \sm(0.5\; \sm(((((O \mm O)\hd R )\hd ((O \mm O)\hd R )) \nonumber\\
       && \;\;- (((O\hd O)\mm (O\hd O))\hd R ))\hd ((O\hd O)\mm D)))\\
        Z_{13} & \equiv & \sm(\sm((((((O\hd O \hd O)\mm D\mm O)\hd R )\hd ((O\mm O)\hd R )) \nonumber\\
       && \;\;  - ((O\hd O \hd O \hd O)\mm D\mm (O\hd O))\hd R )))\\ 
        Z_{14} & \equiv & 0.5\; \sm(\sm(((((O\mm D\mm O)\hd R )\hd ((O \mm O)\hd R ))  \nonumber\\
       && \;\; - (((O\hd O)\mm D\mm (O\hd O))\hd R ))\hd ((O\hd O))))\\
        Z_{15} & \equiv & \sm(\sm((((((O\hd O \hd O)\mm O)\hd R )\hd ((O\mm D\mm O)\hd R ))  \nonumber\\
       && \;\;  -((O\hd O \hd O \hd O)\mm D\mm (O\hd O))\hd R )))\\ 
       \val{E_{55}} & = & 14\;(\sm(0.5\;(\sm(((((O \mm O)\hd R )\hd (((O \mm O))\hd R ))  \nonumber\\
       && \;\; - ((O\hd O)\mm (O\hd O))\hd R )))\hd \sm((O\hd O)\mm D)) -Z_{13} - Z_{12})\\
       \val{E_{56}} & = & 28\;(\sm(0.5\;(\sm(((((O \mm O)\hd R )\hd (((O \mm O))\hd R ))\nonumber\\
       && \;\; - ((O\hd O)\mm (O\hd O))\hd R )))\hd \sm(D\mm (O\hd O))) -Z_{13} - Z_{12})\\
       \val{E_{57}} & = & 14\;(\sm((\sm(((((O\mm D\mm O)\hd R )\hd (((O \mm O))\hd R ))  \nonumber\\
       && \;\; - ((O\hd O)\mm D\mm (O\hd O))\hd R )))\hd \sm((O\hd O)))
            -Z_{13} - Z_{15} -2\;Z_{14})\\
       \val{E_{58}} & = & 14\;(\sm(0.5\; \sm((((((O \mm O)\hd R )\hd ((O \mm O)\hd R ))  \nonumber\\
       && \;\; - (((O\hd O)\mm (O\hd O))\hd R ))\mm D))\hd \sm((O\hd O))) - Z_{15} - Z_{12})\\
       \val{E_{59}} & = & 84\;Z_{12}\\
       \val{E_{60}} & = & 42\;Z_{14}\\
        Z_{25} &=& \trace(M \mm M \mm M\mm M\mm M\mm M\mm M) - \sm(\val{E_1}: \val{E_{60}})\\
        Z_{26} &=& \trace(O \mm O\mm O\mm O\mm O\mm O\mm O) - \val{E_{13}} - \val{E_{14}}- \val{E_{24}} \nonumber\\
       && \;\; - \val{E_{25}}- \val{E_{28}}- \val{E_{38}}- \val{E_{39}} - \sm(\val{E_{42}}: \val{E_{46}}) \\ 
        Z_{16} &\equiv & (1/6)\;((O\mm O\hd R )\hd (O\mm O\hd R )\hd (O\mm O\hd R )   -((O\hd O \hd O)\mm (O\hd O \hd O)\hd R ) \nonumber\\
       && \;\;  -3\;(((O\hd O)\mm (O\hd O)\hd R )\hd (O\mm O\hd R ) - ((O\hd O \hd O)\mm (O\hd O \hd O)\hd R )))\label{eq:matrix73end}
              \end{eqnarray}
              \begin{eqnarray}
       \val{E_{61}} & = & 42\; \sm(\sm(Z_{16}\hd O)) \label{eq:matrix74start}\\
        Z_{17} &\equiv & \sm(\sm(0.5\;((O\mm O\hd R )\hd (O\mm O\hd R )\nonumber\\
       && \;\; - ((O\hd O)\mm (O\hd O)\hd R )))\hd (0.5\;\diag(O\mm O\mm O)))\\
       \val{E_{62}} & = & 28\;(Z_{17} - (6/84)\;\val{E_{61}} - (2/42)\;\val{E_{46}}- (3/56)\;\val{E_{44}}\\
       \val{E_{63}} & = & Z_{26} -\val{E_{61}} -\val{E_{62}}\\
       \val{E_{64}} & = & 7\; \sm(\sm((D\mm O\mm D\mm O\mm D\hd R )\hd (O\mm O\hd R ))) \nonumber\\
       && \;\;- 7\; \sm(\sm(D\mm (O\hd O)\mm D\mm (O\hd O)\mm D\hd R))\\
       \val{E_{65}} & = & 7\; \sm(\sm(D\mm Z_{11}\mm D))\\
        Z_{18} &\equiv & \sm(((O\mm O)\hd R \hd (O\mm O) - ((O\hd O)\mm (O\hd O))\hd R )\hd ((O\hd O)))\\
       \val{E_{66}} & = & 7\; \sm(((\diag(O\mm O\mm O)\hd \diag(O\mm O\mm O)) \nonumber\\
       && \;\; - 2\;\diag((O\hd O)\mm (O\hd O)\mm (O\hd O)) - 4\;Z_{18})\hd \diag(D))\\
        Z_{20} &\equiv & 0.5\; \sm(\sm(((O\mm O\hd R )\hd (O\mm D\mm O\hd R ) - ((O\hd O)\mm D\mm (O\hd O)))\hd (O\hd O)))\\
       \val{E_{67}} & = & 14\;(\sm(\diag(O\mm O\mm O)\hd \diag(O\mm O\mm D\mm O) - 2\;\diag((O\hd O)\mm (O\hd O)\mm D\mm (O\hd O)))\nonumber\\
       && \;\; - 2\; \sm(Z_{18}\hd \diag(D)) - 4\;Z_{20})\\
        Z_{21} &\equiv & (((O\mm D\mm O\mm D)\hd R )\mm O - O\hd (\ones_{d\times 1}\mm \sm(D\mm (O\hd O)\mm D,1)\nonumber\\
       && \;\;- D\mm (O\hd O)\mm D))\hd R\\ 
        Z_{22} &\equiv & (O\hd ((O\mm D\mm O)\hd R ))\hd R \\ 
        Z_{23} &\equiv & (O\hd ((D\mm (O\hd O \hd O)\mm D\mm O)\hd R ))\hd R \\
          Z_{24} &\equiv & (O\hd ((O\mm D\mm (O\hd O \hd O)\mm D)\hd R ))\hd R \\
       \val{E_{68}} & = & 7\; \sm(\sm(((O\mm O)\hd R \hd Z_{21} - (((O\hd O)\mm D\mm Z_{22})\hd R  - Z_{23}) \nonumber\\
       && \;\;- ((Z_{22}\mm D\mm (O\hd O))\hd R  - Z_{24}))))\\ 
       \val{E_{69}} & = & Z_{25} - Z_{26} - \sm(\val{E_{64}}: \val{E_{68}})\label{eq:matrix74end}
\end{eqnarray}

%% file: schatten.bbl
\begin{thebibliography}{61}
\providecommand{\natexlab}[1]{#1}
\providecommand{\url}[1]{\texttt{#1}}
\expandafter\ifx\csname urlstyle\endcsname\relax
  \providecommand{\doi}[1]{doi: #1}\else
  \providecommand{\doi}{doi: \begingroup \urlstyle{rm}\Url}\fi

\bibitem[Achlioptas and McSherry(2001)]{AM01}
Dimitris Achlioptas and Frank McSherry.
\newblock Fast computation of low rank matrix approximations.
\newblock In \emph{Proceedings of the thirty-third annual ACM symposium on
  Theory of computing}, pages 611--618. ACM, 2001.

\bibitem[Alon et~al.(1997)Alon, Yuster, and Zwick]{AYZ97}
N.~Alon, R.~Yuster, and U.~Zwick.
\newblock Finding and counting given length cycles.
\newblock \emph{Algorithmica}, 17\penalty0 (3):\penalty0 209--223, 1997.

\bibitem[Andoni et~al.(2013)Andoni, Nguy{\^e}n, Polyanskiy, and
  Wu]{andoni2013tight}
A.~Andoni, H.~L. Nguy{\^e}n, Y.~Polyanskiy, and Y.~Wu.
\newblock Tight lower bound for linear sketches of moments.
\newblock In \emph{International Colloquium on Automata, Languages, and
  Programming}, pages 25--32. Springer, 2013.

\bibitem[Aune et~al.(2014)Aune, Simpson, and Eidsvik]{ASE14}
E.~Aune, D.~P. Simpson, and J.~Eidsvik.
\newblock Parameter estimation in high dimensional gaussian distributions.
\newblock \emph{Statistics and Computing}, 24\penalty0 (2):\penalty0 247--263,
  2014.

\bibitem[Avron and Toledo(2011)]{AT11}
H.~Avron and S.~Toledo.
\newblock Randomized algorithms for estimating the trace of an implicit
  symmetric positive semi-definite matrix.
\newblock \emph{Journal of the ACM (JACM)}, 58\penalty0 (2):\penalty0 8, 2011.

\bibitem[Bhojanapalli and Jain(2014)]{BJ14}
S.~Bhojanapalli and P.~Jain.
\newblock Universal matrix completion.
\newblock In \emph{Proceedings of the 31st International Conference on Machine
  Learning (ICML-14)}, pages 1881--1889, 2014.

\bibitem[Boutsidis et~al.(2015)Boutsidis, Drineas, Kambadur, Kontopoulou, and
  Zouzias]{BDK15}
C.~Boutsidis, P.~Drineas, P.~Kambadur, E.-M. Kontopoulou, and A.~Zouzias.
\newblock A randomized algorithm for approximating the log determinant of a
  symmetric positive definite matrix.
\newblock \emph{arXiv preprint arXiv:1503.00374}, 2015.

\bibitem[Cand\`es and Recht(2009)]{CR09}
E.~J. Cand\`es and B.~Recht.
\newblock Exact matrix completion via convex optimization.
\newblock \emph{Foundations of Computational Mathematics}, 9\penalty0
  (6):\penalty0 717--772, 2009.

\bibitem[Cand{\`e}s et~al.(2011)Cand{\`e}s, Li, Ma, and Wright]{rpca1}
Emmanuel~J Cand{\`e}s, Xiaodong Li, Yi~Ma, and John Wright.
\newblock Robust principal component analysis?
\newblock \emph{Journal of the ACM (JACM)}, 58\penalty0 (3):\penalty0 11, 2011.

\bibitem[Carb{\'o}-Dorca(2008)]{Car08}
R.~Carb{\'o}-Dorca.
\newblock Smooth function topological structure descriptors based on
  graph-spectra.
\newblock \emph{Journal of Mathematical Chemistry}, 44\penalty0 (2):\penalty0
  373--378, 2008.

\bibitem[Chen(2016)]{Che16}
J.~Chen.
\newblock How accurately should i compute implicit matrix-vector products when
  applying the hutchinson trace estimator?
\newblock \emph{SIAM Journal on Scientific Computing}, 38\penalty0
  (6):\penalty0 A3515--A3539, 2016.

\bibitem[Davis et~al.(2007)Davis, Kulis, Jain, Sra, and Dhillon]{DKJ07}
J.~V. Davis, B.~Kulis, P.~Jain, S.~Sra, and I.~S. Dhillon.
\newblock Information-theoretic metric learning.
\newblock In \emph{Proceedings of the 24th international conference on Machine
  learning}, pages 209--216. ACM, 2007.

\bibitem[Di~Napoli et~al.(2016)Di~Napoli, Polizzi, and Saad]{DPS16}
E.~Di~Napoli, E.~Polizzi, and Y.~Saad.
\newblock Efficient estimation of eigenvalue counts in an interval.
\newblock \emph{Numerical Linear Algebra with Applications}, 2016.

\bibitem[Elenberg et~al.(2015)Elenberg, Shanmugam, Borokhovich, and
  Dimakis]{elenberg2015beyond}
E.~R. Elenberg, K.~Shanmugam, M.~Borokhovich, and A.~G. Dimakis.
\newblock Beyond triangles: A distributed framework for estimating 3-profiles
  of large graphs.
\newblock In \emph{Proceedings of the 21th ACM SIGKDD International Conference
  on Knowledge Discovery and Data Mining}, pages 229--238. ACM, 2015.

\bibitem[Elenberg et~al.(2016)Elenberg, Shanmugam, Borokhovich, and
  Dimakis]{elenberg2016distributed}
E.~R. Elenberg, K.~Shanmugam, M.~Borokhovich, and A.~G. Dimakis.
\newblock Distributed estimation of graph 4-profiles.
\newblock In \emph{Proceedings of the 25th International Conference on World
  Wide Web}, pages 483--493. International World Wide Web Conferences Steering
  Committee, 2016.

\bibitem[Estrada(2000)]{Est00}
E.~Estrada.
\newblock Characterization of 3d molecular structure.
\newblock \emph{Chemical Physics Letters}, 319\penalty0 (5):\penalty0 713--718,
  2000.

\bibitem[Estrada and Hatano(2007)]{EH07}
E.~Estrada and N.~Hatano.
\newblock Statistical-mechanical approach to subgraph centrality in complex
  networks.
\newblock \emph{Chemical Physics Letters}, 439\penalty0 (1):\penalty0 247--251,
  2007.

\bibitem[Estrada and Rodriguez-Vel{\'a}zquez(2005)]{ER05}
E.~Estrada and J.~A. Rodriguez-Vel{\'a}zquez.
\newblock Spectral measures of bipartivity in complex networks.
\newblock \emph{Physical Review E}, 72\penalty0 (4):\penalty0 046105, 2005.

\bibitem[Feige and Ofek(2005)]{FeO05}
U.~Feige and E.~Ofek.
\newblock Spectral techniques applied to sparse random graphs.
\newblock \emph{Random Struct. Algorithms}, 27\penalty0 (2):\penalty0 251--275,
  2005.

\bibitem[Flum and Grohe(2004)]{FG04}
J.~Flum and M.~Grohe.
\newblock The parameterized complexity of counting problems.
\newblock \emph{SIAM Journal on Computing}, 33\penalty0 (4):\penalty0 892--922,
  2004.

\bibitem[Friedman et~al.(1989)Friedman, Kahn, and Szemer{\'e}di]{FKS89}
J.~Friedman, J.~Kahn, and E.~Szemer{\'e}di.
\newblock On the second eigenvalue in random regular graphs.
\newblock In \emph{Proceedings of the Twenty-First Annual ACM Symposium on
  Theory of Computing}, pages 587--598, Seattle, Washington, USA, may 1989.
  ACM.

\bibitem[G{\"o}tze and Tikhomirov()]{gotzerate}
F.~G{\"o}tze and A.~Tikhomirov.
\newblock On the rate of convergence to the semicircular law, preprint (2011).
\newblock \emph{arXiv preprint arXiv:1109.0611}.

\bibitem[Halford and Chugg(2006)]{HC06}
T.~R. Halford and K.~M. Chugg.
\newblock An algorithm for counting short cycles in bipartite graphs.
\newblock \emph{IEEE Transactions on Information Theory}, 52\penalty0
  (1):\penalty0 287--292, 2006.

\bibitem[Halko et~al.(2011)Halko, Martinsson, and Tropp]{rand2}
Nathan Halko, Per-Gunnar Martinsson, and Joel~A Tropp.
\newblock Finding structure with randomness: Probabilistic algorithms for
  constructing approximate matrix decompositions.
\newblock \emph{SIAM review}, 53\penalty0 (2):\penalty0 217--288, 2011.

\bibitem[Han et~al.(2015)Han, Malioutov, and Shin]{HMS15}
I.~Han, D.~Malioutov, and J.~Shin.
\newblock Large-scale log-determinant computation through stochastic chebyshev
  expansions.
\newblock In \emph{ICML}, pages 908--917, 2015.

\bibitem[Han et~al.(2016)Han, Malioutov, Avron, and Shin]{HMAS16}
I.~Han, D.~Malioutov, H.~Avron, and J.~Shin.
\newblock Approximating the spectral sums of large-scale matrices using
  chebyshev approximations.
\newblock \emph{arXiv preprint arXiv:1606.00942}, 2016.

\bibitem[Hutchinson(1990)]{Hut90}
Michael~F Hutchinson.
\newblock A stochastic estimator of the trace of the influence matrix for
  laplacian smoothing splines.
\newblock \emph{Communications in Statistics-Simulation and Computation},
  19\penalty0 (2):\penalty0 433--450, 1990.

\bibitem[Ingster and Suslina(2012)]{IS12}
Yuri Ingster and Irina~A Suslina.
\newblock \emph{Nonparametric goodness-of-fit testing under Gaussian models},
  volume 169.
\newblock Springer Science \& Business Media, 2012.

\bibitem[Jain et~al.(2013)Jain, Netrapalli, and Sanghavi]{JNS13}
P.~Jain, P.~Netrapalli, and S.~Sanghavi.
\newblock Low-rank matrix completion using alternating minimization.
\newblock In \emph{STOC}, pages 665--674, 2013.

\bibitem[Karimi and Banihashemi(2013)]{KB13}
M.~Karimi and A.~H. Banihashemi.
\newblock Message-passing algorithms for counting short cycles in a graph.
\newblock \emph{IEEE Transactions on Communications}, 61\penalty0 (2):\penalty0
  485--495, 2013.

\bibitem[Keshavan and Oh(2009)]{keshavan2009gradient}
R.~H. Keshavan and S.~Oh.
\newblock A gradient descent algorithm on the {G}rassman manifold for matrix
  completion.
\newblock \emph{arXiv preprint arXiv:0910.5260}, 2009.

\bibitem[Keshavan et~al.(2010{\natexlab{a}})Keshavan, Montanari, and
  Oh]{KMO10IT}
R.~H. Keshavan, A.~Montanari, and S.~Oh.
\newblock Matrix completion from a few entries.
\newblock \emph{Information Theory, IEEE Transactions on}, 56\penalty0
  (6):\penalty0 2980--2998, 2010{\natexlab{a}}.

\bibitem[Keshavan et~al.(2010{\natexlab{b}})Keshavan, Montanari, and
  Oh]{KMO10JMLR}
R.~H Keshavan, A.~Montanari, and S.~Oh.
\newblock Matrix completion from noisy entries.
\newblock \emph{Journal of Machine Learning Research}, 11\penalty0
  (2057-2078):\penalty0 1, 2010{\natexlab{b}}.

\bibitem[Kloks et~al.(2000)Kloks, Kratsch, and M{\"u}ller]{KKM00}
T.~Kloks, D.~Kratsch, and H.~M{\"u}ller.
\newblock Finding and counting small induced subgraphs efficiently.
\newblock \emph{Information Processing Letters}, 74\penalty0 (3):\penalty0
  115--121, 2000.

\bibitem[Kong and Valiant(2016)]{KV16}
W.~Kong and G.~Valiant.
\newblock Spectrum estimation from samples.
\newblock \emph{arXiv preprint arXiv:1602.00061}, 2016.

\bibitem[Le et~al.(2015)Le, Levina, and Vershynin]{LLV15}
C.~M. Le, E.~Levina, and R.~Vershynin.
\newblock Sparse random graphs: regularization and concentration of the
  laplacian.
\newblock \emph{arXiv preprint arXiv:1502.03049}, 2015.

\bibitem[Li and Woodruff(2016)]{LW16}
Y.~Li and D.~P. Woodruff.
\newblock On approximating functions of the singular values in a stream.
\newblock \emph{arXiv preprint arXiv:1604.08679}, 2016.

\bibitem[Li et~al.(2014)Li, Nguy{\^e}n, and Woodruff]{LNW14}
Y.~Li, H.~L. Nguy{\^e}n, and D.~P. Woodruff.
\newblock On sketching matrix norms and the top singular vector.
\newblock In \emph{Proceedings of the Twenty-Fifth Annual ACM-SIAM Symposium on
  Discrete Algorithms}, pages 1562--1581. Society for Industrial and Applied
  Mathematics, 2014.

\bibitem[Liu and Wang(2006)]{LW06}
H.~Liu and J.~Wang.
\newblock A new way to enumerate cycles in graph.
\newblock In \emph{AICT/ICIW}, page~57, 2006.

\bibitem[Mahoney et~al.(2011)]{rand1}
Michael~W Mahoney et~al.
\newblock Randomized algorithms for matrices and data.
\newblock \emph{Foundations and Trends{\textregistered} in Machine Learning},
  3\penalty0 (2):\penalty0 123--224, 2011.

\bibitem[Mason and Handscomb(2002)]{mason2002chebyshev}
J.~C. Mason and D.~C. Handscomb.
\newblock \emph{Chebyshev polynomials}.
\newblock CRC Press, 2002.

\bibitem[Negahban and Wainwright(2012)]{NW11}
S.~Negahban and M.~J. Wainwright.
\newblock Restricted strong convexity and (weighted) matrix completion: Optimal
  bounds with noise.
\newblock \emph{Journal of Machine Learning Research}, 2012.
\newblock To appear; posted at http://arxiv.org/abs/1009.2118.

\bibitem[Netrapalli et~al.(2014)Netrapalli, Niranjan, Sanghavi, Anandkumar, and
  Jain]{rpca2}
Praneeth Netrapalli, UN~Niranjan, Sujay Sanghavi, Animashree Anandkumar, and
  Prateek Jain.
\newblock Non-convex robust pca.
\newblock In \emph{Advances in Neural Information Processing Systems}, pages
  1107--1115, 2014.

\bibitem[Pace and Le{S}age(2004)]{PL04}
R.~K. Pace and J.~P. Le{S}age.
\newblock Chebyshev approximation of log-determinants of spatial weight
  matrices.
\newblock \emph{Computational Statistics \& Data Analysis}, 45\penalty0
  (2):\penalty0 179--196, 2004.

\bibitem[Polizzi(2009)]{eigen1}
Eric Polizzi.
\newblock Density-matrix-based algorithm for solving eigenvalue problems.
\newblock \emph{Physical Review B}, 79\penalty0 (11):\penalty0 115112, 2009.

\bibitem[Roosta-Khorasani and Ascher(2015)]{RA15}
F.~Roosta-Khorasani and U.~Ascher.
\newblock Improved bounds on sample size for implicit matrix trace estimators.
\newblock \emph{Foundations of Computational Mathematics}, 15\penalty0
  (5):\penalty0 1187--1212, 2015.

\bibitem[Rue and Held(2005)]{RH05}
H.~Rue and L.~Held.
\newblock \emph{Gaussian Markov random fields: theory and applications}.
\newblock CRC Press, 2005.

\bibitem[Saade et~al.(2015)Saade, Krzakala, and Zdeborov{\'a}]{saade2015matrix}
A.~Saade, F.~Krzakala, and L.~Zdeborov{\'a}.
\newblock Matrix completion from fewer entries: Spectral detectability and rank
  estimation.
\newblock In \emph{Advances in Neural Information Processing Systems}, pages
  1261--1269, 2015.

\bibitem[Sakurai and Sugiura(2003)]{eigen2}
Tetsuya Sakurai and Hiroshi Sugiura.
\newblock A projection method for generalized eigenvalue problems using
  numerical integration.
\newblock \emph{Journal of computational and applied mathematics}, 159\penalty0
  (1):\penalty0 119--128, 2003.

\bibitem[Schofield et~al.(2012)Schofield, Chelikowsky, and Saad]{eigen3}
G.~Schofield, J.~R. Chelikowsky, and Y.~Saad.
\newblock A spectrum slicing method for the kohn--sham problem.
\newblock \emph{Computer Physics Communications}, 183\penalty0 (3):\penalty0
  497--505, 2012.

\bibitem[Schudy and Sviridenko(2011)]{schudy2011bernstein}
W.~Schudy and M.~Sviridenko.
\newblock Bernstein-like concentration and moment inequalities for polynomials
  of independent random variables: multilinear case.
\newblock \emph{arXiv preprint arXiv:1109.5193}, 2011.

\bibitem[Shen-Orr et~al.(2002)Shen-Orr, Milo, Mangan, and Alon]{SMM02}
S.~S. Shen-Orr, R.~Milo, S.~Mangan, and U.~Alon.
\newblock Network motifs in the transcriptional regulation network of
  escherichia coli.
\newblock \emph{Nature genetics}, 31\penalty0 (1):\penalty0 64--68, 2002.

\bibitem[Stathopoulos et~al.(2013)Stathopoulos, Laeuchli, and Orginos]{SLO13}
A.~Stathopoulos, J.~Laeuchli, and K.~Orginos.
\newblock Hierarchical probing for estimating the trace of the matrix inverse
  on toroidal lattices.
\newblock \emph{SIAM Journal on Scientific Computing}, 35\penalty0
  (5):\penalty0 S299--S322, 2013.

\bibitem[Tian et~al.(2004)Tian, Jones, Villasenor, and
  Wesel]{tian2004selective}
T.~Tian, C.~R. Jones, J.~D. Villasenor, and R.~D. Wesel.
\newblock Selective avoidance of cycles in irregular ldpc code construction.
\newblock \emph{IEEE Transactions on Communications}, 52\penalty0 (8):\penalty0
  1242--1247, 2004.

\bibitem[Ugander et~al.(2013)Ugander, Backstrom, and Kleinberg]{UBK13}
J.~Ugander, L.~Backstrom, and J.~Kleinberg.
\newblock Subgraph frequencies: Mapping the empirical and extremal geography of
  large graph collections.
\newblock In \emph{Proceedings of the 22nd international conference on World
  Wide Web}, pages 1307--1318. ACM, 2013.

\bibitem[Van~Aelst and Rousseeuw(2009)]{VR09}
S.~Van~Aelst and P.~Rousseeuw.
\newblock Minimum volume ellipsoid.
\newblock \emph{Wiley Interdisciplinary Reviews: Computational Statistics},
  1\penalty0 (1):\penalty0 71--82, 2009.

\bibitem[Wang et~al.(2014)Wang, Lui, Ribeiro, Towsley, Zhao, and Guan]{WLR14}
P.~Wang, J.~Lui, B.~Ribeiro, D.~Towsley, J.~Zhao, and X.~Guan.
\newblock Efficiently estimating motif statistics of large networks.
\newblock \emph{ACM Transactions on Knowledge Discovery from Data (TKDD)},
  9\penalty0 (2):\penalty0 8, 2014.

\bibitem[Wigner(1955)]{wigner1955}
E.~P. Wigner.
\newblock Characteristic vectors of bordered matrices with infinite dimensions.
\newblock \emph{Annals of Mathematics}, page 548–564, 1955.

\bibitem[Wu et~al.(2016)Wu, Laeuchli, Kalantzis, Stathopoulos, and
  Gallopoulos]{WLK16}
L.~Wu, J.~Laeuchli, V.~Kalantzis, A.~Stathopoulos, and E.~Gallopoulos.
\newblock Estimating the trace of the matrix inverse by interpolating from the
  diagonal of an approximate inverse.
\newblock \emph{Journal of Computational Physics}, 326:\penalty0 828--844,
  2016.

\bibitem[Zhang and Leithead(2007)]{ZL07}
Y.~Zhang and W.~E. Leithead.
\newblock Approximate implementation of the logarithm of the matrix determinant
  in gaussian process regression.
\newblock \emph{Journal of Statistical Computation and Simulation}, 77\penalty0
  (4):\penalty0 329--348, 2007.

\bibitem[Zhang et~al.(2015)Zhang, Wainwright, and Jordan]{ZWJ15}
Y.~Zhang, M.~J. Wainwright, and M.~I. Jordan.
\newblock Distributed estimation of generalized matrix rank: Efficient
  algorithms and lower bounds.
\newblock \emph{arXiv preprint arXiv:1502.01403}, 2015.

\end{thebibliography}
